\documentclass{article}

\usepackage{arxiv,times}

\pdfoutput=1

\usepackage{hyperref}
\hypersetup{
    colorlinks=true,
    linkcolor=red,
    filecolor=magenta,
    urlcolor=blue,
    citecolor=purple,
    pdftitle={Overleaf Example},
    pdfpagemode=FullScreen,
    }

\usepackage[utf8]{inputenc} 
\usepackage[T1]{fontenc}    
\usepackage{url}            
\usepackage{booktabs}       
\usepackage{amsfonts}       
\usepackage{amsmath}
\usepackage{nicefrac}       
\usepackage{microtype}      
\usepackage[Algorithm,ruled]{algorithm}
\usepackage{algorithmicx}
\usepackage{algpseudocode}
\usepackage{bm}
\usepackage{graphicx}
\usepackage{natbib}
\usepackage{doi}
\usepackage{adjustbox}
\usepackage{float}

\usepackage{xcolor}

\title{Neural Distillation as a State Representation Bottleneck in Reinforcement Learning}

\author{Valentin Guillet \\
       ISAE-SUPAERO, Université de Toulouse, France \\
       \texttt{valentin.guillet@isae-supaero.fr} \\
       \And
       Dennis Wilson \\
       ISAE-SUPAERO, Université de Toulouse, France \\
       \texttt{dennis.wilson@isae-supaero.fr} \\
       \And
       Carlos Aguilar-Melchor \\
       Sandbox AQ, France \\
       \texttt{carlos@sandboxaq.com} \\
       \And
       Emmanuel Rachelson \\
       ISAE-SUPAERO, Université de Toulouse, France \\
       \texttt{emmanuel.rachelson@isae-supaero.fr} \\
}

\begin{document}
\maketitle

\let\thefootnote\relax\footnotetext{Work published at the 1st Conference on Lifelong Learning Agents (CoLLAs), 2022}

\begin{abstract}
    Learning a good state representation is a critical skill when dealing with multiple tasks in Reinforcement Learning as it allows for transfer and better generalization between tasks.
    However, defining what constitute a useful representation is far from simple and there is so far no standard method to find such an encoding.
    In this paper, we argue that distillation --- a process that aims at imitating a set of given policies with a single neural network --- can be used to learn a state representation displaying favorable characteristics.
    In this regard, we define three criteria that measure desirable features of a state encoding: the ability to select important variables in the input space, the ability to efficiently separate states according to their corresponding optimal action, and the robustness of the state encoding on new tasks.
    We first evaluate these criteria and verify the contribution of distillation on state representation on a toy environment based on the standard inverted pendulum problem, before extending our analysis on more complex visual tasks from the Atari and Procgen benchmarks.
\end{abstract}

\section{Introduction}

Despite the impressive successes of modern reinforcement learning (RL) \citep{sutton2018reinforcement} methods in designing efficient specialized control policies for a wide variety of difficult tasks, many studies have highlighted the limited ability of RL agents to generalize to variations of such tasks that would appear easy to a human being \citep{farebrother2018generalization, packer2018assessing, zhang2018study, song2020observational, cobbe2019quantifying}.
This work is motivated by the idea that networks trained for specific tasks build state representations that can easily be fooled by the ambiguity between observation variables.
For instance, in some platform video games, it is possible to design an optimal policy for a given level based solely on background features and progression indicators, rather than on the position of platforms and enemies \citep{song2020observational}.
While very efficient on this specific level, such a policy might not perform well on another.
Conversely, we formulate and evaluate the hypothesis that a network trained to imitate several such specialized policies on a limited set of task variations induces a state representation that lifts the ambiguity and filters out confounding observation variables.
Specifically, we investigate whether the process of network distillation \citep{hinton2015distilling,rusu2016policy}, inspired by knowledge consolidation in cognitive systems \citep{wilson1994reactivation,ashworth2014sleep,mcclelland1995why}, induces valuable state representations.
The interplay between distillation and state representation appears to have received little attention so far.
We endeavor to fill this gap and investigate how neural distillation can act as a state representation bottleneck in RL.

Our contributions are as follows:
\begin{itemize}
    \item We propose a generic experimental protocol to evaluate the effects of imitation (via distillation) on state representation.
    \item We demonstrate, on controlled experiments, that distillation across tasks acts as a filter over state representations that mitigates observational overfitting \citep{song2020observational} and its consequences.
    \item We extend this analysis to visual RL tasks from the Arcade Learning Environment \citep{bellemare2013arcade} and the Procgen \citep{cobbe2020leveraging} benchmark suites, and evaluate to what extent state representation disambiguation is a key to generalization across tasks.
\end{itemize}

Section \ref{sec:generalization} proposes a discussion and an overview of the links between the works in generalization and state representation in multi-task RL.
It motivates the study of the emergence of relevant state encodings from the distillation of previous knowledge.
Section \ref{sec:distillation} discusses properties we expect from a good state representation learning procedure, defines quantitative criteria, and presents the controlled experiment designed to evaluate them.
Section \ref{sec:visual} transposes this analysis to the more complex case of visual control tasks.
Section \ref{sec:conclu} summarizes our main findings and conclusions.

\section{Generalization and state representation in RL}
\label{sec:generalization}

Generalization is the ability for a supervised learning predictor to output a correct label for any sample that has not been seen during training.
This implicitly supposes that all samples $(x,y)$ stem from a unique distribution, whose density is often noted $p(x,y)$.
Then, given a loss $\mathcal{L}(y,y')$ and a sample set of $N$ samples $(x_i,y_i)$, closing the generalization gap consists of minimizing the difference between the risk and the empirical risk \citep{vapnik1992principles} $gap(f) = \mathbb{E}_{x,y\sim p}[\mathcal{L}(f(x),y)] - 1/N \sum_i \mathcal{L}(f(x_i),y_i)$.

Closing the generalization gap in RL is recognized as a difficult problem \citep{cobbe2019quantifying,song2020observational} that has been surveyed recently by \citet{kirk2021survey}.
Recent contributions on this topic exploit data augmentation \citep{raileanu2020automatic,laskin2020reinforcement,yarats2020image,lee2020network}, regularization \citep{cobbe2019quantifying,igl2019generalization,cobbe2020leveraging,wang2020improving,bertoin2022local},  or architecture separation and auxiliary losses \citep{raileanu2021decoupling}.
However, few of these works that aim at generalization properties of the policy explicitly draw a connection with the multi-task learning literature.
Such a connection can be found in the representation learning literature, either under the form of auxiliary losses \citep{bellemare2019geometric}, self-learning \citep{grill2020bootstrap}, or open-ended learning \citep{team2021open} for instance.

The general case of multi-task (supervised) learning \citep{caruana1997multitask,wilson2007multi-task,ruder2017overview,crawshaw2020multi} covers the problem of learning functions from \emph{several} distributions over $(x,y)$, each corresponding to a different task.
In that case, there is not necessarily a single optimal label that can be attached to a given $x$ and (empirical) risk minimization does not capture the full notion of generalization across tasks.
A key problem in this context becomes that of \emph{representation learning} \citep{bengio2013representation}, that is learning low-dimensional descriptive features $\varphi(x)$ that are useful to all tasks $p(x,y)$ (or at least a subset of tasks).
Formally, for all tasks $p(x,y)$, one searches for a low-dimensional $\varphi$ such that $\exists g \textrm{ s.t. } \mathbb{E}_{x,y\sim p} [ \mathcal{L}(g(\varphi(x)), y) ] \leq \epsilon$, for a given risk $\epsilon$.

This definition translates straightforwardly to RL: a desirable property for generalization of RL policies is the emergence of common useful features in policies trained on a subset of tasks.
Formally, for a given set of MDPs, and all corresponding optimal policies $\pi(a|s)$ inducing a state distribution $\rho^\pi$, one searches for a low-dimensional $\varphi$ such that $\exists g \textrm{ s.t. } \mathbb{E}_{s \sim \rho^\pi, a \sim \pi(\cdot |s)} [ \mathcal{L}(g(\varphi(s)), a) ] \leq \epsilon$, for a given risk $\epsilon$.
The definition of multi-task RL proposed by \citet{yu2020meta} covers a somewhat restricted case of multi-task learning, where there exists at least one function that performs well on all tasks.
This is essentially due to a task identifier being included in the agent's observation, separating the definition domains of different tasks' policies into disjoint sets, thus allowing to merge $g\circ \varphi$ into a single policy for all tasks.
In that framework, the question of representation learning translates to that of \emph{state representation learning} \citep{lesort2018state,raffin2019decoupling,caselles2021s,montoya2021decoupling,schwarzer2021pretraining}.
In particular, the recent work of \citet{grill2020bootstrap} illustrates how such state representations can be achieved in an unsupervised, task-independent way that exploits data augmentation and self-learning.
The line of work we develop here is somewhat orthogonal to their contribution since we focus on \emph{inter-task} imitation, while they exploit \emph{intra-task} data augmentation.

Methods tackling multi-task RL problems aim, for instance, at playing all Atari games, or all levels within a Procgen game. Since the corresponding visual states permit discriminating what game is being played, they already implicitly contain the task identifier information and indeed fall within the above definition.
Different approaches have been developed to tackle multi-task RL \citep{rusu2016progressive,fernando2017pathnet,hessel2019multi,sodhani2021multi}.
We focus here on those that perform imitation learning \citep{parisotto2016actor-mimic,rusu2016policy,teh2017distral} by distilling \citep{hinton2015distilling} the policies represented by several specialized neural networks into a common network performing all tasks.
The contribution of these methods to the question of learning good state representations in RL has not been discussed in the literature.
This work endeavors to fill this gap.

\section{Distillation as an information bottleneck}
\label{sec:distillation}

We wish to draw generic conclusions concerning the effect of distillation on state representation. For this purpose, this section proposes a generic, toy example, where useful state information is entangled with noise and redundant variables, to assess whether distillation is able to recover the useful variables.

\subsection{Evaluation criteria for distillation}
\label{sec:pend_eval_crit}

A good representation learning procedure should output a low-dimensional set of features that are useful for generalization to unseen states in the task at hand, and to the full span of tasks presented to the agent.
In contrast, a bad state encoding within a good policy for a given task transforms the state into features that will not be relevant in other tasks, or that do not permit generalization to new states within the same task.
For instance, in the CoinRun Procgen game, it is possible to learn a good policy for a given level, solely based on background elements and not the actual position of enemies, as illustrated by the saliency maps of \citet{song2020observational}. The resulting policy is good, but the learned state encoding does not permit generalization.
\citet{zahavy2016graying} illustrate that one can extract options from the state representation learned by DQN \citep{mnih2015human} on Atari games.
However, in these representations, two states with different actions could easily have very close state encodings, thus risking poor generalization to unseen states.

A desirable property for a learning procedure is that it outputs a policy that is both efficient and that relies on a state encoding that permits generalization. We translate this property into the following evaluation criteria:
\begin{itemize}
    \item \textbf{Ability to select important variables.} The final policy should depend strongly on important variables and filter out confounding, uninformative, redundant, or noisy inputs.
    \item \textbf{State separability.} The encoding should cluster together states sharing the same optimal action to ease generalization across states.
    \item \textbf{Robustness.} Application of the final policy to new tasks should preserve a certain amount of performance.
\end{itemize}

It is possible to interpret these criteria from a bias/variance tradeoff perspective.
One advantage of using (deep) neural networks is the small bias they introduce, compared to more constrained families of functions such as linear combination of features.
This, in turn, makes them more susceptible to focusing on confounding variables, which creates variance in the hypothesis space.
Although previous work showed how stochastic gradient descent acts as a regularizer \citep{gunasekar2017implicit}, empirical risk minimization remains intrinsically prone to variance.
In particular, \citet{song2020observational} illustrate how a high dimensional observation space with a low dimensional underlying state space can induce multiple solution policies, some of which are not generalizable to other MDPs.
Distillation then acts as a form of enhanced regularization, filtering out value functions (and the corresponding policies) that have equivalent empirical risk but bad generalization properties.
Conversely, such a mechanism would be less beneficial in hypothesis spaces that induce more bias and less variance (such as linear representations).

\subsection{Synthetic experimental protocol}
\label{sec:exp_pend}

We design a synthetic, toy benchmark, to evaluate these criteria, based on the inverted pendulum environment (Pendulum-v0) from the OpenAI Gym suite \citep{brockman2016gym}.
The purpose of this experiment is to control the introduction of noise and redundancy in state variables, in order to assess whether distillation is able to recover an efficient encoding of the state, in terms of the above criteria.
The original benchmark consists of bringing and stabilizing a pendulum to the upright position by applying discrete torque actions.
It features a 3-dimensional observation $(\cos \theta, \sin \theta, \dot{\theta})$ at each time step.
We introduce different ``levels'' of this same game by enriching the observation and making it 30-dimensional in a fashion inspired by the Madelon dataset \citep{guyon2003design}.
Specifically, for a given level, the variables are organized as follows:
\begin{itemize}
    \item Index 1 to 3. The original state variables.
    \item Index 4 to 13. For each level, we draw uniformly between -1 and 1 a fixed set of 10 linear combination weights that allow to recombine the three first variables together. This yields 10 new variables. This combination changes from one level to the next. This creates information redundancy within the level that does not permit generalization in other levels.
    \item Index 14 to 23. We draw uniformly between -1 and 1 a fixed set of 10 linear combination weights over the three first variables, yielding 10 more variables, to which we add a centered Gaussian noise ($\sigma = 0.05$). These weights are shared across levels, so this creates information redundancy that permits generalization but includes measurement noise.
    \item Index 24 to 26. A copy of 3 random variables among the 24 previous ones, re-drawn for each level.
    \item Index 28 to 30. Four centered Gaussian noise ($\sigma=1$) variables.
\end{itemize}

We draw $N_{levels}$ different levels of the benchmark and individually train (on each level) a C51 \citep{bellemare2017distributional} agent as implemented in Dopamine \citep{castro2018dopamine}, with default hyperparameters.
The network is adapted to take the 30-dimensional state as input, and features two hidden layers of size 32, with ReLU activation functions.
We refer to such networks as \emph{experts}.
As per the protocol of \citet{rusu2016policy} we then learn from scratch a distillation agent, referred to as \emph{student}, that imitates the distribution of outputs of the experts.
Specifically, we train the student to imitate the distribution of Q-values output by the C51 agent for each state.
The student plays a full episode on a given level and fills a level-specific replay buffer with examples of actions recommended by the corresponding expert, before switching to another level.
A slight difference with the protocol of \citet{rusu2016policy} is that, as recommended by \citet{parisotto2016actor-mimic}, we let the student (instead of the expert) draw which actions are performed on the level.
The mini-batches drawn to train the student at a given step are drawn  from the replay buffer of the current episode. Appendix \ref{app:pseudocode} provides a pseudo-code for this procedure.

\subsection{Results and discussion}
\label{sec:pend_results}

\begin{figure}
    \centering
    \includegraphics[width=0.32\textwidth]{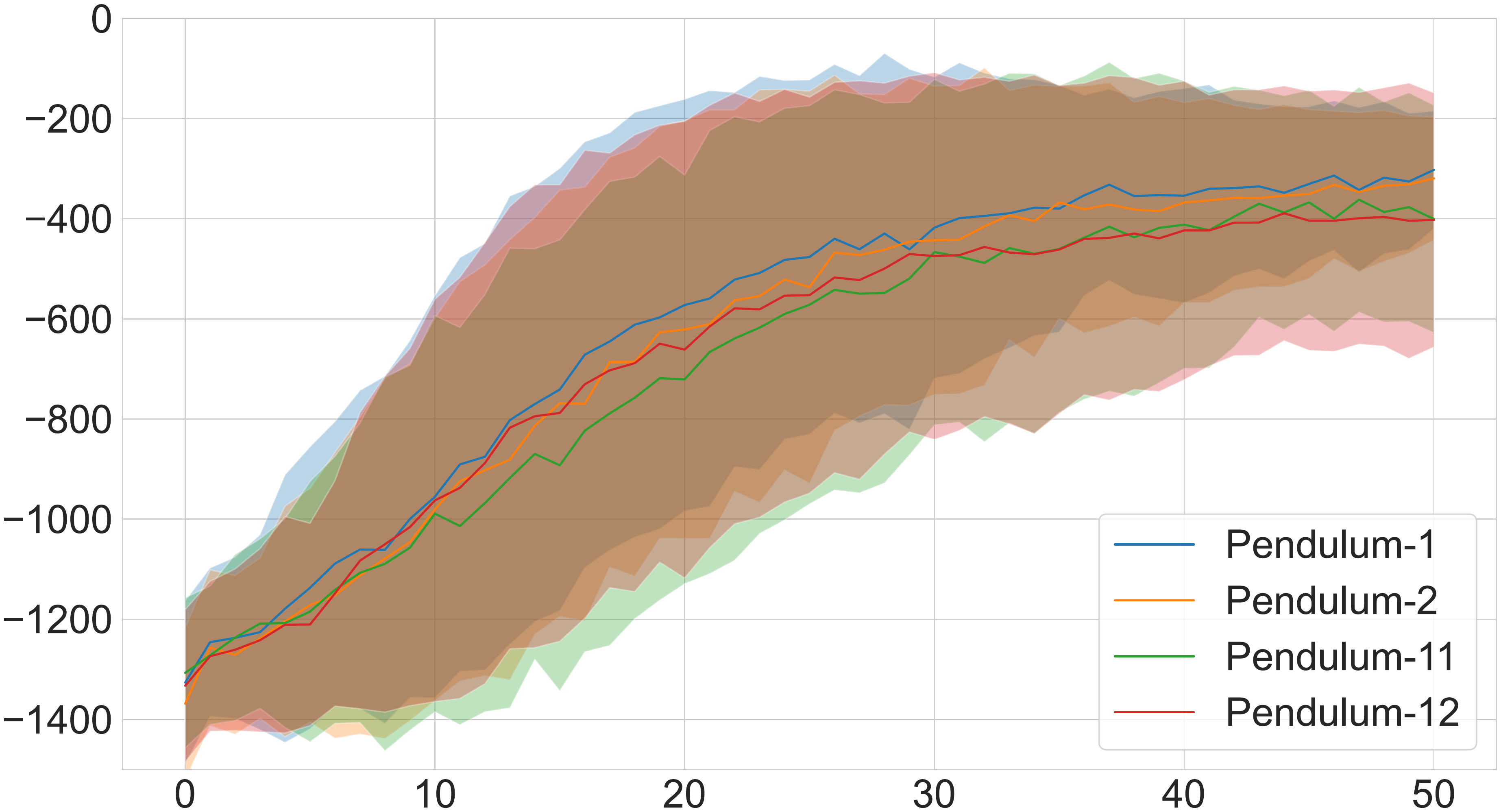}
    \includegraphics[width=0.32\textwidth]{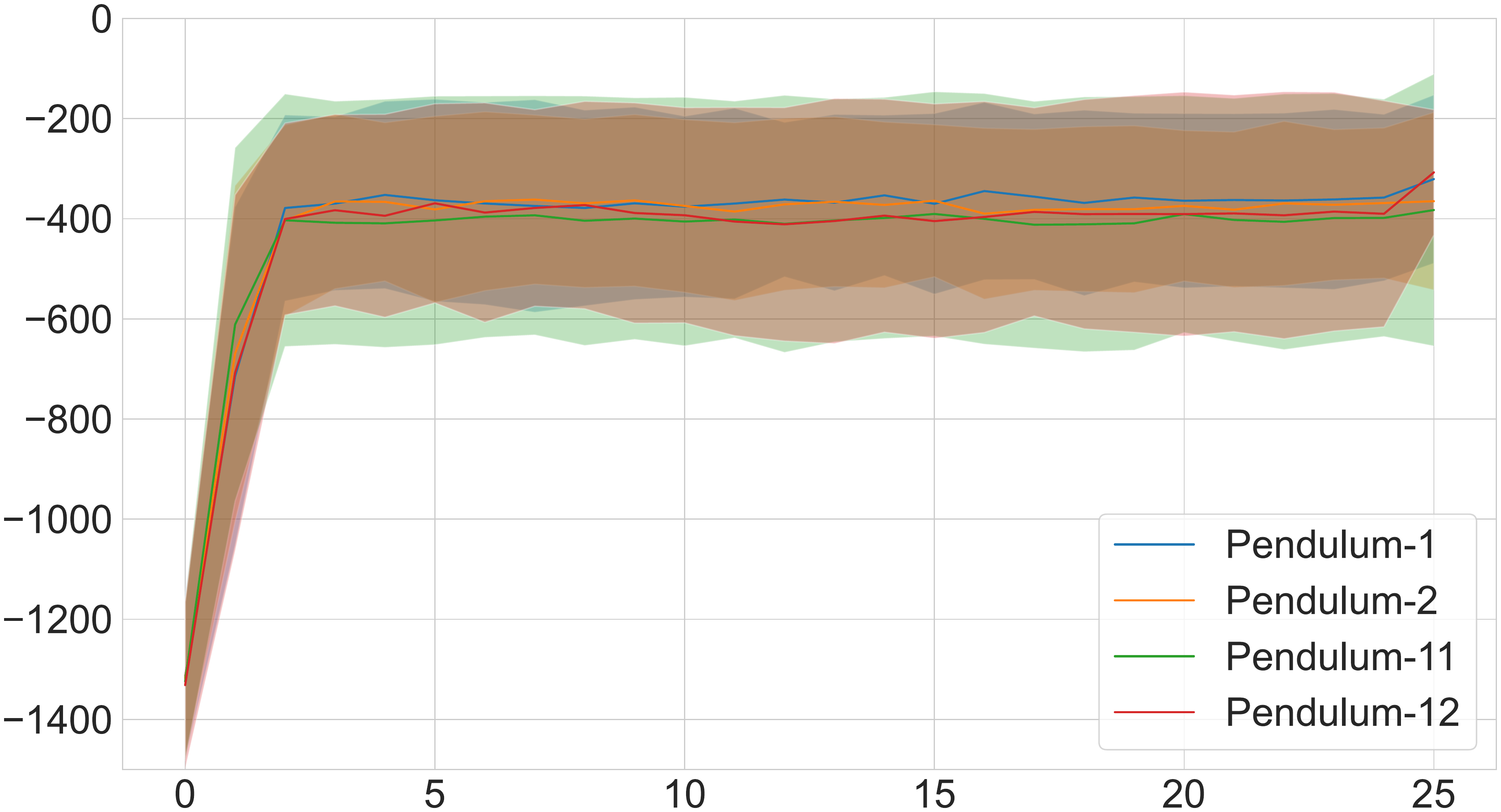}
    \includegraphics[width=0.32\textwidth]{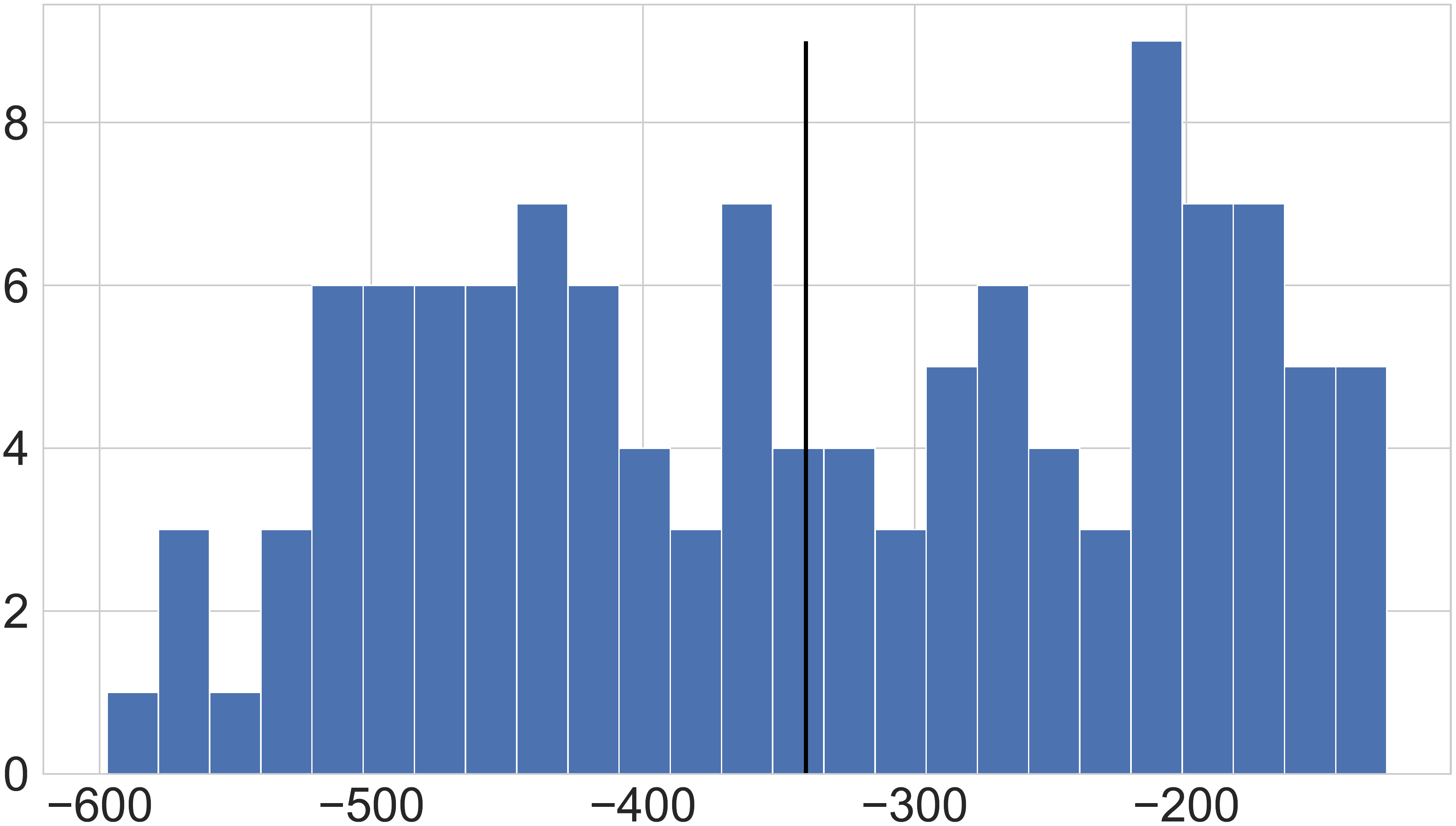}
    \caption{Training curves for experts (left) and student (middle), shaded area is one standard deviation}. Histogram of an expert's final score (right). 4 of the 12 levels are shown due to overlap in the remaining levels.
    \label{fig:train_expert_student}
\end{figure}

Figure \ref{fig:train_expert_student} reports the learning curves of all experts and the student.
A single curve in this figure corresponds to a single level and is the average, across 150 training runs, of the expert's score with respect to time (1 unit on the x-axis corresponds to 5 episodes of 200 time steps each).
Therefore, $150\times N_{levels}$ agents were trained overall and each $N_{levels}$-tuple of experts was distilled into a student, yielding 150 student networks. Unless stated otherwise, $N_{levels}=12$ in this section's discussion.
The evaluation is performed on a fixed grid of 50 randomly drawn initial states.
As all levels share the same underlying dynamics and reward models, the training curves of all experts are very similar and converge to the same average score.
The non-negligible training variance illustrates that the training process is rather sensitive and might fail.
To put this result in perspective, Figure \ref{fig:train_expert_student} also reports a representative histogram of the 150 experts' score after the last iteration, on a single level (the black bar indicates the average).

\begin{figure}[b]
    \centering
    \includegraphics[width=\textwidth]{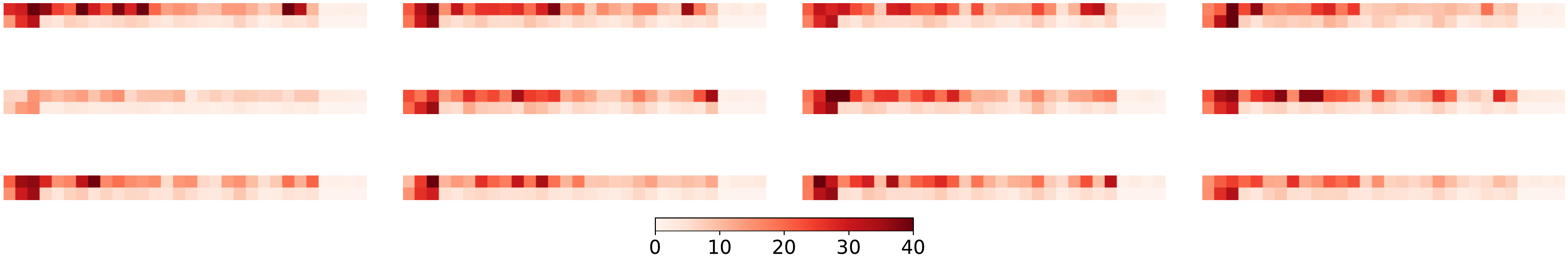}
    \caption{Feature importance in the experts (top lines) and the student (bottom lines) for all levels.}
    \label{fig:pend_feat_importance}
\end{figure}

To evaluate the ability of the student to filter out variables that do not lead to generalization, we evaluate each feature's importance for the experts and the student.
For this, we compute the gradient of the network's output with respect to the state variables.
Figure \ref{fig:pend_feat_importance} reports graphically the gradient amplitudes of the $N_{levels}$ experts and of the student, averaged over a pre-defined grid $\{s_j\}$ of states, for a subset of levels.
Each cell represents $\sum_j | (\partial net / \partial s^i) (s_j) |$, where $net$ is either the expert (top) or the student (bottom), and where $s^i$ is the $i$th state variable.
The lines corresponding to the experts in Figure \ref{fig:pend_feat_importance} (each top line) show that the expert relies both on the ``true'' state variables and on the confounding variables.
Specifically, as expected, the expert is fooled by the redundant variables ( variables 4 to 13).
To a certain extent, it is able to filter out the redundant but noisy variables (variables 14 to 23), while sometimes still relying on them to define the policy.
Also as expected, the repeated variables (variables 24 to 26) have a rather strong influence on the policy.
Finally, the experts are able to filter out the purely random variables (variables 28 to 30).
The bottom lines show that the student has filtered out all confounding variables and only relies on the three original state variables.
Interestingly, this property is true even for variables 14 to 23 even though these variables are shared across levels.
We conclude that consolidation on as few as 12 levels is enough to disambiguate between variables and filter out noise.

\begin{figure}
    \centering
    \includegraphics[width=\textwidth]{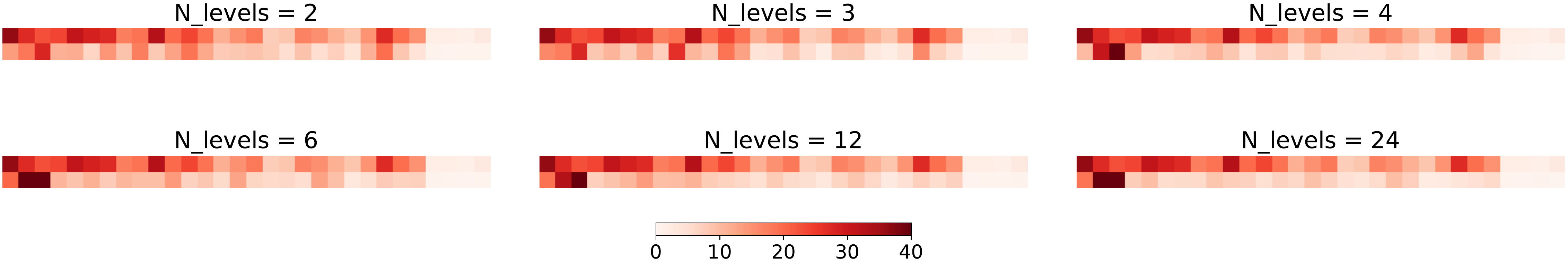}
    \caption{Feature importance as the number of experts increases of experts (top lines) and students (bottom lines).}
    \label{fig:pend_feat_levels}
\end{figure}

In order to assess the impact of $N_{levels}$, Figure \ref{fig:pend_feat_levels} reports the gradient amplitudes on a single level, when the student is trained to imitate a growing number of experts.
The filtering effect of distilling the experts into the student seems to appear with as little as 4 experts (with minor dependence on some confounding variables) and persists consistently for larger numbers of training levels.

\begin{figure}
    \centering
    \includegraphics[width=0.49\textwidth]{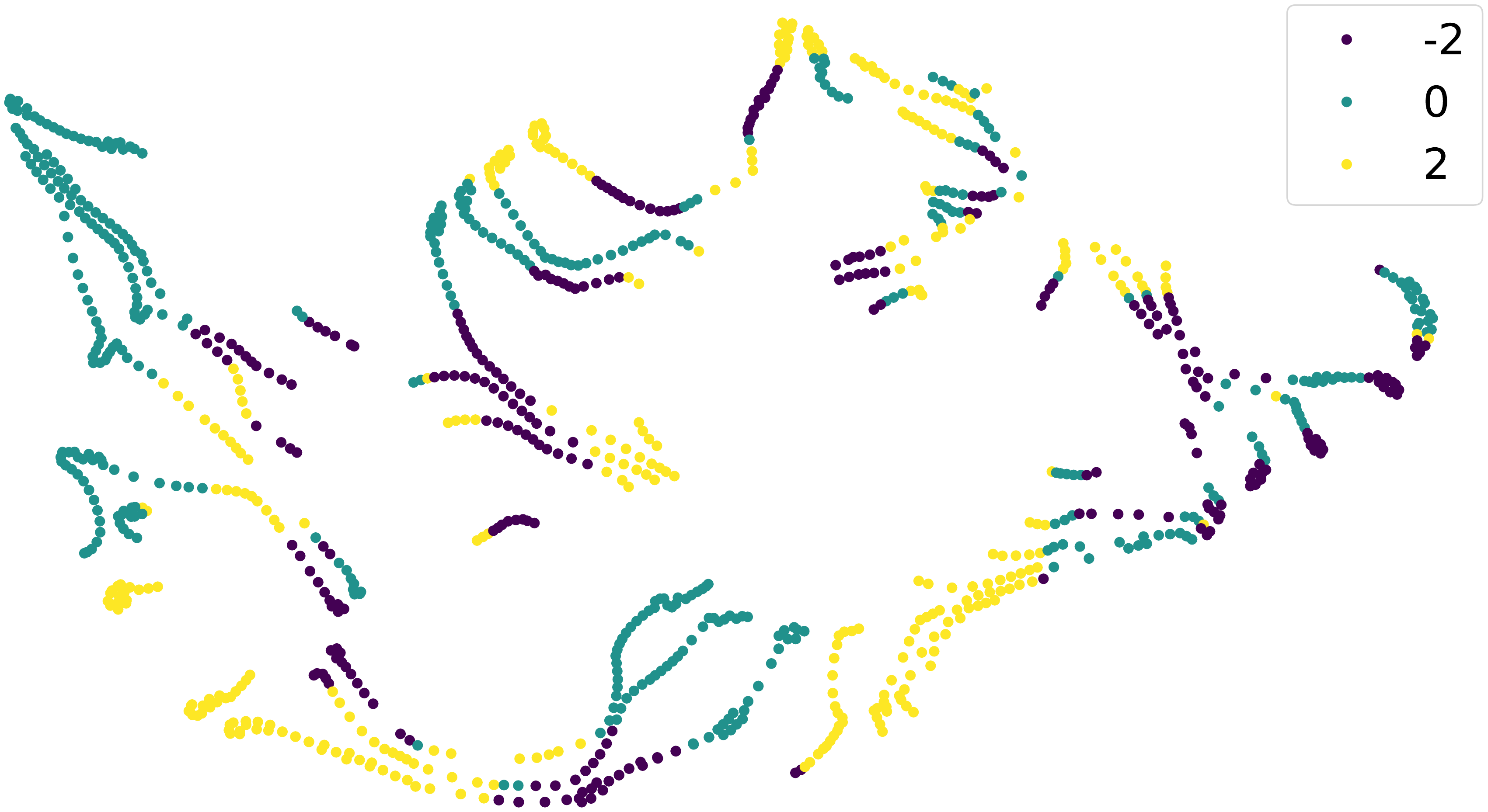}
    \includegraphics[width=0.49\textwidth]{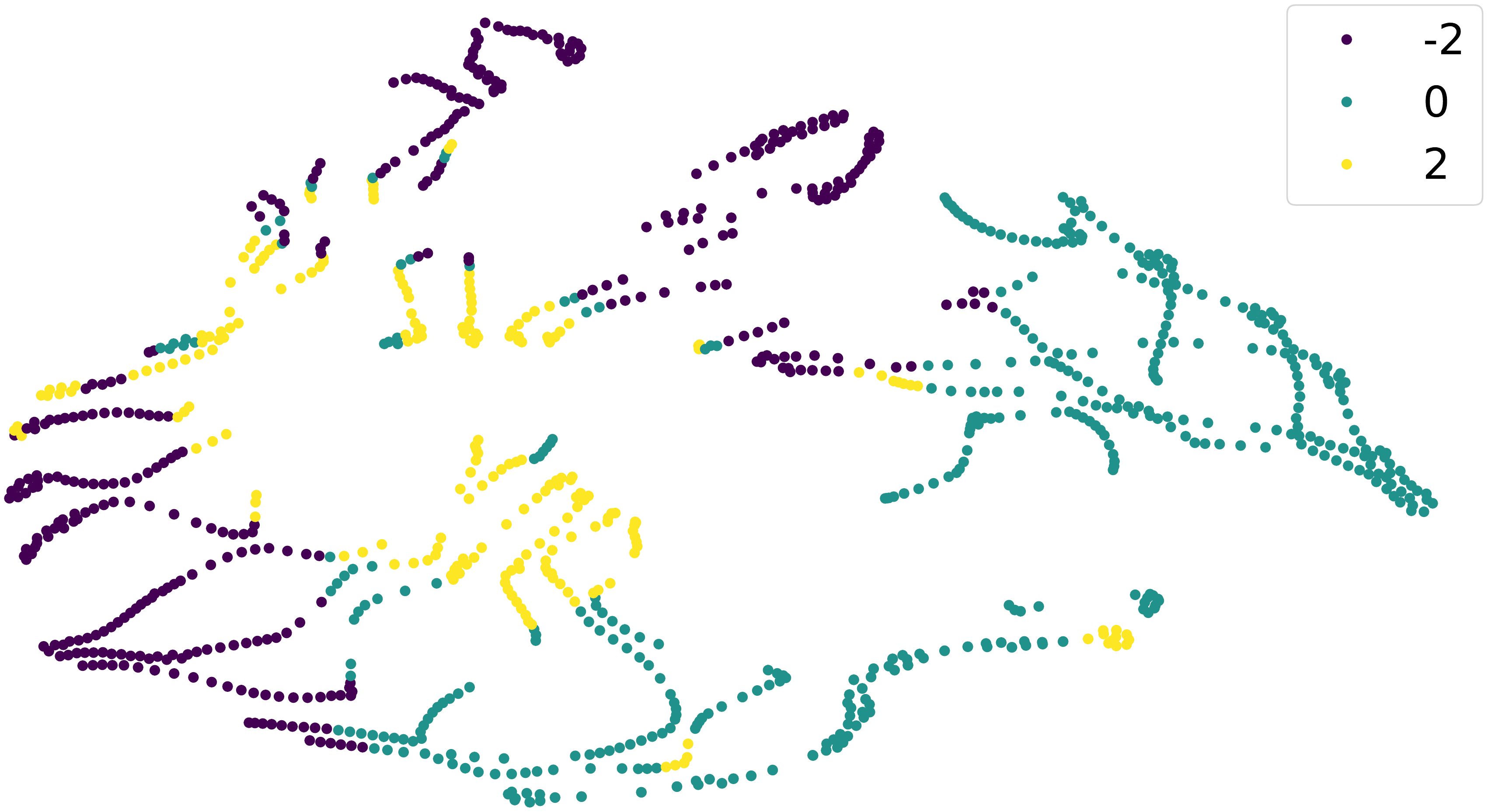}\\
    \vspace{1em}
    \includegraphics[width=0.49\textwidth]{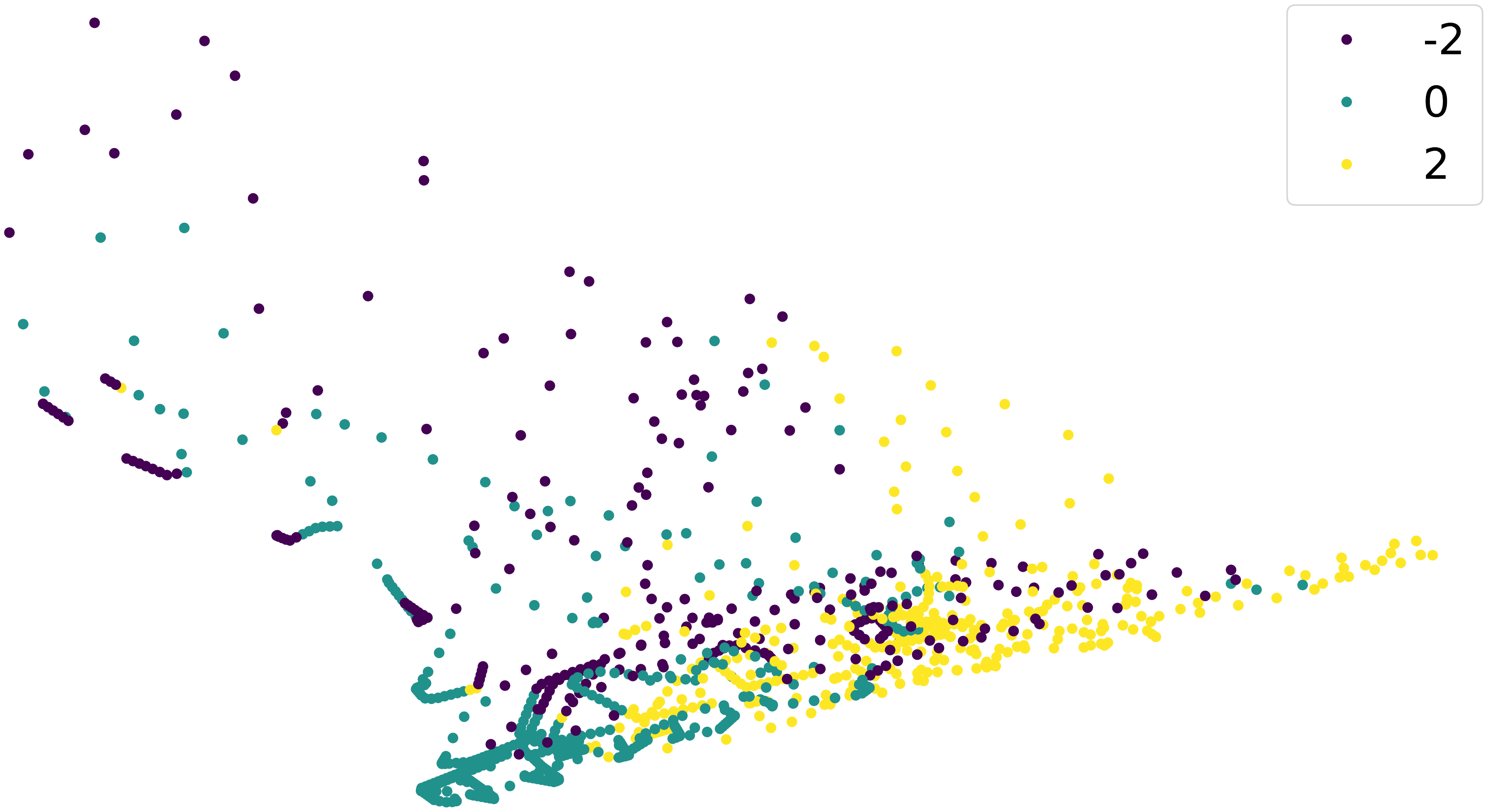}
    \includegraphics[width=0.49\textwidth]{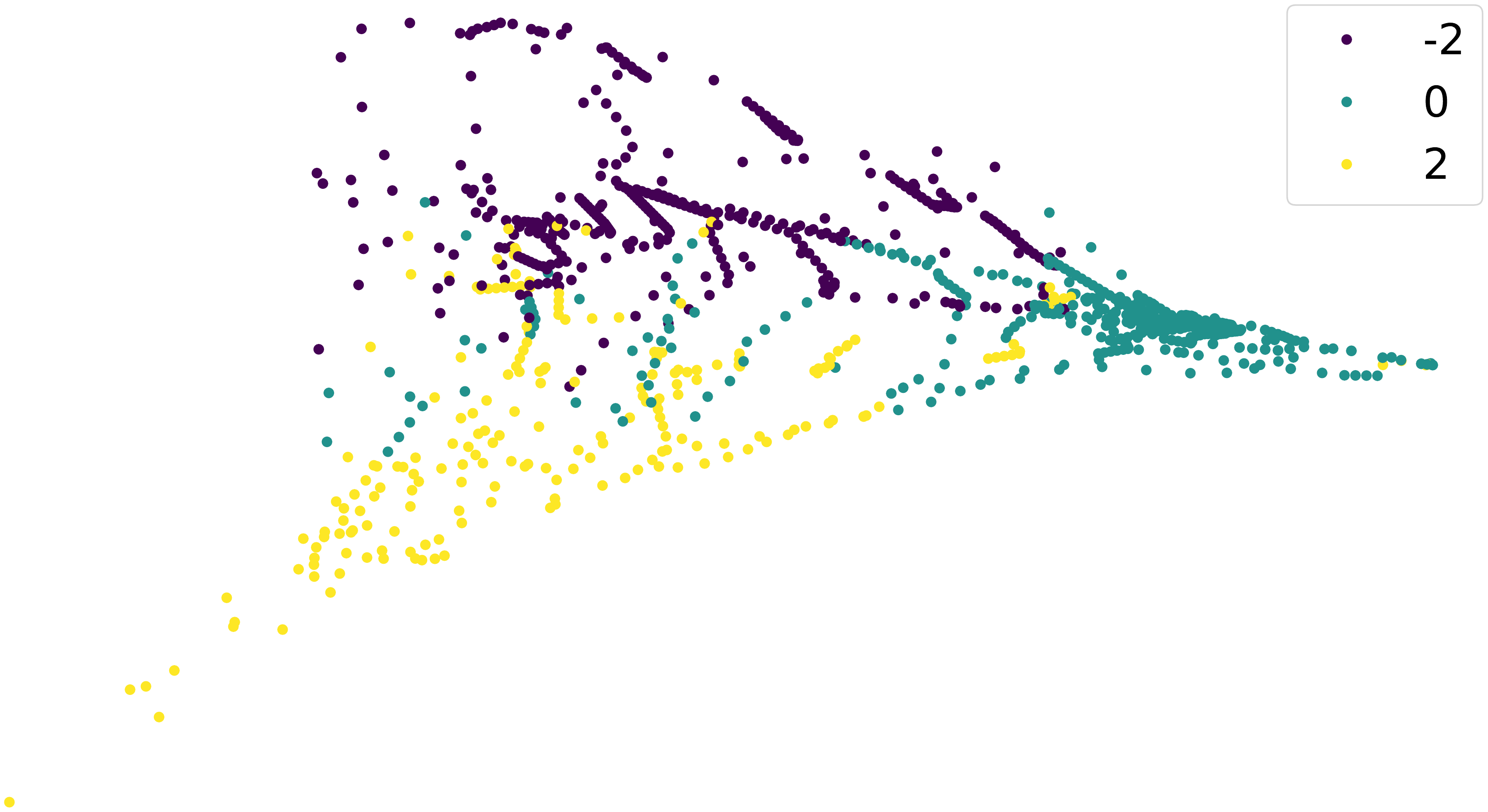}
    \caption{t-SNE (top) and PLS (bottom) plots of the state embedding for an expert (left) and the student (right).}
    \label{fig:pend_tsne}
\end{figure}

In order to evaluate the second criterion of Section \ref{sec:pend_eval_crit}, state separability, we plot a t-SNE visualization  \citep{van2008visualizing} of the state embedding, i.e. of the features after the second hidden layer.
Since t-SNE captures correlations between descriptive variables but not correlations with the network's output, we also compute a partial least squares regression (PLS) \citep{wold1984collinearity} of the action taken by the agent with respect to the embedding and plot the projection of the state embeddings within the two first principal directions.
Figure \ref{fig:pend_tsne} reports these visualizations of the state representation embedding, where we color points according to their optimal action (-2, 0, or 2 with our discretized inverted pendulum).
Graphically, it appears that distillation favors better proximity of states that share the same action.
In order to lift any graphical interpretation bias and quantify this measure of separability, we train a linear SVM to separate the data points in the
space spanned by the PLS projection and report the accuracy of the obtained classifier in Table \ref{tab:pend_pls} (more detailed results in Appendix \ref{app:pend_accuracy}).
Specifically, we train this linear SVM both when using the projection on the 3 first principal directions computed by the PLS and on all 32 directions (which is equivalent to not computing the principal directions at all).
The results consistently demonstrate that (except for $N_{levels}=2$), the embedding of the student is better suited for linear separability than that of the expert (whether we use 3 or 32 dimensions).
Increasing the variety of training levels monotonically increases the linear separability of states sharing the same action.

\begin{table}
    \begin{adjustbox}{width=\columnwidth,center}
    \begin{tabular}{c|cc|cc|cc|cc}
         & \multicolumn{2}{c|}{$N_{levels}=2$}  & \multicolumn{2}{c|}{$N_{levels}=6$} & \multicolumn{2}{c|}{$N_{levels}=12$}    & \multicolumn{2}{c}{$N_{levels}=24$} \\
         & dim = 3 & dim = 32       & dim = 3 & dim = 32        & dim = 3 & dim = 32        & dim = 3 & dim = 32\\
         \hline
        expert & 78.05 (4.90) & 91.29 (2.06) & 78.05 (4.90) & 91.29 (2.06) & 78.05 (4.90) & 91.29 (2.06) & 78.05 (4.90) & 91.29 (2.06) \\
        student & 75.56 (6.27) & 91.60 (2.16) & 83.06 (5.66) & 93.68 (1.87) & 88.03 (1.56) & 95.37 (0.75) & 91.01 (4.60) & 95.81 (1.78)
    \end{tabular}
    \end{adjustbox}

    \caption{Average (and standard deviation of) accuracy for a linear classifier built on the state representation.}
    \label{tab:pend_pls}
\end{table}

\begin{figure}
    \centering
    \includegraphics[width=0.32\textwidth]{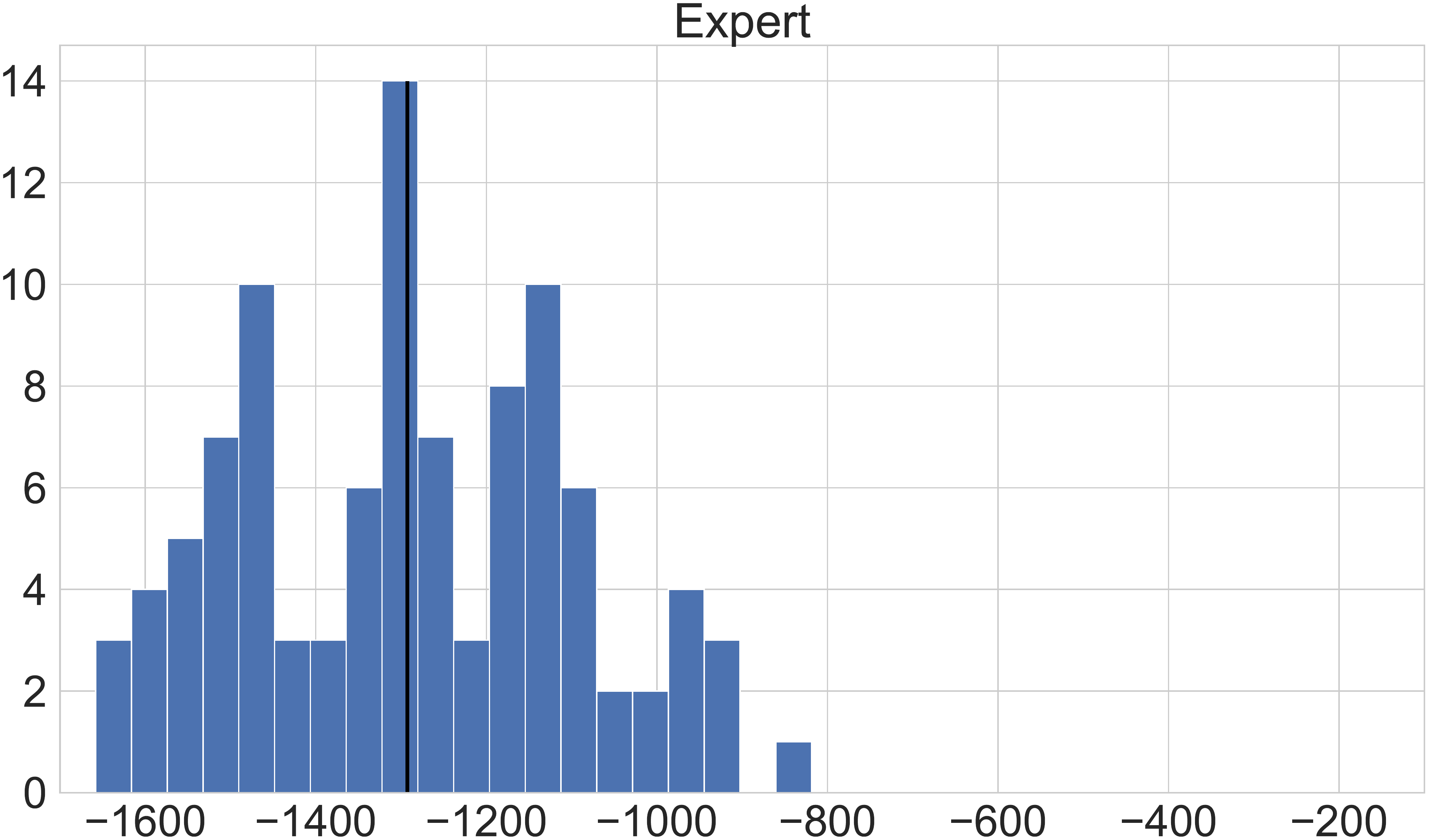}
    \includegraphics[width=0.32\textwidth]{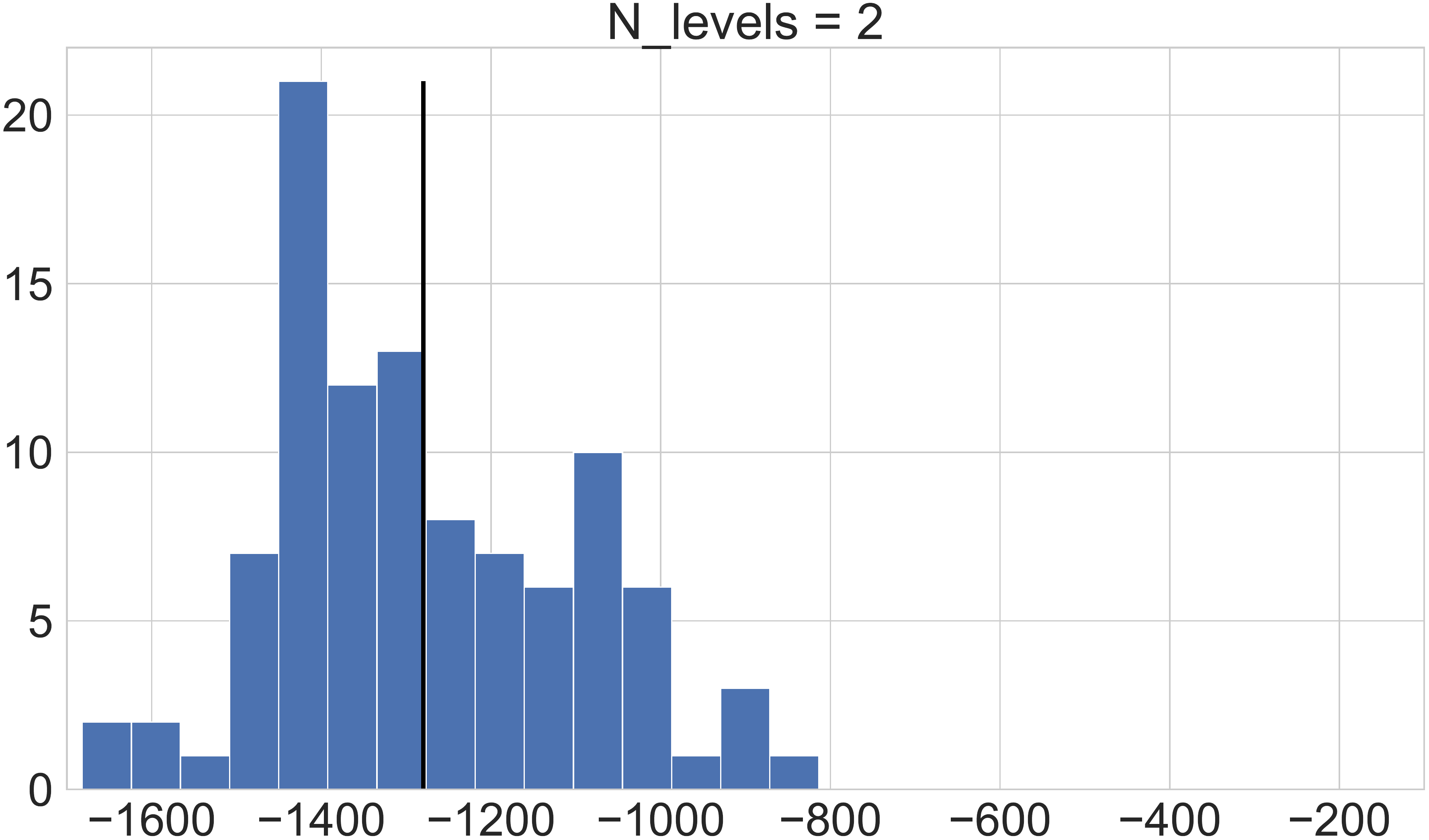}
    \includegraphics[width=0.32\textwidth]{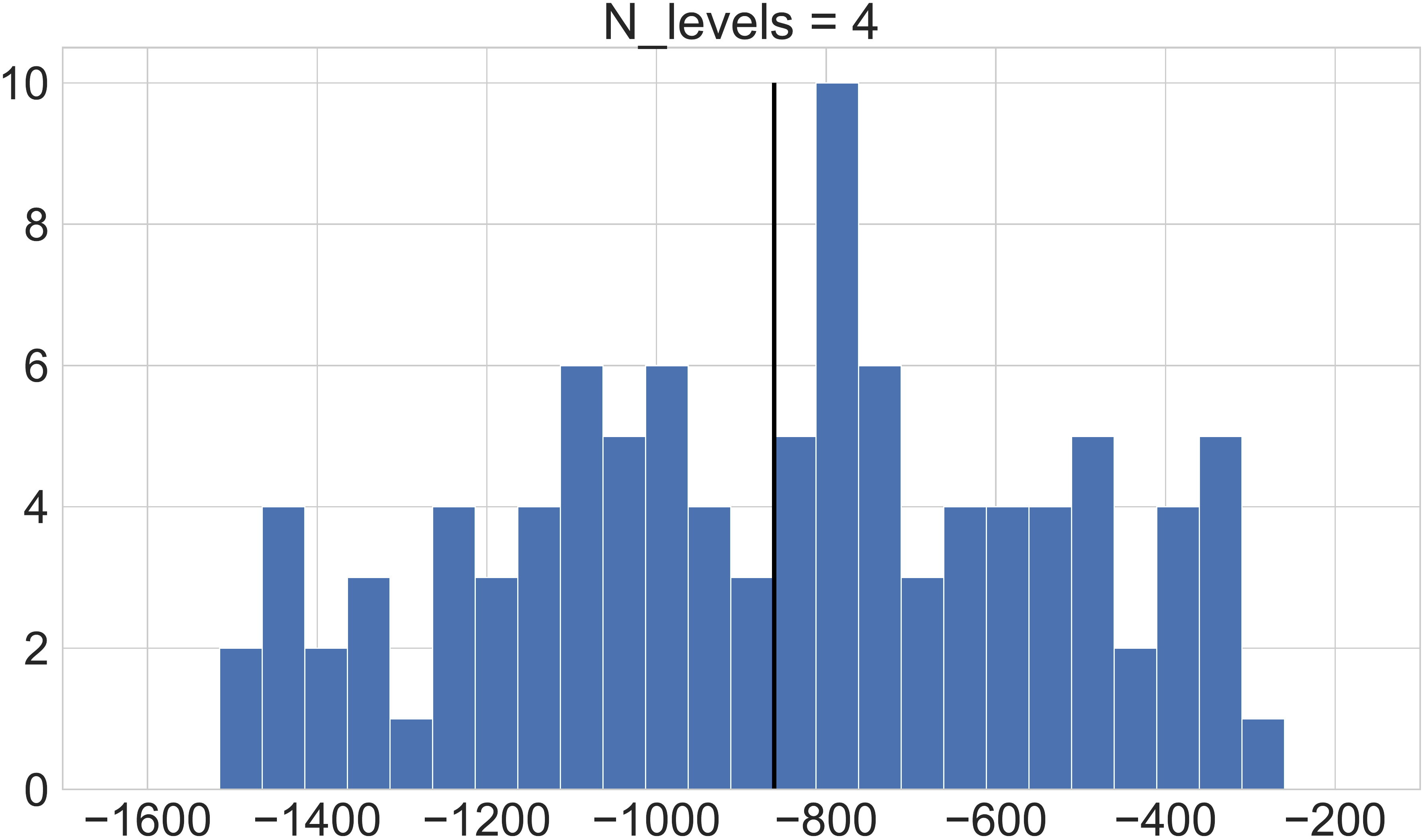}
    \includegraphics[width=0.32\textwidth]{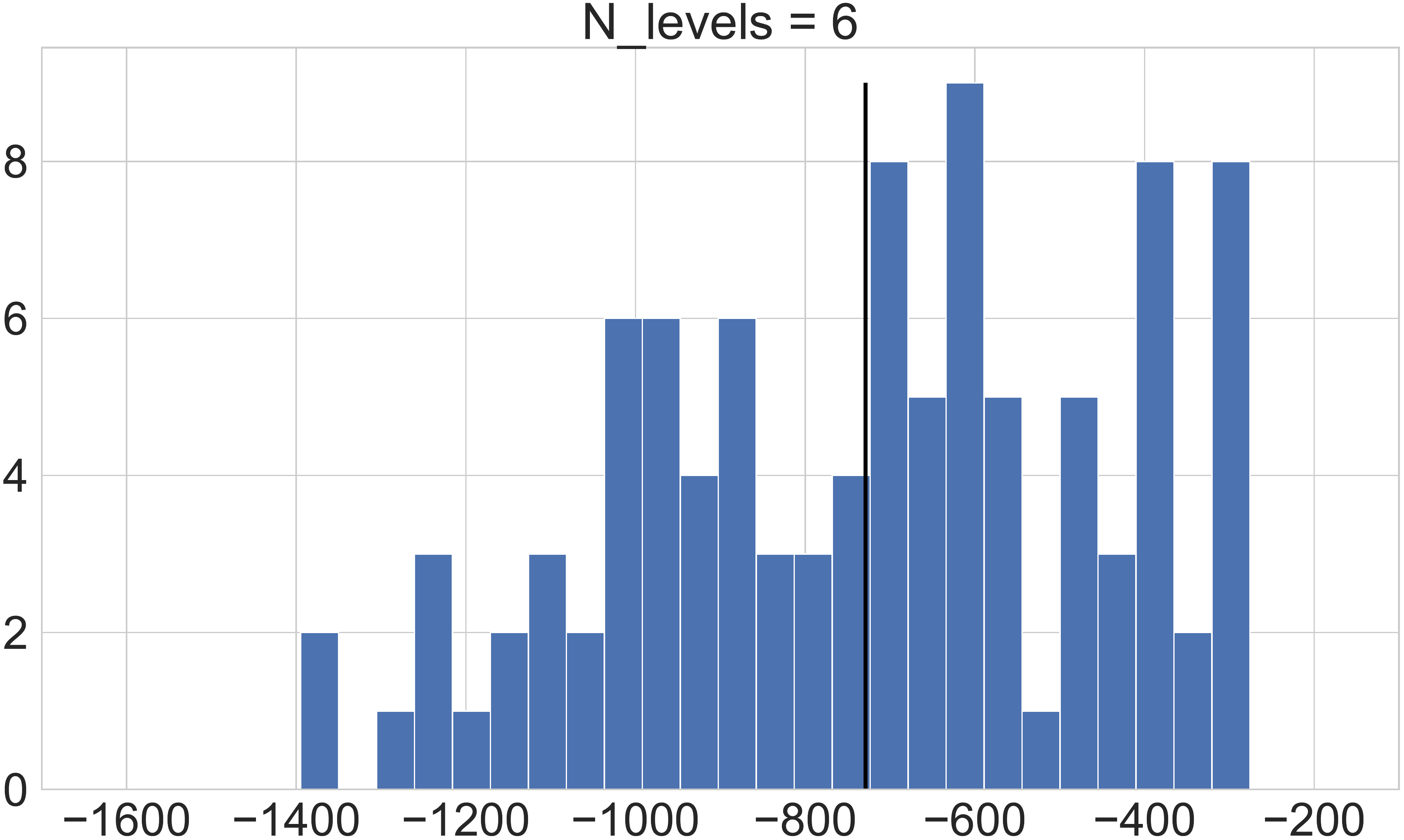}
    \includegraphics[width=0.32\textwidth]{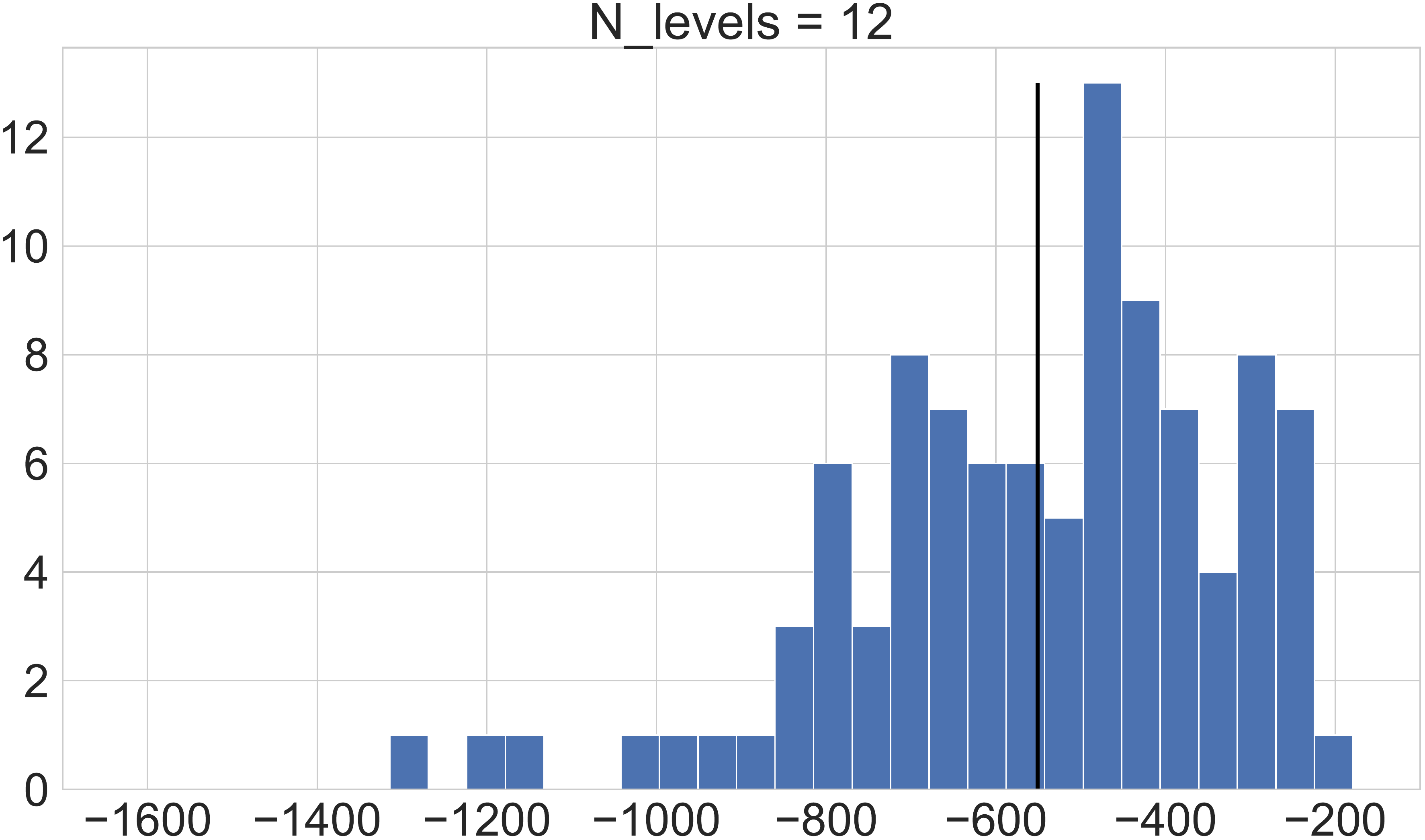}
    \includegraphics[width=0.32\textwidth]{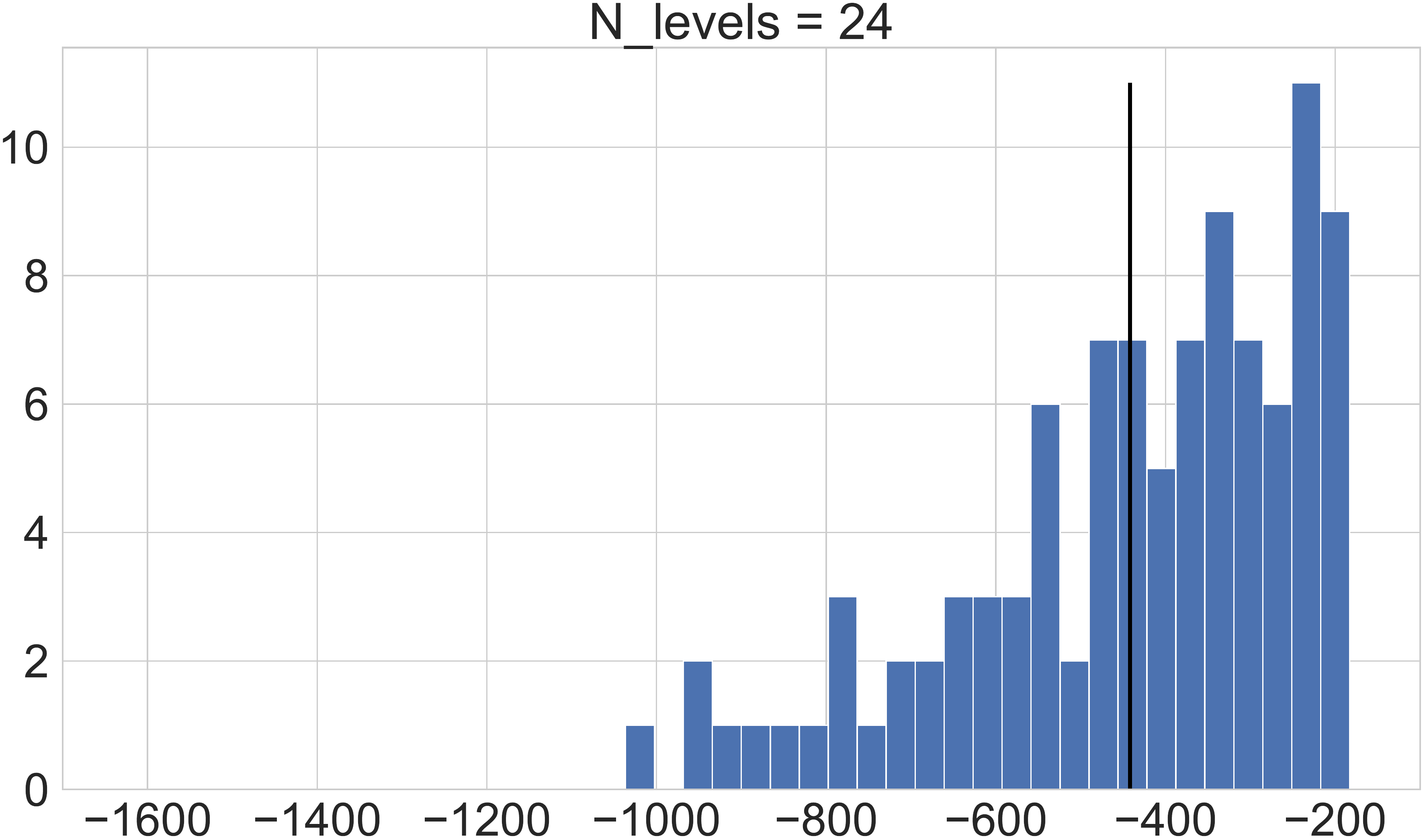}
    \caption{Robustness of an expert and of the student policy for varying values of $N_{levels}$.}
    \label{fig:pend_robust}
\end{figure}

We expect the student policy to perform well on a variety of levels, both those upon which the experts were trained and new ones. Figure \ref{fig:pend_robust} reports the distribution of scores obtained by an expert and by the student on a set of testing levels, drawn independently from the ones the experts were trained on.
The black bar indicates the distribution's mean.
As $N_{levels}$ increases, the student becomes more and more robust to new levels and, eventually, its average score reaches that of individual experts.
Note that this is different from identifying relevant variables (which happens already with 4 levels), and from performing as well as each expert individually on the training levels (which was illustrated on Figure~\ref{fig:train_expert_student}).

While this toy problem consisted of a simple task, inverted pendulum, the results clearly demonstrate that distillation can create new student networks which focus on important features, ignore confounding features, and reach the performance of expert networks, benefiting from additional experts trained on different levels.
One possible perspective of this work (although beyond the scope of this paper) is to design a curriculum of training tasks that minimize the number of corresponding experts which are necessary for good generalization performance.

\section{Visual state encodings from distillation}
\label{sec:visual}

The previous section quantified the effect of distillation as an information bottleneck on the state representation, via the proposed controlled experiment.
This bottleneck filters out information redundancy and noise, leading to features that permit better generalization and robustness.
The next question is to assess whether this type of filtering is actually a missing feature that would improve generalization and robustness in applications.
For instance, do this experimental protocol's results remain when evaluated on a set of Atari games or a variety of levels for several Procgen games?

\subsection{Visual benchmarks}
\label{sec:visual_benchmarks}

The evaluation protocol we follow remains essentially the same, with slight adaptations to the visual RL tasks at hand.
Specifically, the network architecture is now the classic DQN architecture from \citet{mnih2015human}, adapted to the C51 output \citep{bellemare2017distributional}.
The training algorithm we use for the experts is the Dopamine \citep{castro2018dopamine} implementation of Rainbow \citep{hessel2018rainbow}.
For the student, we adapt the Actor-Mimic Networks (AMN) algorithm of \citet{parisotto2016actor-mimic}.
Compared to plain distillation, AMN adds a feature regression loss that regularizes the features of the student network towards the features of the experts, which was shown to improve the distillation efficiency.
The goal for this regularization term is not to reproduce exactly the expert features, but rather to constrain the AMN features to contain the same information as them.
An adaptation layer is therefore added after the AMN feature layer so that the order and dimension of each feature vectors does not matter for the regularization term.
Contrarily to the original AMN implementation that applies the feature regression loss on the last hidden layer, we instead apply it to the output of the last convolutional layer,
since we would like to encourage structured feature maps that permit generalization, rather than less structured feature spaces after fully-connected layers (pseudo-code in Appendix \ref{app:pseudocode}).

To evaluate the feature importance criterion of Section \ref{sec:pend_eval_crit}, instead of computing network gradients with respect to the input, we turn to saliency maps, which essentially represent the same information in a more structured and interpretable fashion.
Specifically, we implement perturbation-based saliency maps, since they seem to provide the most interpretable results \citep{zahavy2016graying,greydanus2018visualizing,puri2020explain,huber2021local}.
We retain an analysis in terms of t-SNE and PLS embeddings for the separability of states in the representation space.
Finally, the robustness criterion is adapted to the different situations as discussed below.

Many Atari games share common visual features. For instance, Pong, Breakout and VideoPinball need to detect the position of a ball.
It seems desirable that the policy depends on this position.
Conversely, SpaceInvaders and Carnival share similar visual features, but it is unclear whether they can be reused in a common policy.
It is important to note that Atari games enjoy completely disjoint observation spaces and unrelated dynamics and reward models.
This does not prevent applying distillation on the corresponding agents, without assuming that the policies will be transferable from one game to another.
In this case, the intention of performing expert distillation on this type of benchmark, where the underlying dynamics are very different from one game to the other, is really to extract meaningful state representations and avoid concentrating on the ``wrong'' features, keeping in mind that it is likely that few features might transfer from one game to another.
We test these intuitions on the five aforementioned games.

The case of Procgen games is different, as this benchmark was designed precisely to evaluate generalization across visual variations of the same game, and there is thus an existing measure of task similarity.
Within the same Procgen game, we aim to both extract features that remain valid across levels, and evaluate how the student agent generalizes to unseen levels.
Hence, this benchmark appears much closer to the controlled experiment of Section \ref{sec:distillation}.
We focus specifically on CoinRun, BigFish and Jumper. On each game, we train experts on $N_{levels}=50$ levels, with each expert trained on a single level, then train the student network based on all expert networks.

\subsection{Results on Visual Tasks}
\label{sec:visual_results}

\begin{figure}
    \centering
    \includegraphics[width=0.32\textwidth]{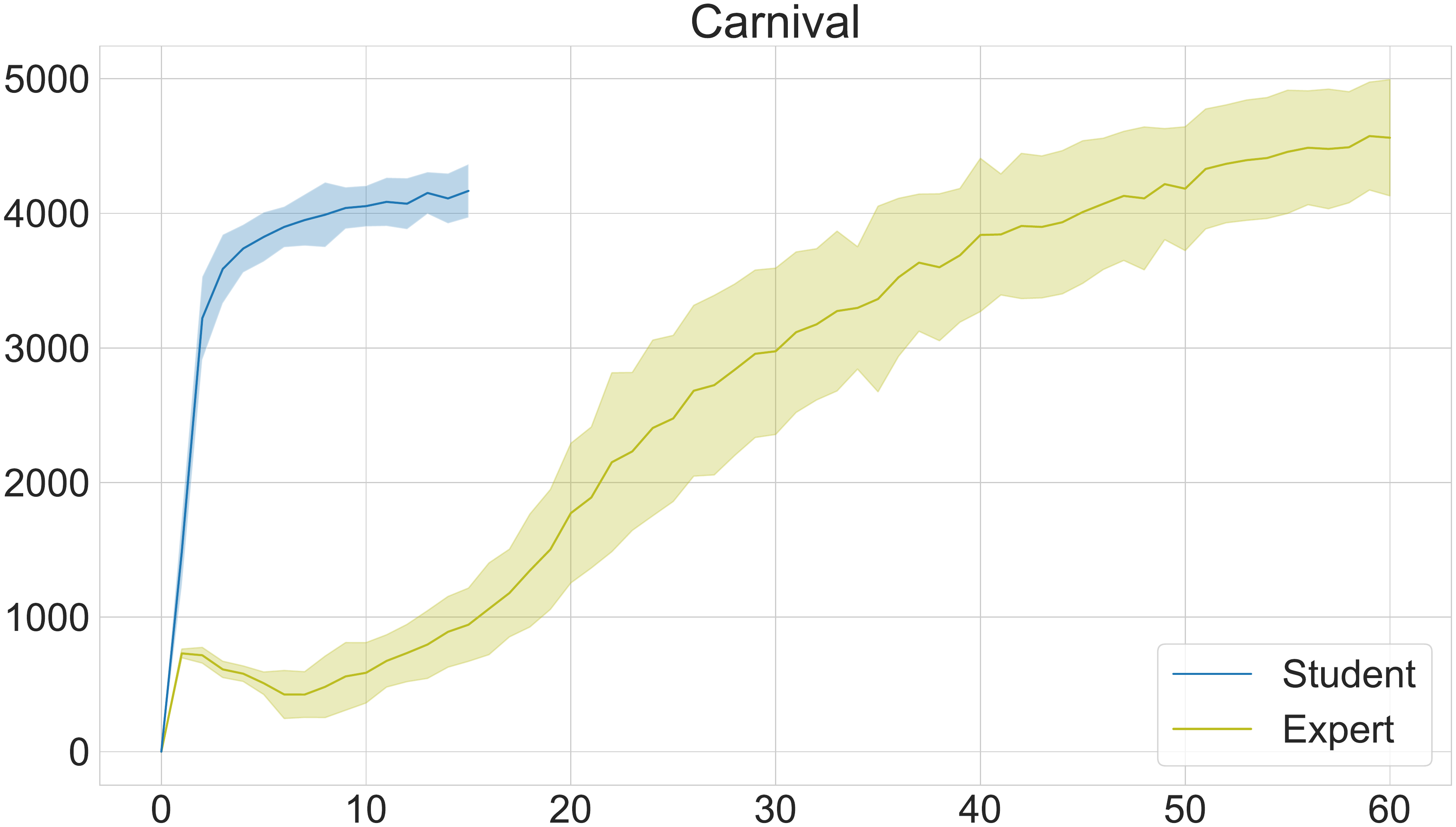}
    \includegraphics[width=0.32\textwidth]{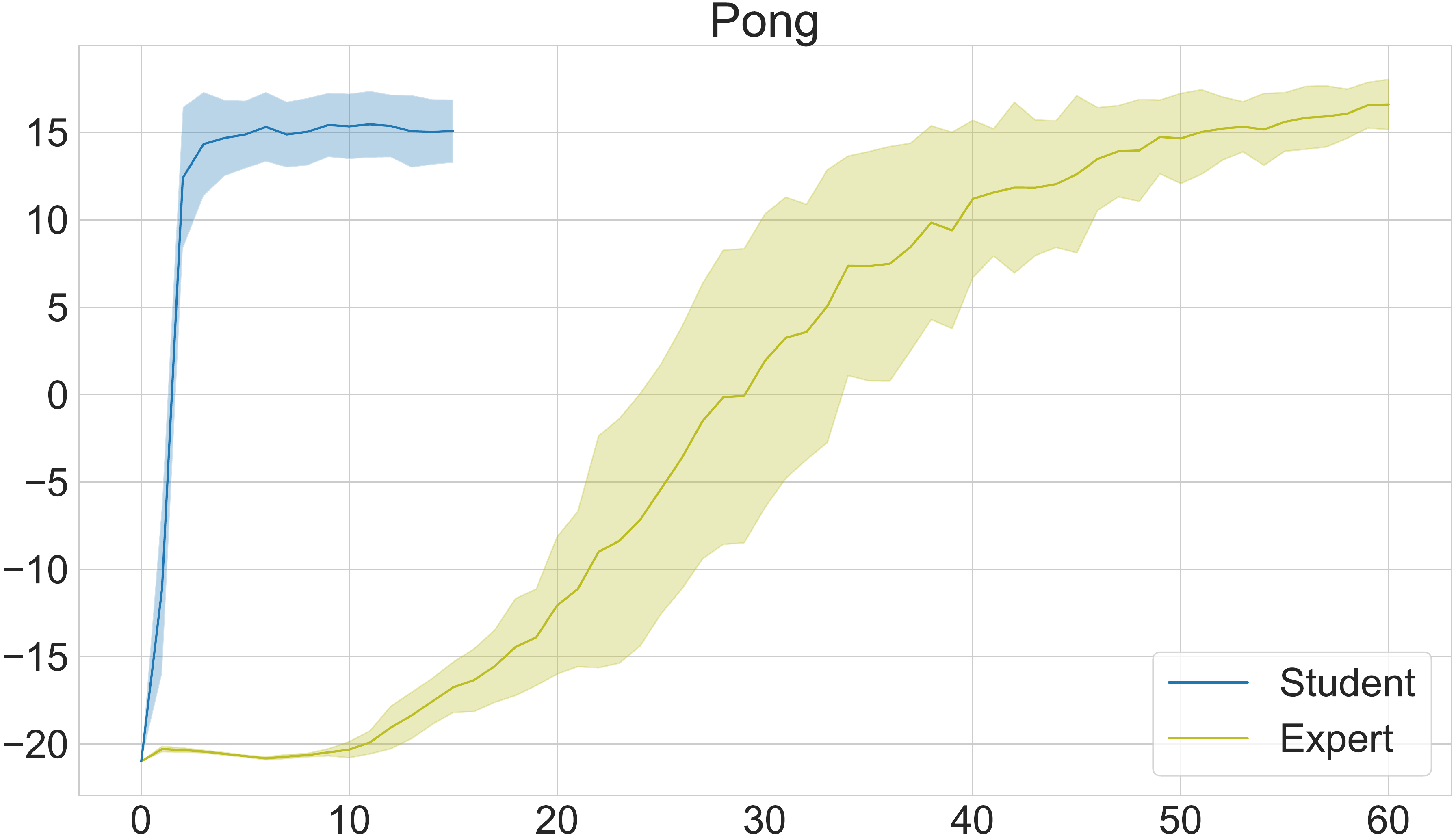}
    \includegraphics[width=0.32\textwidth]{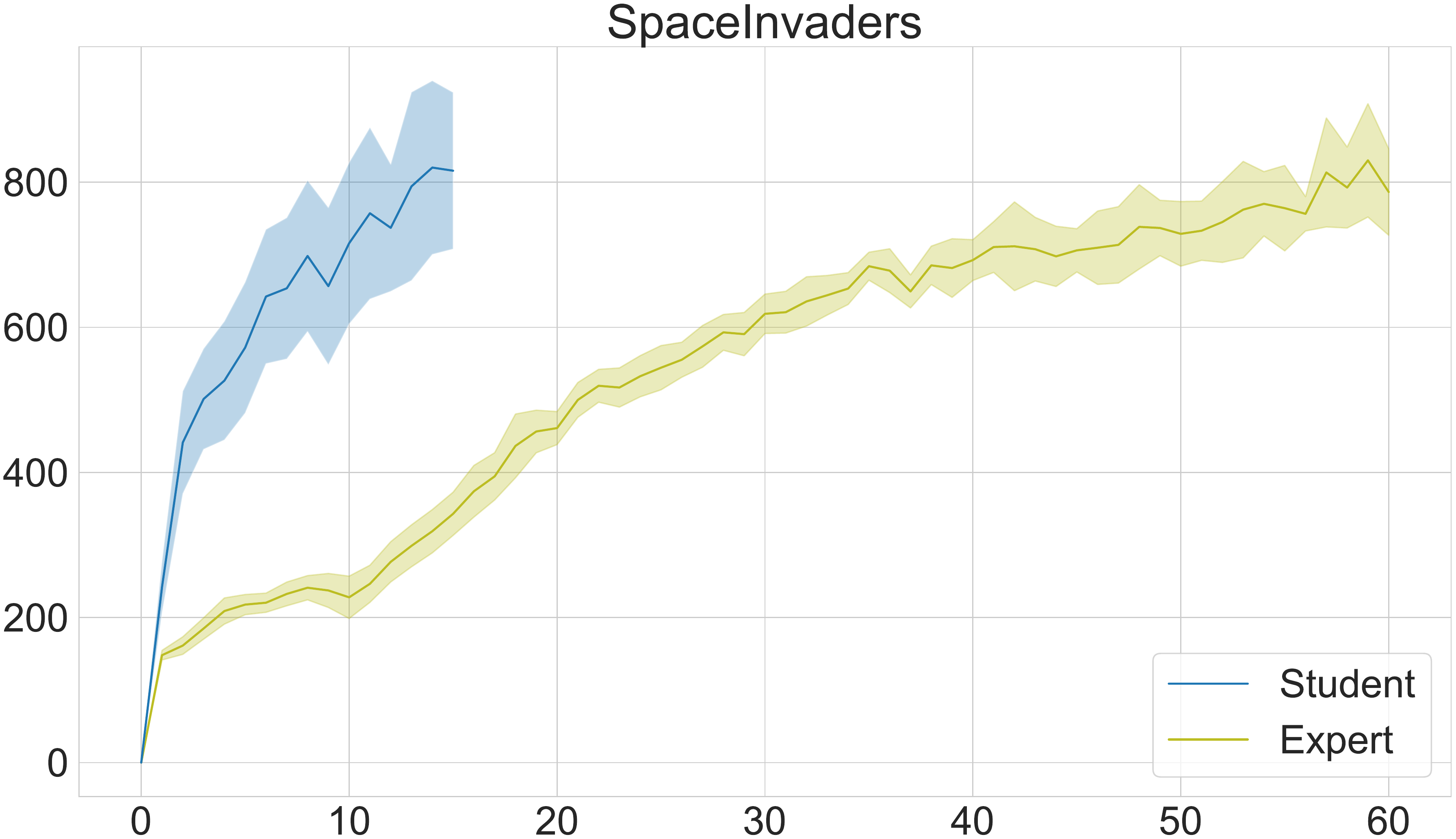}

    \includegraphics[width=0.32\textwidth]{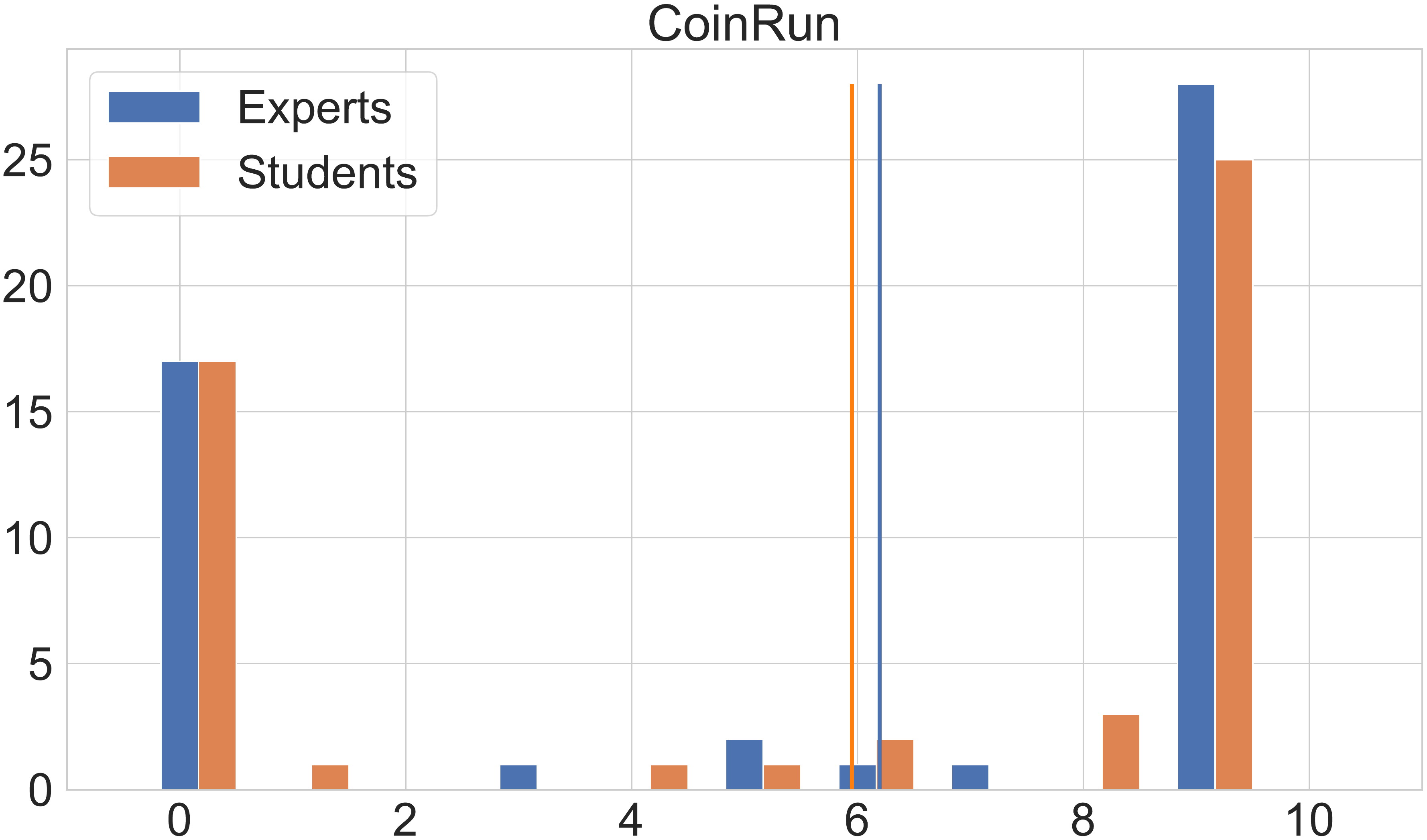}
    \includegraphics[width=0.32\textwidth]{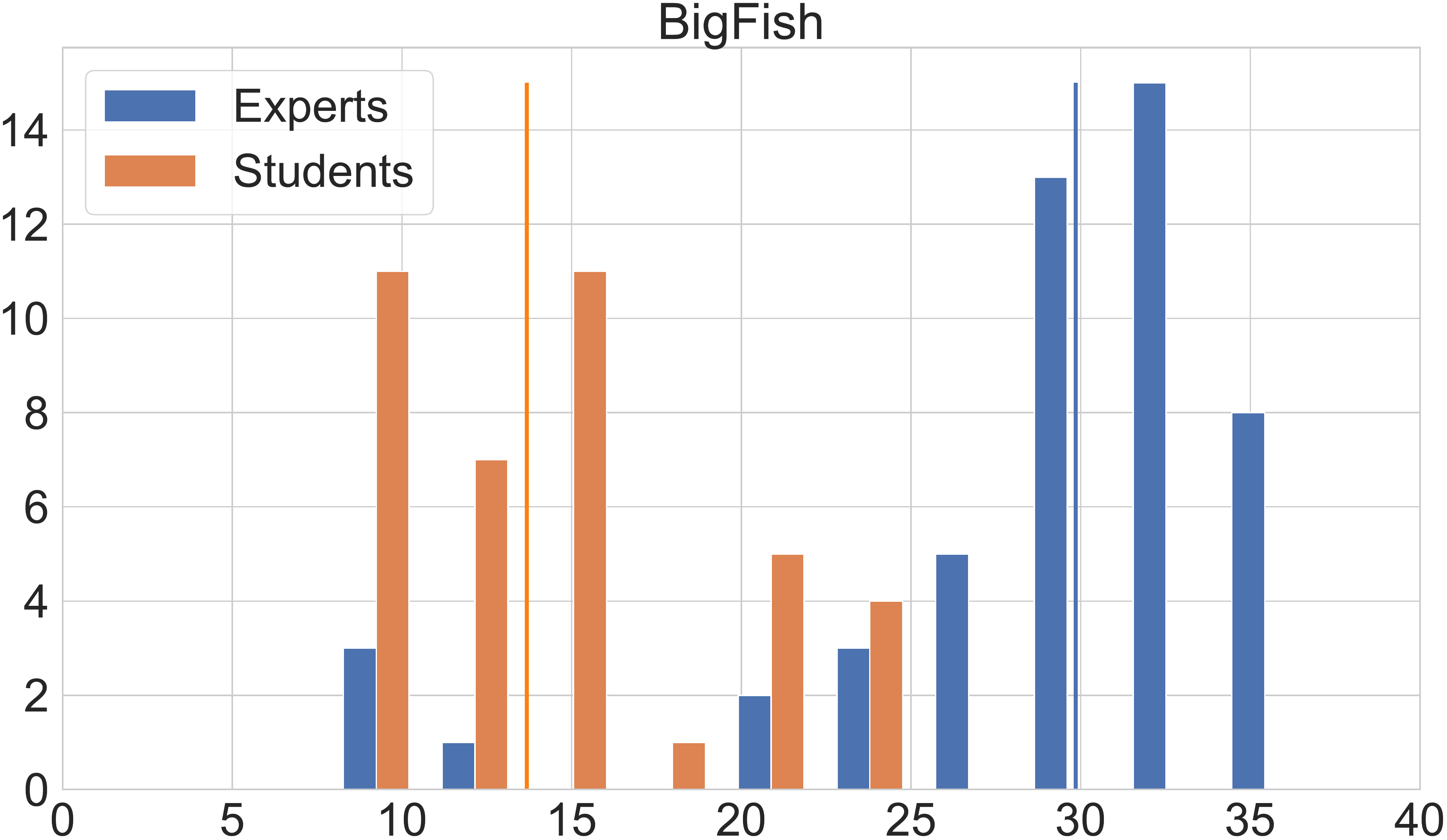}
    \includegraphics[width=0.32\textwidth]{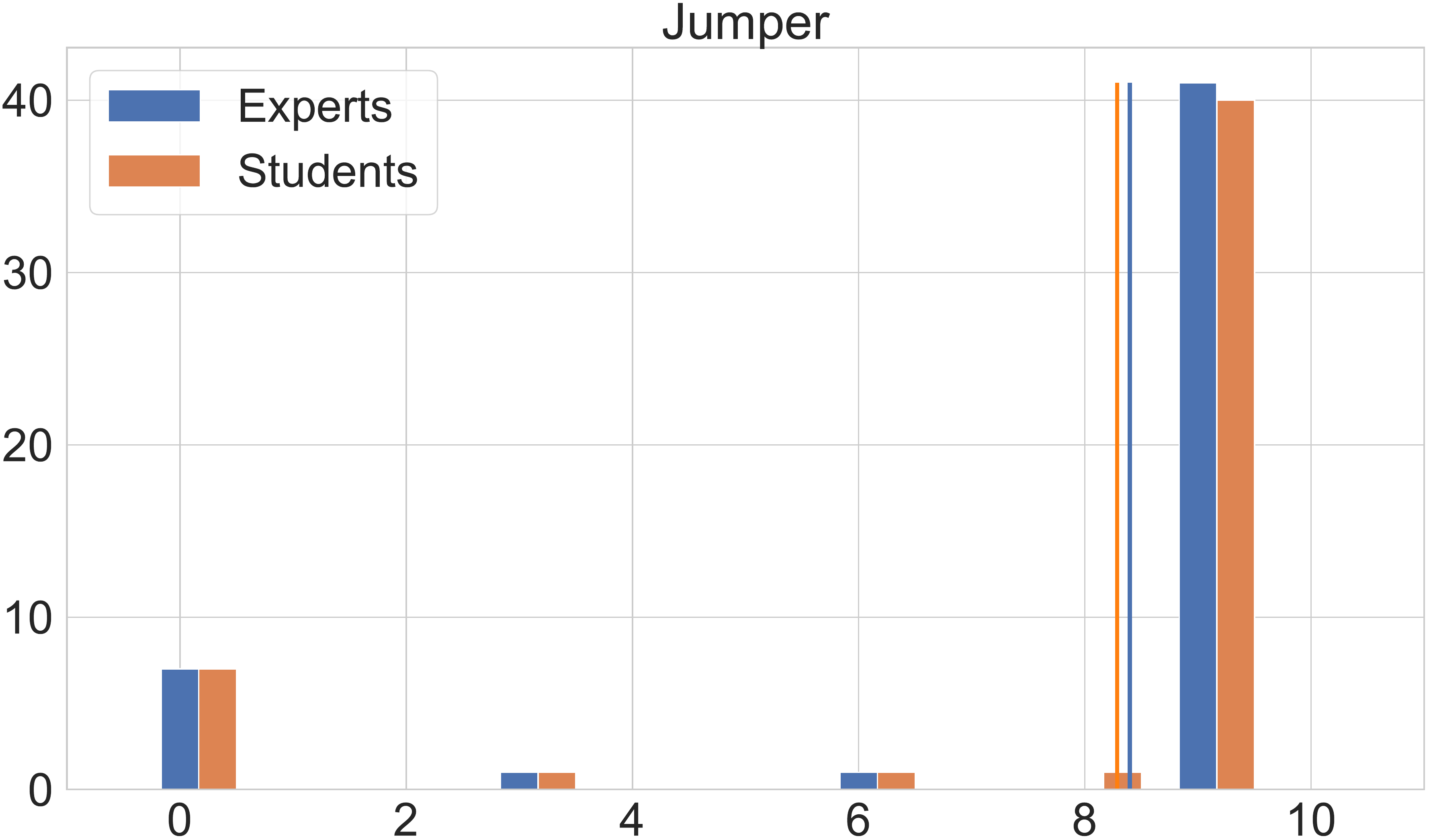}

    \includegraphics[width=0.32\textwidth]{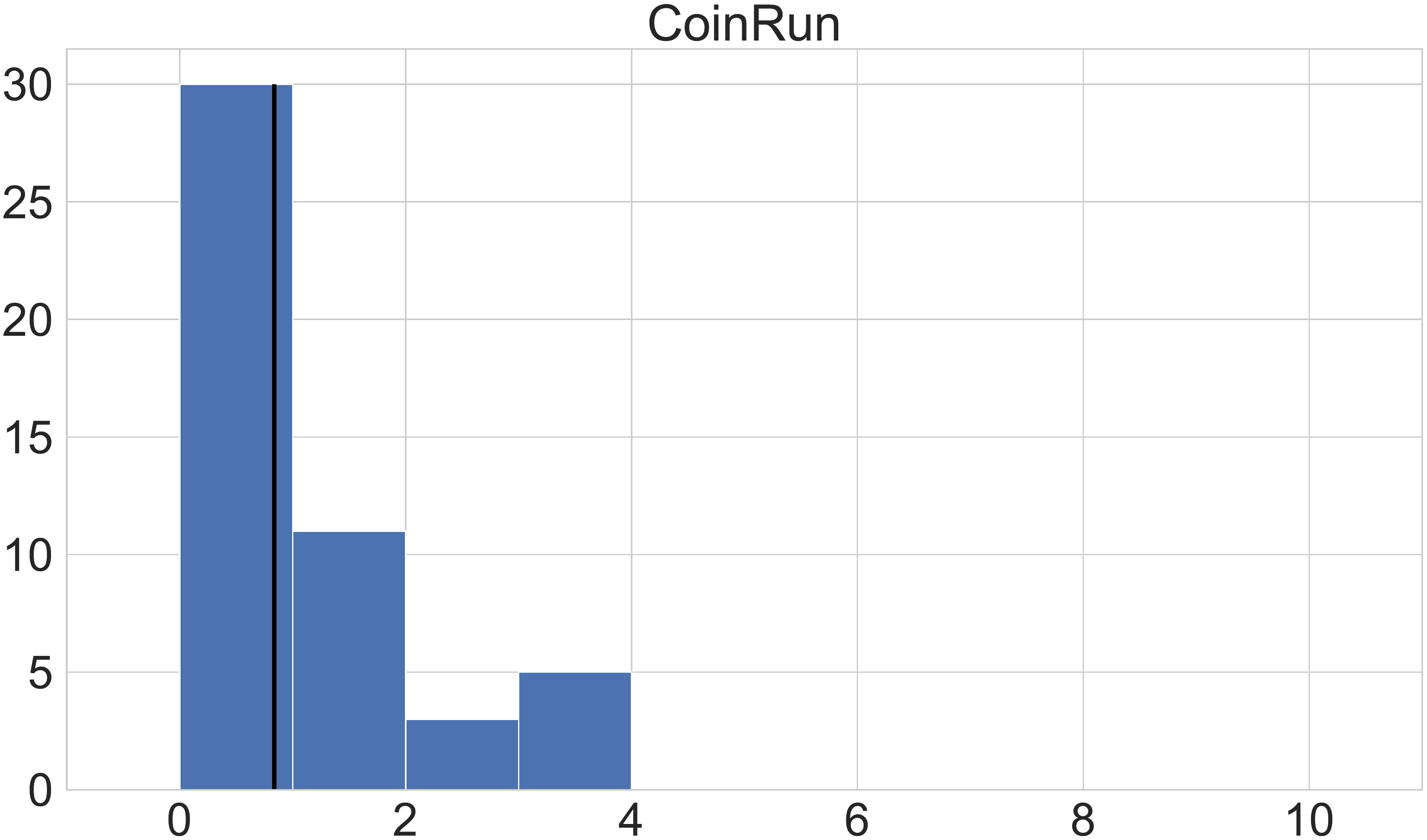}
    \includegraphics[width=0.32\textwidth]{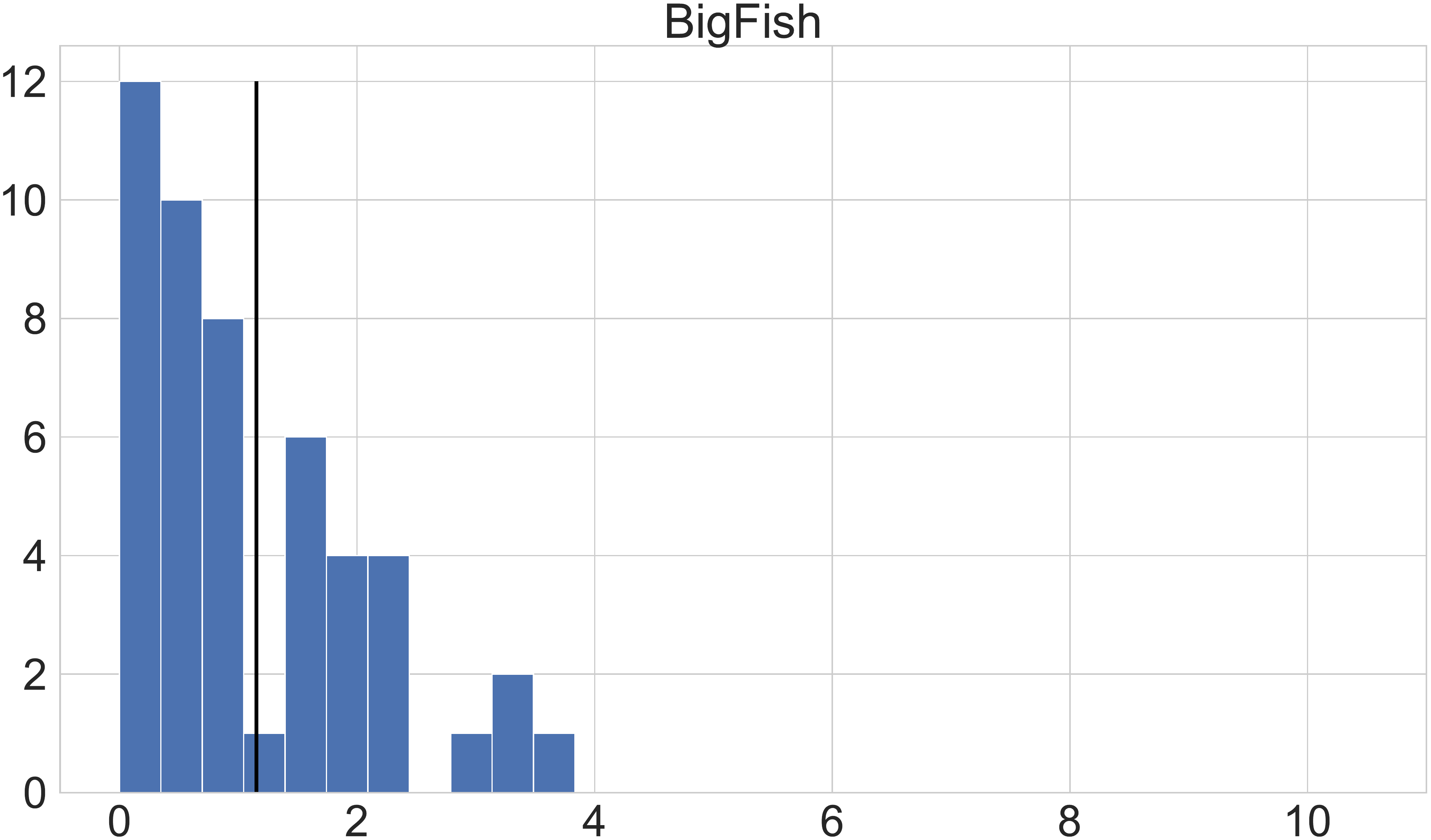}
    \includegraphics[width=0.32\textwidth]{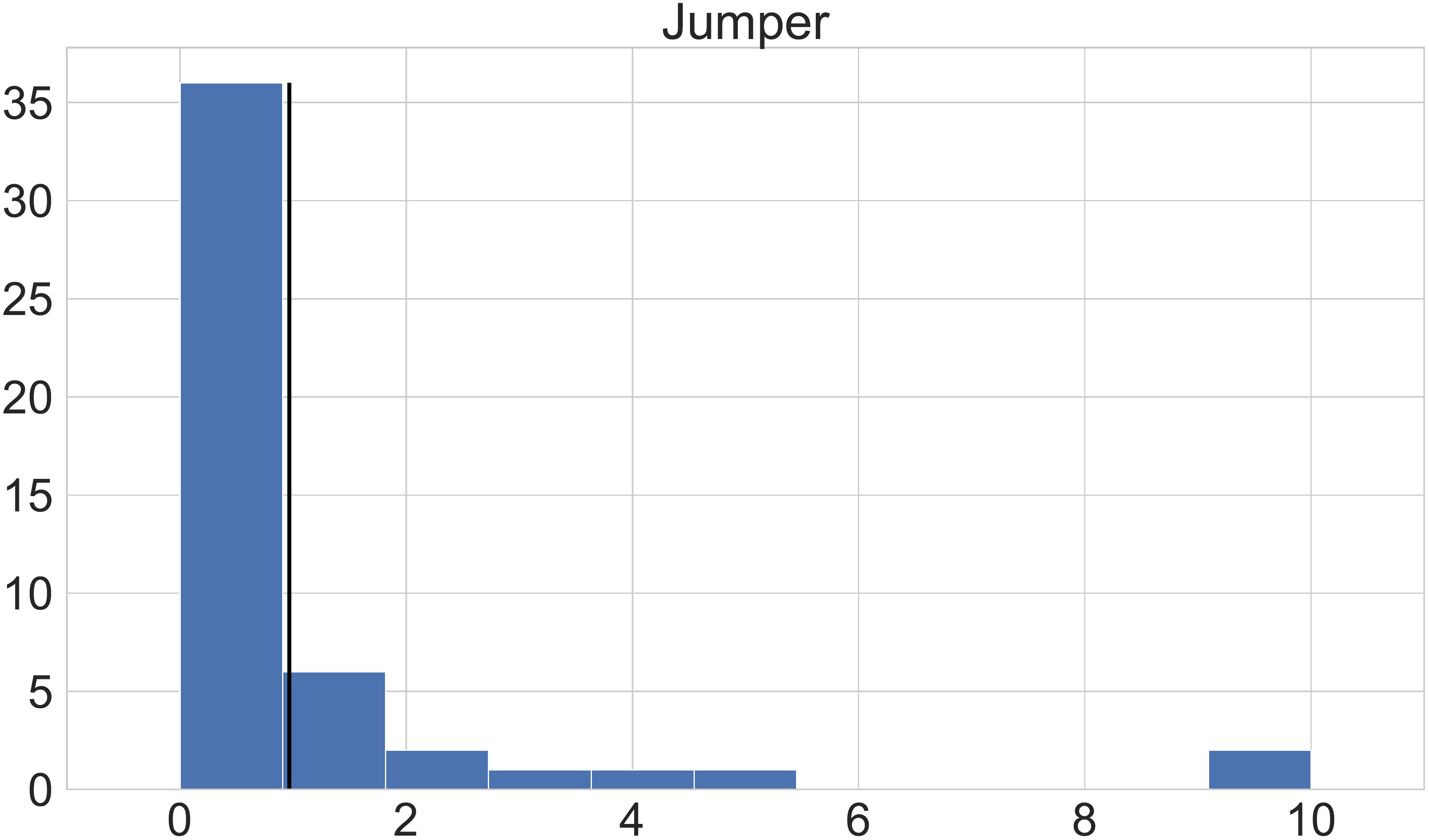}
    \caption{(Top) Training curves for the experts and the student on three Atari tasks. (Middle) Distribution of scores of trained experts on their respective training level and the student network over all 50 levels; mean is represented with a thin line. (Bottom) Distribution of scores of randomly selected trained experts on the 50 training levels (Note that BigFish x-axis is not the same between the middle and bottom rows).}
    \label{fig:visual_tasks_training}
\end{figure}

Figure \ref{fig:visual_tasks_training} reports training curves for the experts and the student on a subset of tasks (complete results in Appendix \ref{app:visual_tasks_training}).
Each training run was repeated 10 times for Atari games and 5 times for each level of each Procgen game.
One unit on the x-axis corresponds to 50k time steps for Atari games.
Procgen games display a large variance across runs so we report the score distribution of experts after 750000 time steps across all $N_{levels}=50$ training levels and 5 random seeds, for each game.
Average scores are represented with a thin vertical line.
Note that the training of both the experts and the student are sufficient to reach a non-trivial score but are also representative of reasonably small training times (details in Appendix \ref{app:reproducibility}) during which the experts do not necessarily reach state-of-the-art performance. Students do reproduce the performance of experts in almost all cases, with the exception of BigFish (discussed in Appendix \ref{app:bigfish}) and Breakout (shown in Appendix \ref{app:visual_tasks_training}).
As such, we find the allocated training time sufficient for common features to emerge within the state representation and to study distillation.

We note that, for the Procgen games of CoinRun and Jumper, the student is able to reproduce expert performance over the 50 levels.
While each expert is trained on a single level, the student is able to distill the information from each of these experts into a single network capable of performing similarly across the different levels.
We compare this to the generalization of randomly selected trained experts over the same 50 levels: although most experts are able to solve the level they are trained on, Figure \ref{fig:visual_tasks_training} confirms this is not enough to ensure transfer to any other level.
We posit that distillation is therefore beneficial for generalization to different levels in Procgen; further discuss robustness to level changes below.

We next evaluate the dependence of experts and students on various input features using saliency maps. Interpreting saliency maps from a human perspective can be a tricky endeavor; examples of such maps are reported and discussed in Appendix \ref{app:saliencies}.
Instead, we can evaluate the cardinality of the set of input features that strongly condition the network's output.
As mentioned in Section \ref{sec:pend_eval_crit}, it is desirable for this set to be small.
Contrarily to the toy experiment of Section \ref{sec:exp_pend} however, it is not necessarily always the same pixels in the saliency map that are important for the network's output and, on visual tasks, we mostly would like to filter out irrelevant background elements and only focus on level-independent visual cues.
In order to quantify this as a metric, we set all pixels in the saliency image to 1 if their intensities are above a certain threshold $\epsilon$, and 0 otherwise, then sum over all pixels and average this value across a predefined set of game images.
The intention is to reflect a similar criterion as that introduced in Section \ref{sec:pend_results}, under the form of a $\sum_i \sum_j \bm{1}( (\partial net / \partial s^i) (s_j) > \epsilon )$ value, where $net$ is either the expert or the student, $s_j$ is the $j$th game image in the testing set, and $s^i$ is the variable standing for the perturbation leading to the $i$th pixel in the saliency map.\footnote{Note that this is close to approximating an $\ell_0$ ``norm'' of the rounded gradient \citep{donoho2003optimally}.}
To account for the distribution of gradients independently of their amplitudes, we linearly scale the saliency maps pixel's intensity between 0 and 1 and take $\epsilon = 1/20$.

Table \ref{tab:saliencies} reports this metric for all games tested in the Atari and Procgen suites.
For the Atari games, the ``expert'' line reports the metric for each game's expert, while for the Procgen games, the reported value for a given game is the average over experts for all training levels.
There are far fewer important pixels for the students than for the experts on all environments, with some environments  showing a drastic reduction in the number of important variables; e.g. Carnival, which uses fewer than $1/7\times$ pixels in the student network.
It is worth noting that the performance is not impacted by the reduction in features, for example in Carnival, or in Pong, where the same performance is achieved by student networks which use fewer than half the pixels used by expert networks. This indicates that the distillation process leads the student's decision function to rely on specific visual features which are necessary for reproducing expert output. Distillation therefore seems to act as a regularizer among features by not creating dependence in the student network on unnecessary features.

\begin{table}
    \begin{adjustbox}{width=\columnwidth,center}
    \begin{tabular}{c||c|c|c|c|c||c|c|c}
         & Breakout & Pong & VideoPinball & Carnival & SpaceInvaders & CoinRun & BigFish & Jumper \\
         \hline
         expert & 239 (75) & 148 (81) & 331 (246) & 616 (278) & 2583 (772) & 1497 (203) & 727 (128) & 2628 (276) \\
         student & 70 (34) & 68 (50) & 173 (135) & 84 (108) & 1488 (784) & 1443 (193) & 460 (126) & 2116 (54) \\
    \end{tabular}
    \end{adjustbox}
    \caption{Saliency intensity (and standard deviation) as a proxy for the number of important variables.}
    \label{tab:saliencies}
\end{table}

\begin{figure}
    \centering
    \includegraphics[width=0.49\textwidth]{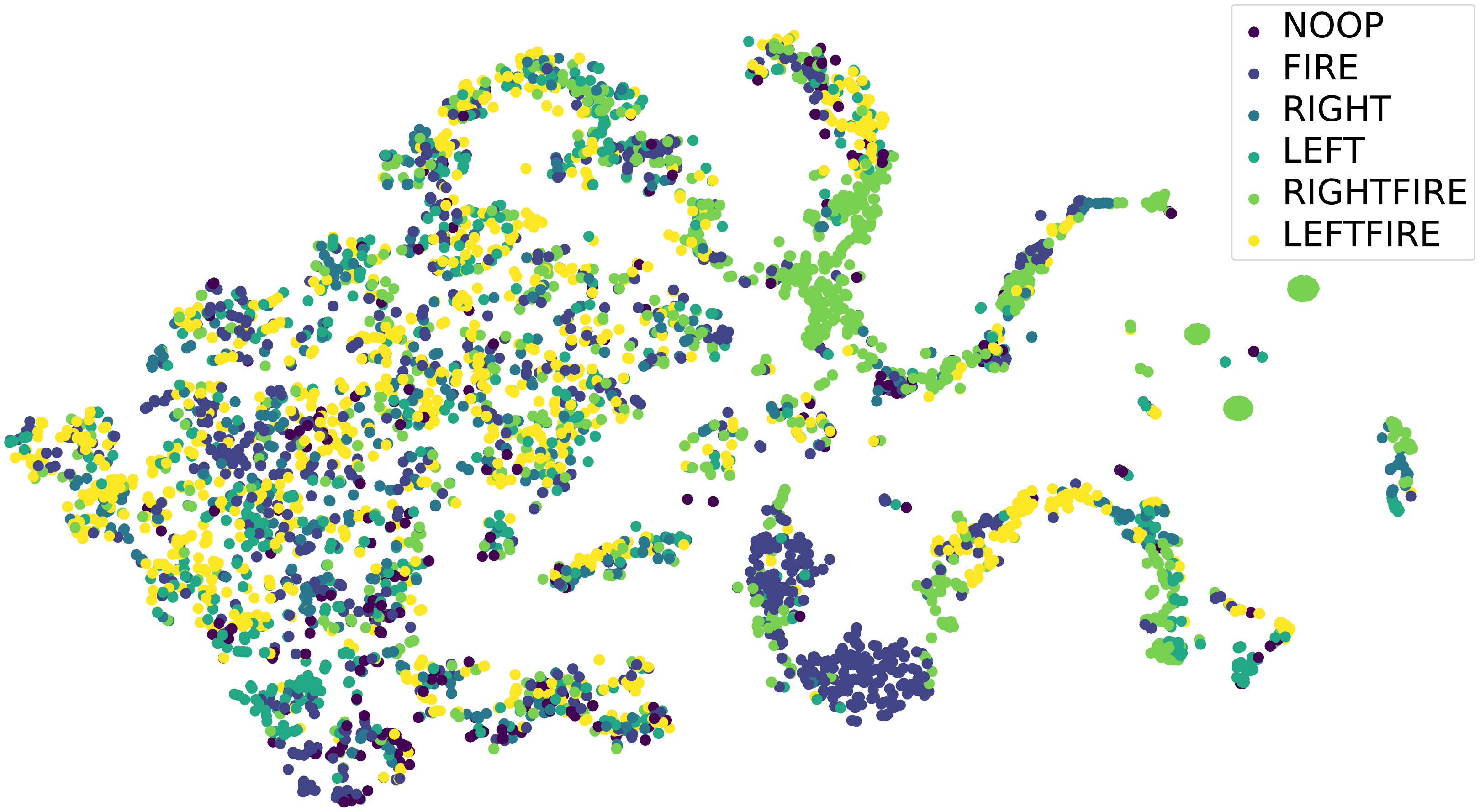}
    \includegraphics[width=0.49\textwidth]{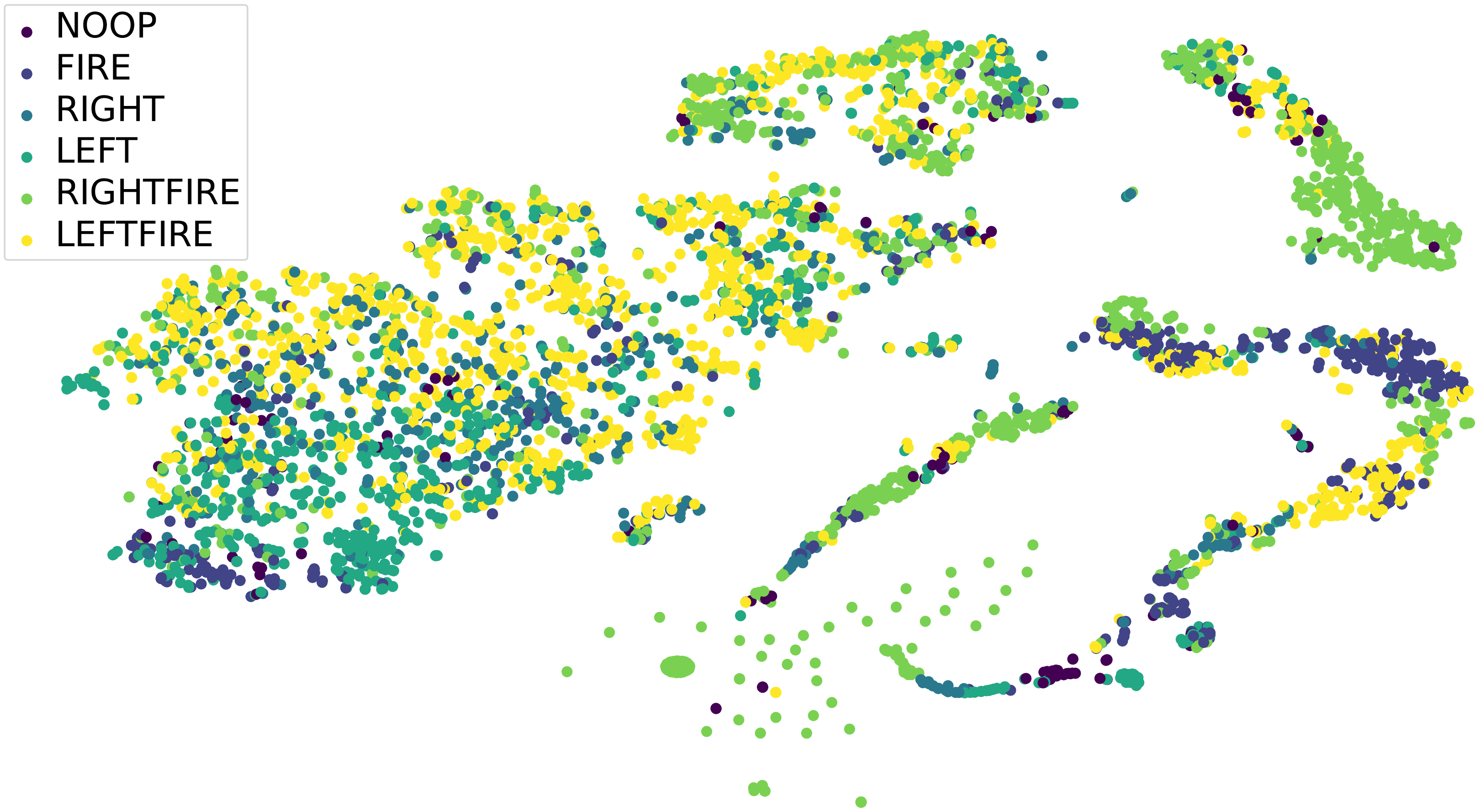}
    \vspace{1em}
    \includegraphics[width=0.49\textwidth]{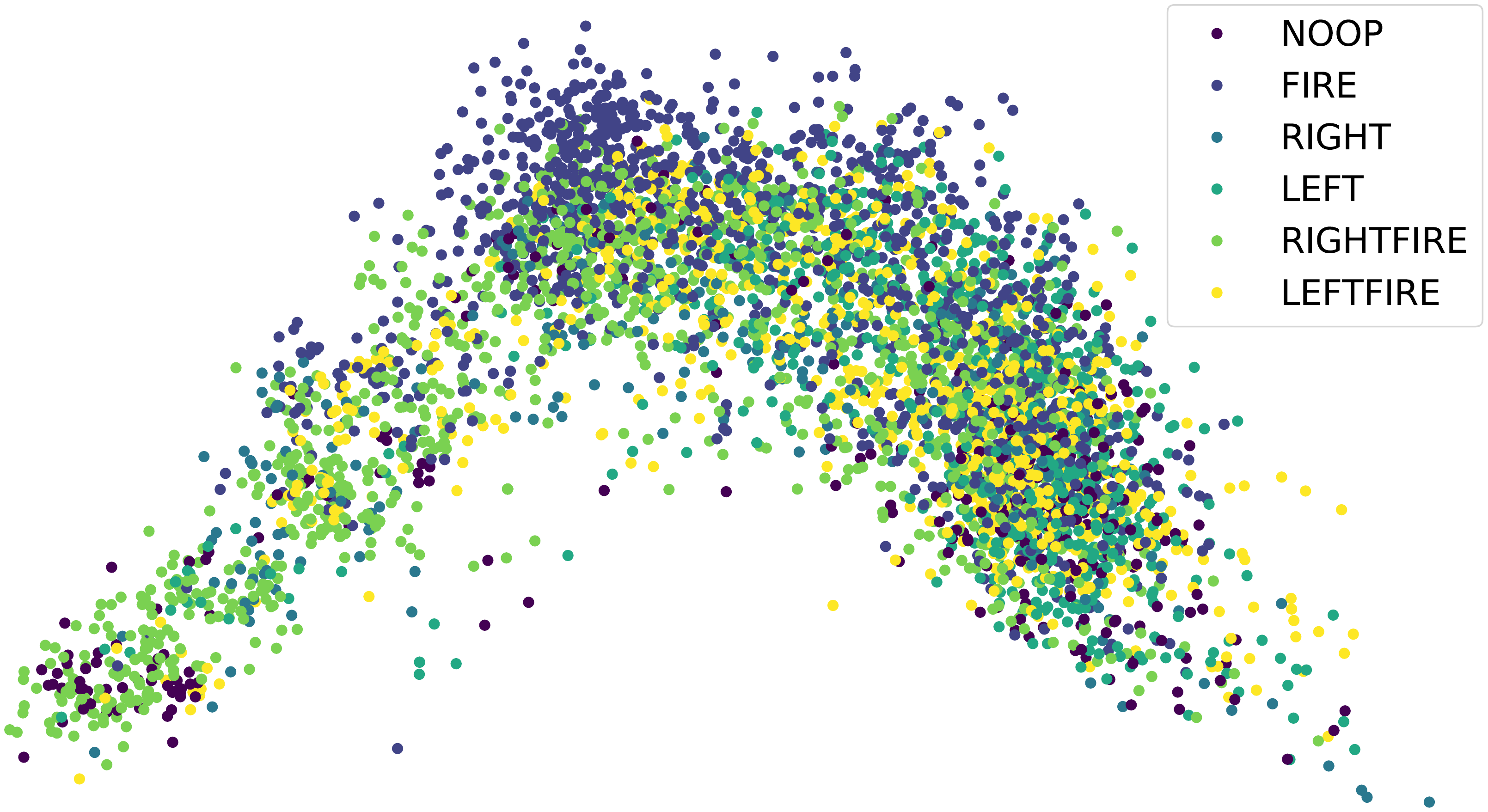}
    \includegraphics[width=0.49\textwidth]{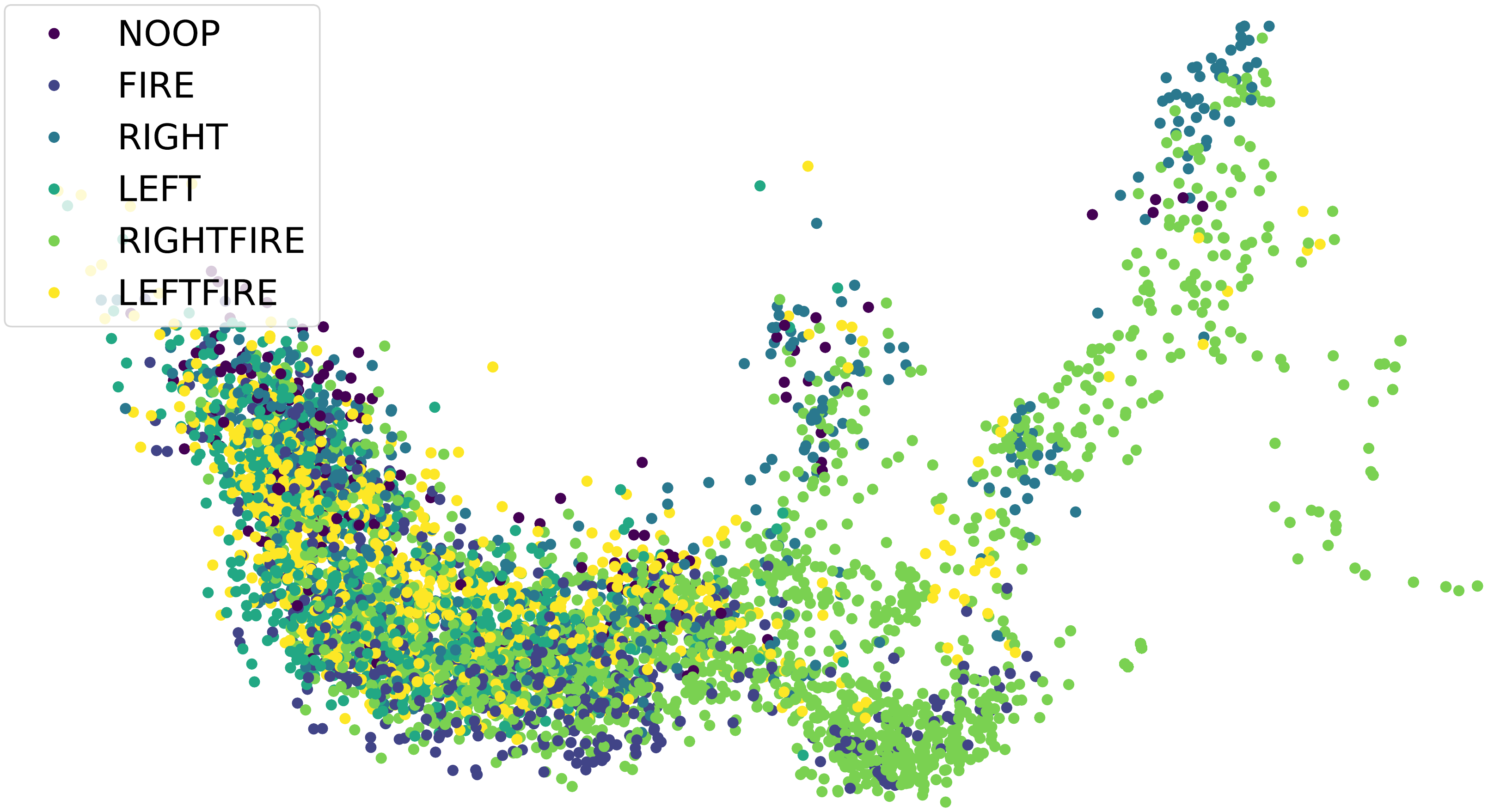}
    \caption{t-SNE (top) and PLS (bottom) state embeddings for an expert (left) and the student (right) on Space Invaders.}
    \label{fig:visual_tasks_tsne}
    \vspace{-1em}
\end{figure}

Figure \ref{fig:visual_tasks_tsne} presents the t-SNE and PLS representations of a set of images from the Space Invaders game, colored by their associated optimal action (similar representations for other games in Appendix \ref{app:sep_visual}). We note that features are more clearly grouped in the student networks based on action in both the t-SNE and PLS representations, especially for the RIGHTFIRE action seed in the PLS representation. We posit that distillation concentrates features based on final output, leading to more separable projected states.
To deepen this analysis, we repeat the same quantification of separability of projected states as in Section \ref{sec:pend_results}, by training a linear SVM on projected states.
Table \ref{tab:visual_tasks_sep} illustrates how the state representation learned by the student lends itself better to a linear classification, which indicates better geometrical proximity for states sharing the same action. This indicates that distillation creates networks with a better chance at avoiding overfitting and generalizing to visual changes as their feature representation groups states based on correlated action and not on visual features that have low correlation with the final output.

\begin{table}
    \begin{adjustbox}{width=\columnwidth,center}
    \begin{tabular}{c||c|c|c|c|c||c|c|c}
         & Breakout & Pong & VideoPinball & Carnival & SpaceInvaders & CoinRun & BigFish & Jumper \\
         \hline
         expert & 77.00 (1.25) & 75.29 (1.90) & 71.23 (3.58) & 70.07 (1.13) & 76.49 (2.37) & 92.53 (0.41) & 95.49 (2.99) & 76.10 (3.43) \\
         student & 79.03 (3.50) & 80.99 (1.78) & 80.83 (8.43) & 72.36 (2.15) & 95.47 (1.98) & 97.19 (0.30) & 97.96 (2.01) & 81.15 (2.71) \\
    \end{tabular}
    \end{adjustbox}

    \caption{Average (and standard deviation) of accuracy for a linear classifier built on the state representation.}
    \vspace{-1em}
    \label{tab:visual_tasks_sep}
\end{table}

\begin{figure}
    \centering
    \includegraphics[width=0.32\textwidth]{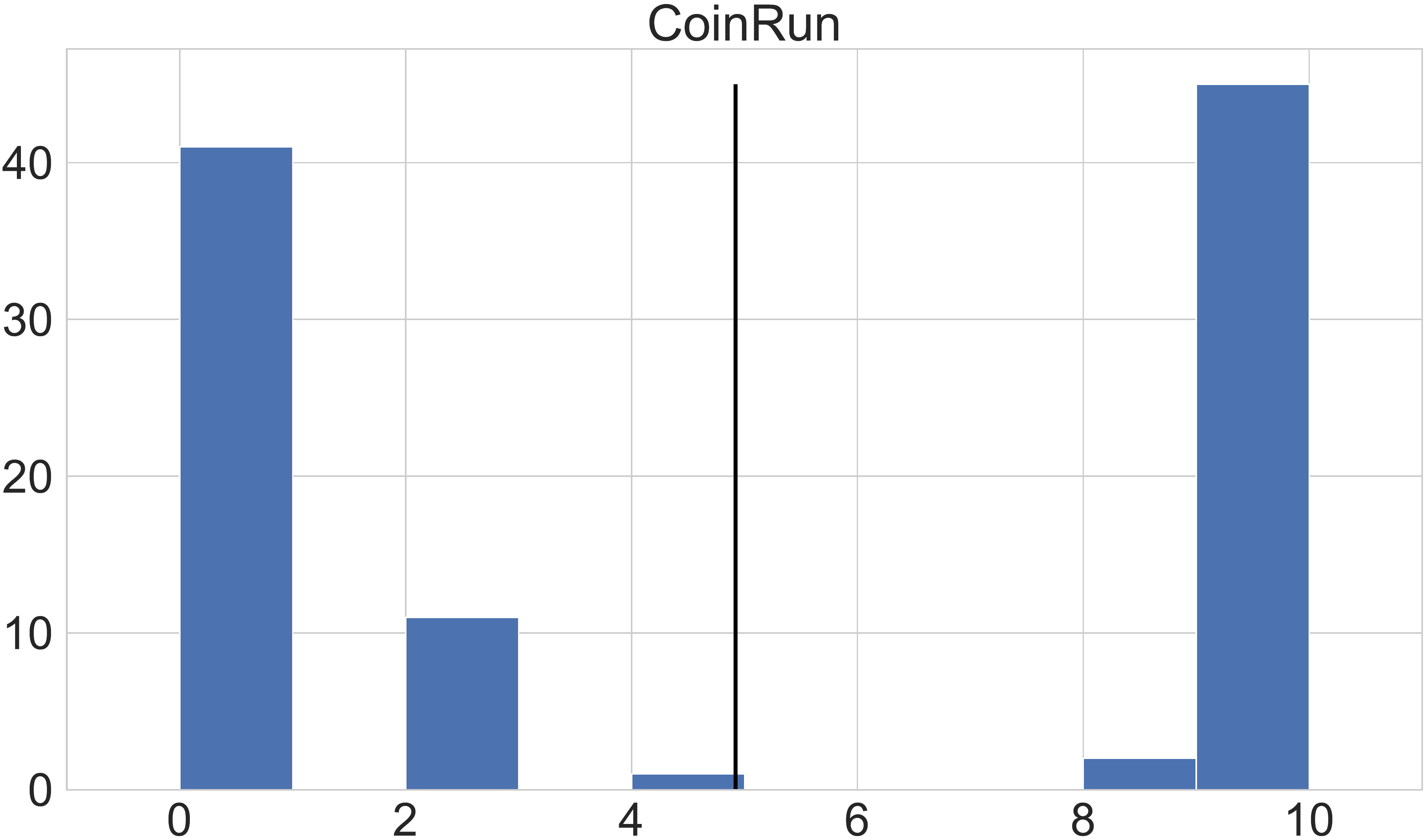}
    \includegraphics[width=0.32\textwidth]{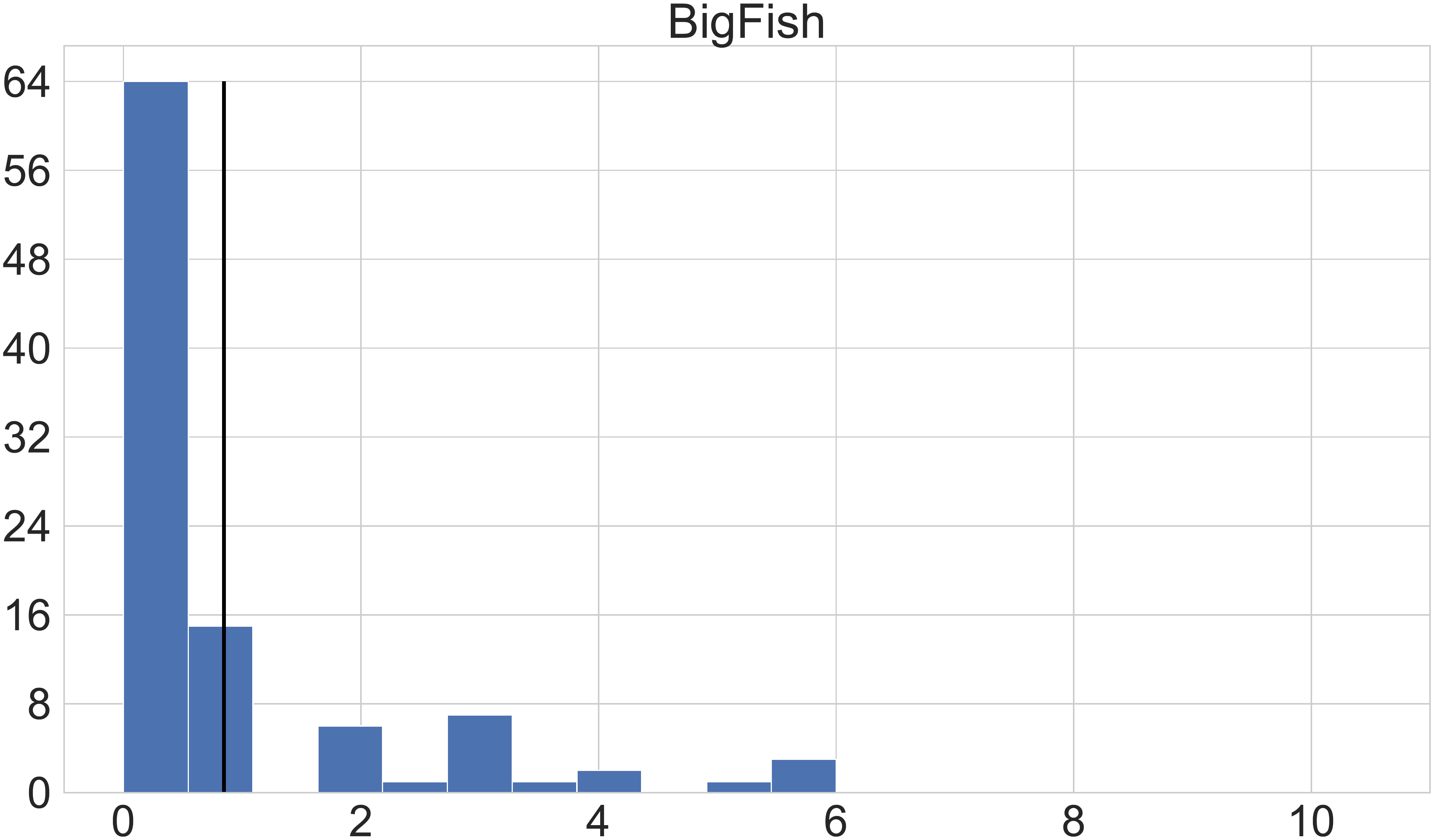}
    \includegraphics[width=0.32\textwidth]{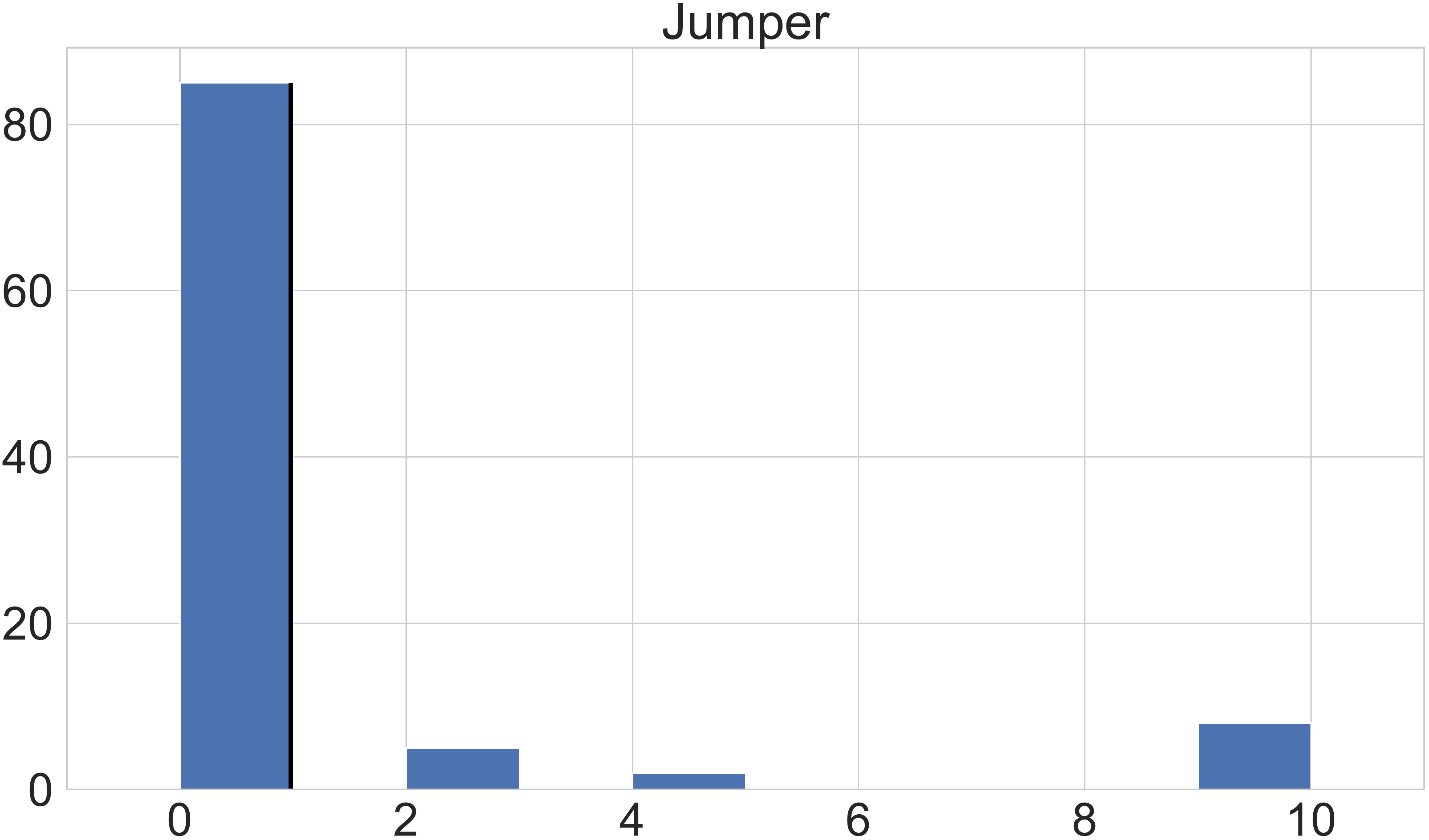}
    \caption{Robustness evaluation of the student's policy on the testing distribution of Procgen games.}
    \label{fig:procgen_robust}
\end{figure}

To understand the importance of a good state representation in adapting to new environments, we evaluate the student's robustness to new levels in the selected Procgen games. We discuss the possibility of using distillation for the Atari benchmark in \ref{app:atari_annex}, but focus here on Procgen.
Figure \ref{fig:procgen_robust} reports the distribution of scores when the student is evaluated on levels from the testing distribution of Procgen, i.e. levels with different layouts, visual representation, and dynamics such as enemy number and placement.

In the case of the BigFish and Jumper environments, no simple policy exists common to every level that allows for a good performance, thus almost no agent is able to succeed without retraining on new levels. We note, however, that the small number of levels (50) used to train the experts contributes to the lack of generalization; in  \citep{cobbe2020leveraging}, individual networks trained on 100, 300, and even 1000 levels had similar performance on test levels for BigFish and Jumper.
However, the performance of the student network on CoinRun is interesting as the simple policy of running right while jumping as soon as possible is often enough to finish many levels.
In fact, Figure \ref{fig:procgen_robust} shows that the student is able to solve more than half of the 100 unseen levels we tested it on.
On the other hand, the fail cases are in a large majority due to the presence of enemies in the path of the agent when following this simple policy.
This confirms that, although simple, the policy extracted by the distillation process is quite general and can transfer as is to multiple new levels.

\section{Conclusion}
\label{sec:conclu}

In this paper we explore the impact of policy imitation via distillation on state representation and showed that distillation can act as an information bottleneck that mitigates observational overfitting.
More specifically, we isolated three criteria to evaluate the ability of a policy to generalize to various environments and proposed experimental means of measuring each one.
Our experiments on a toy environment confirmed that distillation is an efficient way to filter out task-specific features, thus allowing trained policies to focus on variables in the observation space that are relevant for multiple environments.
The extension to more difficult visual tasks supports this idea and reinforces the use of multi-task policy imitation as an interesting mechanism for general feature extraction, including when the training of the experts or of the student are interrupted before total convergence.
One notable finding from this work is that distillation requires a limited number of pre-trained experts, and no specific knowledge of task boundaries or identifiers, to achieve efficient generalization.
The ability to learn a good state representation from a set of various tasks is a crucial skill for any agents evolving in multiple environments.
In the lifelong learning setting in particular, an efficient encoding can allow for better transfer and generalization between tasks.
An additional interesting perspective lies in the combination of self-learning of good representations that aim at observational robustness within a given task \citep{grill2020bootstrap,bertoin2022local}, with inter-task distillation as presented here.

\subsubsection*{Acknowledgements}
    This work was granted access to the HPC resources of the CALMIP supercomputing center under allocation p21001.
    The authors acknowledge the support of the French Defence Innovation Agency (AID) under grant ARAC.


\appendix

\section{Reproducibility}
\label{app:reproducibility}

All the code used in this paper can be found at \url{https://anonymous.4open.science/r/DistillationAsBottleneck/}.
The code is based on the Dopamine implementation of standard RL algorithms which can be found at \url{https://github.com/google/dopamine}.

\begin{table}[H]
    \centering
    \begin{tabular}{c||c|c|c}
         & Toy Pendulum & Procgen & Atari \\
         \hline
         algorithm & C51 & Rainbow & Rainbow \\
         feature size & 32 & 512 & 512 \\
         optimizer &  \multicolumn{3}{c}{Adam} \\
         learning rate & 5e-3 & 6.25e-5 & 6.25e-5 \\
         memory capacity & 50000 & 250000 & 1000000 \\
         batch size & 128 & 32 & 32 \\
         target update period & 100 & 4000 & 8000 \\
    \end{tabular}

    \caption{Algorithm hyperparameters.}
    \label{tab:hyperparameters}
\end{table}

The hyperparameters used for our algorithm on each environment can be found in table \ref{tab:hyperparameters}.

We also used the scikit-learn \citep{pedregosa2011scikit} implementation of the PLS, t-SNE and SVM algorithms with their default parameters.
For the saliency map computation, we re-implemented the perturbation-based method proposed by \citet{greydanus2018visualizing}.
This technique consists in measuring the saliency of a given pixel by computing the difference in the network outputs when fed with a sampled state or with a blurred version of that same state (using a Gaussian blur).

Every network training for this paper was realised on GPUs, mainly on Nvidia V100s cards.
For the toy environment of section \ref{sec:distillation}, we trained each expert for 50 iterations of 1000 time steps which took around 15 minutes per expert, and each student for 25 iterations of 1000 time steps which took around 1h30.
For each Procgen Game, each expert was trained for 30 iterations of 25000 time steps for a computation time of around 1h20, and each student for 8 iterations of 25000 time steps during around 12h.
Finally we trained each expert on any Atari games for 60 iterations of 50000 time steps (around 6h of computation) and each student for 15 iterations of 50000 time steps (around 2 days).

\section{Separability of state embeddings on the controlled experiment of Section \ref{sec:distillation}}
\label{app:pend_accuracy}

A more detailed version of Table \ref{tab:pend_pls} is given in Table \ref{tab:pend_pls_extended}.
For each of the 24 levels, we train an expert, compute the PLS features, keep either the 3 main ones or all 32, and train the linear SVM on these features to imitate the expert's policy.
We compute the classification accuracy, and then average the results across experts.
We repeat this training 20 times and average across these 20 repetitions.
This computation is done once and for all and thus we report the same values along the ``expert'' line, whatever the number of levels used for distillation.
These are the same experts that are then used for distillation.
For the ``students'' line however, only experts from levels 1 and 2 are used in the first column, then levels 1 up to 6 for the second and so on.
The reported value is thus the average over the 20 repetitions.

\begin{table}[H]
    \centering
    \begin{tabular}{c|cc|cc|cc}
         & \multicolumn{2}{c|}{$N_{levels}=2$}  & \multicolumn{2}{c|}{$N_{levels}=3$} & \multicolumn{2}{c}{$N_{levels}=4$}  \\
         & dim = 3 & dim = 32       & dim = 3 & dim = 32        & dim = 3 & dim = 32\\
         \hline
        expert & 78.05 (4.90) & 91.29 (2.06) & 78.05 (4.90) & 91.29 (2.06) & 78.05 (4.90) & 91.29 (2.06) \\
        \hline
        student & 75.56 (6.27) & 91.60 (2.16) & 80.65 (4.87) & 93.60 (1.22) & 81.33 (4.67) & 93.43 (1.87)
    \end{tabular}\\
    \vspace{1em}
    \begin{tabular}{c|cc|cc|cc}
         & \multicolumn{2}{c|}{$N_{levels}=6$} & \multicolumn{2}{c|}{$N_{levels}=12$}    & \multicolumn{2}{c}{$N_{levels}=24$} \\
         & dim = 3 & dim = 32       & dim = 3 & dim = 32        & dim = 3 & dim = 32\\
         \hline
        expert & 78.05 (4.90) & 91.29 (2.06) & 78.05 (4.90) & 91.29 (2.06) & 78.05 (4.90) & 91.29 (2.06) \\
        \hline
        student & 83.06 (5.66) & 93.68 (1.87) & 88.03 (1.56) & 95.37 (0.75) & 91.01 (4.60) & 95.81 (1.78)
    \end{tabular}

    \caption{Average (and standard deviation) of accuracy for a linear classifier built on the state representation.}
    \label{tab:pend_pls_extended}
\end{table}

\section{Additional details on the Atari benchmark}
\label{app:atari_annex}

In this section, we motivate decisions made concerning the Atari benchmark, notably stopping training early and the lack of a robustness study.

For Atari games, we stopped training before convergence for experts and students. For instance, the expert had not converged after 60 iterations in Breakout, and 15 iterations is not enough for the student to match the expert performance on Breakout.
This was motivated by different factors:
\begin{itemize}
    \item the convergence is not guaranteed as the task of imitating 5 experts at the same task is far from trivial,
    \item the distillation process aims at extracting general features and we argue this can be done before convergence of the policy,
    \item the computation time would have been intractable in order to run 10 seeds for each experiment.
\end{itemize}
For comparison, our training time was around 6h for 60 iterations on a single seed; Dopamine \citep{castro2018dopamine} measured iterations of 250,000 timesteps ($5\times$ ours) and trained for 200 iterations, in other words, roughly 16 times more timesteps than our experiments. We determined that running each seed for 100 GPU hours was unnecessary for studying distillation.

In section \ref{sec:visual_results}, we choose to focus on Procgen for evaluating the robustness of student networks. We argue that evaluating the robustness of the student network on Atari games is not as relevant for this work.
We already know from the distillation process that the student is efficient on different instances of the same game used to train the experts (Figure \ref{fig:visual_tasks_training}).
However, as the underlying dynamics of two Atari games are very different from each other, we would not expect the student's policy to perform well on a new game based solely on a good state representation.
An open question would be to evaluate how the learned state representations could be used to initialize a new expert and how the training would compare to an uninformed initialization.
This would amount to using distillation as a means to perform meta-learning as in the work of \citet{finn2017model-agnostic,nichol2018first-order}.
Following the recent contribution of \citet{schwarzer2021pretraining} we argue that the question of transferring state representations as good initializations for a neural network is a separate topic and reserve it for future work.

\clearpage

\section{Distillation pseudo-code}
\label{app:pseudocode}

Algorithm \ref{alg:distillation} summarizes the pseudo code corresponding to the algorithmic operations of Sections \ref{sec:distillation} and \ref{sec:visual}.

\begin{algorithm}
    \caption{Expert training then distillation}
    \label{alg:distillation}
    \begin{algorithmic}[1]
        \For{$i \in [1,N_{levels}]$}
	        \Comment Train each expert separately
            \State Draw MDP $\mathcal{M}$ and default initialization of $\pi_E$
            \State Train $\pi_E$ on $\mathcal{M}$ for $K$ episodes
            \State Store $(\pi_E,\mathcal{M})$ in the list $\mathcal{E}$ of experts
        \EndFor
        \State Default initialization of $\pi_{student}$
        \Repeat
            \Comment Train student
            \For{$\pi_E,\mathcal{M} \in \mathcal{E}$}
                \For{all time steps $t$ in one episode}
                    \State Add sample $(s_t,\pi_E(a|s_t))$ in a training set $D_\mathcal{M}$ by playing $\pi_{student}$ in $\mathcal{M}$
                    \State Sample a batch of states $\mathbf{s} \sim D_\mathcal{M}$
	                \State Take one gradient step on $\pi_{student}$ to minimize imitation loss $L$
	            \EndFor
            \EndFor
        \Until{maximum number of episodes reached}
        \State
        \State For plain distillation (Section \ref{sec:distillation}): $L = \sum_\mathbf{s} [ \pi_E(a | s) \log \pi_{student}(a | s) ]$
        \State For AMN on visual tasks (Section \ref{sec:visual}): $L = \sum_\mathbf{s} [ \pi_E(a | s) \log \pi_{student}(a | s) ] + \beta \lVert \phi_E(s) - f(\phi_{student}(s)) \rVert_2^2$,
        \State \hspace{1em} with $f$ the adaptation layer and $\phi$ the feature layer of each network (see Section \ref{sec:visual_benchmarks} and \citep{parisotto2016actor-mimic}).
    \end{algorithmic}
\end{algorithm}

\section{Discussion on the results on the BigFish benchmark}
\label{app:bigfish}

The results of the student agent on the BigFish game might appear mitigated and deserve a separate discussion.
One first remark is that the student does not do so badly: experts trained on a level and evaluated on another have an average score around 2 (figure \ref{fig:visual_tasks_training}, bottom center).
The student network has an average score of around 14.
This is well below the average score of experts evaluated solely on their expertise level (30), but still better than naive transfer.
Possible immediate causes for such a mitigated result are (a) insufficiently long distillation, (b) insufficient variety of training levels or (c) convergence to a policy that does not generalize enough.
Reason (a) is unlikely since various lengths were tested for distillation.
It should be noted that the standard protocol proposed by \citet{cobbe2020leveraging} uses between 100 and 100000 training levels, and states that at least 10000 levels are necessary to close the generalization gap for single agent training.
Recent works that use this benchmark (e.g. RAD \citep{laskin2020reinforcement}, IDAAC \citep{raileanu2021decoupling} or CLOP \citep{bertoin2022local}) typically use 500 levels for training, which makes (b) a plausible explanation.
This hypothesis is also supported by the fact that even though the student has performance 14 on the training levels, it does not generalize at all on the testing ones (Figure \ref{fig:procgen_robust}).
Finally, it is known that SGD acts as an implicit regularizer \citep{gunasekar2017implicit} that leads the optimization process towards a single minimum even with few samples, but this minimum has no guarantee of being generalizable across tasks, as studied by \citet{song2020observational}, hence hypothesis (c).

\section{Training curves on visual control tasks}
\label{app:visual_tasks_training}

\begin{figure}[H]
    \centering
    \includegraphics[width=0.49\textwidth]{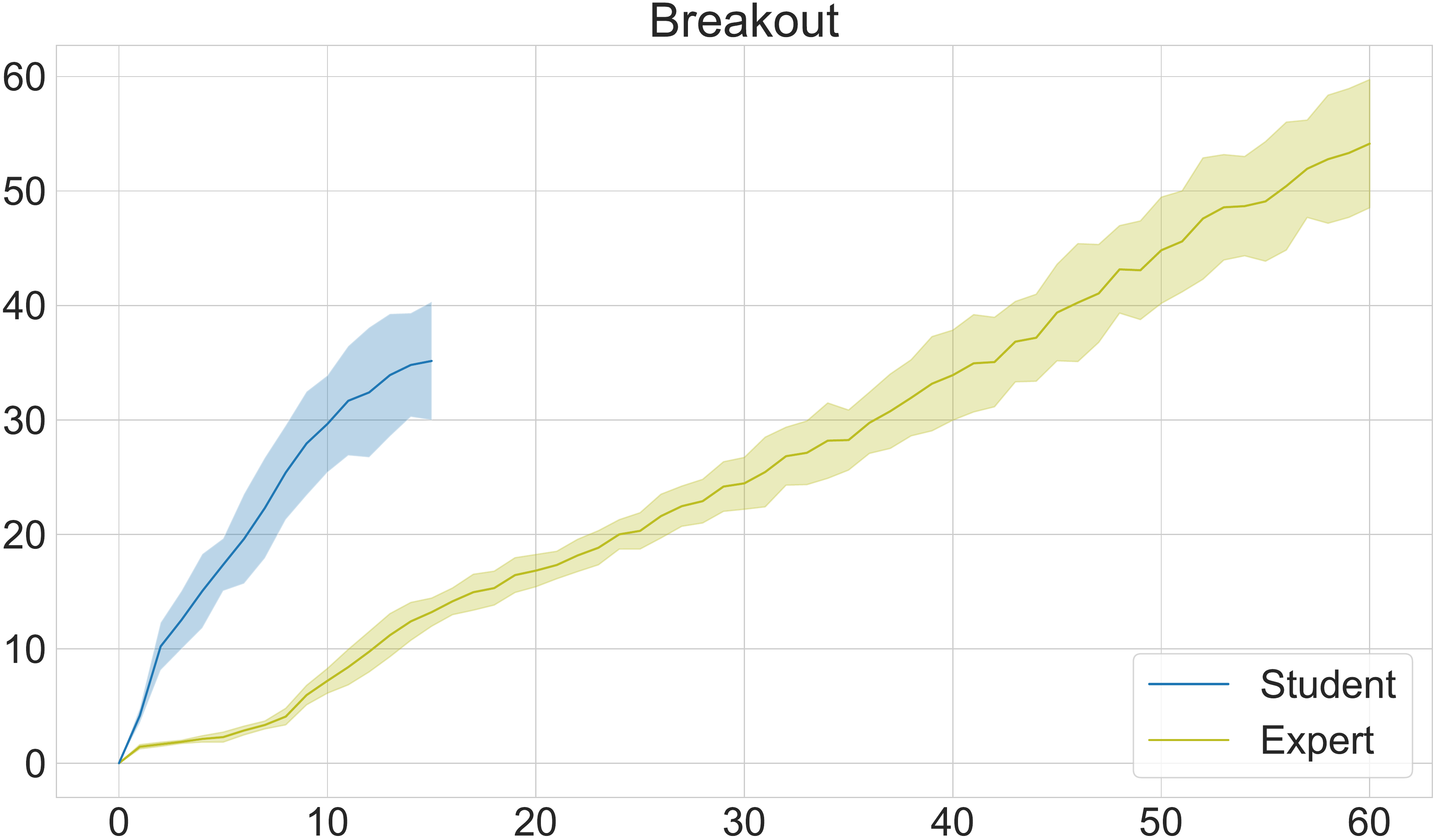}
    \includegraphics[width=0.49\textwidth]{figs/Atari/Perf_Carnival_grid.pdf}
    \includegraphics[width=0.49\textwidth]{figs/Atari/Perf_Pong_grid.pdf}
    \includegraphics[width=0.49\textwidth]{figs/Atari/Perf_SpaceInvaders_grid.pdf}
    \includegraphics[width=0.49\textwidth]{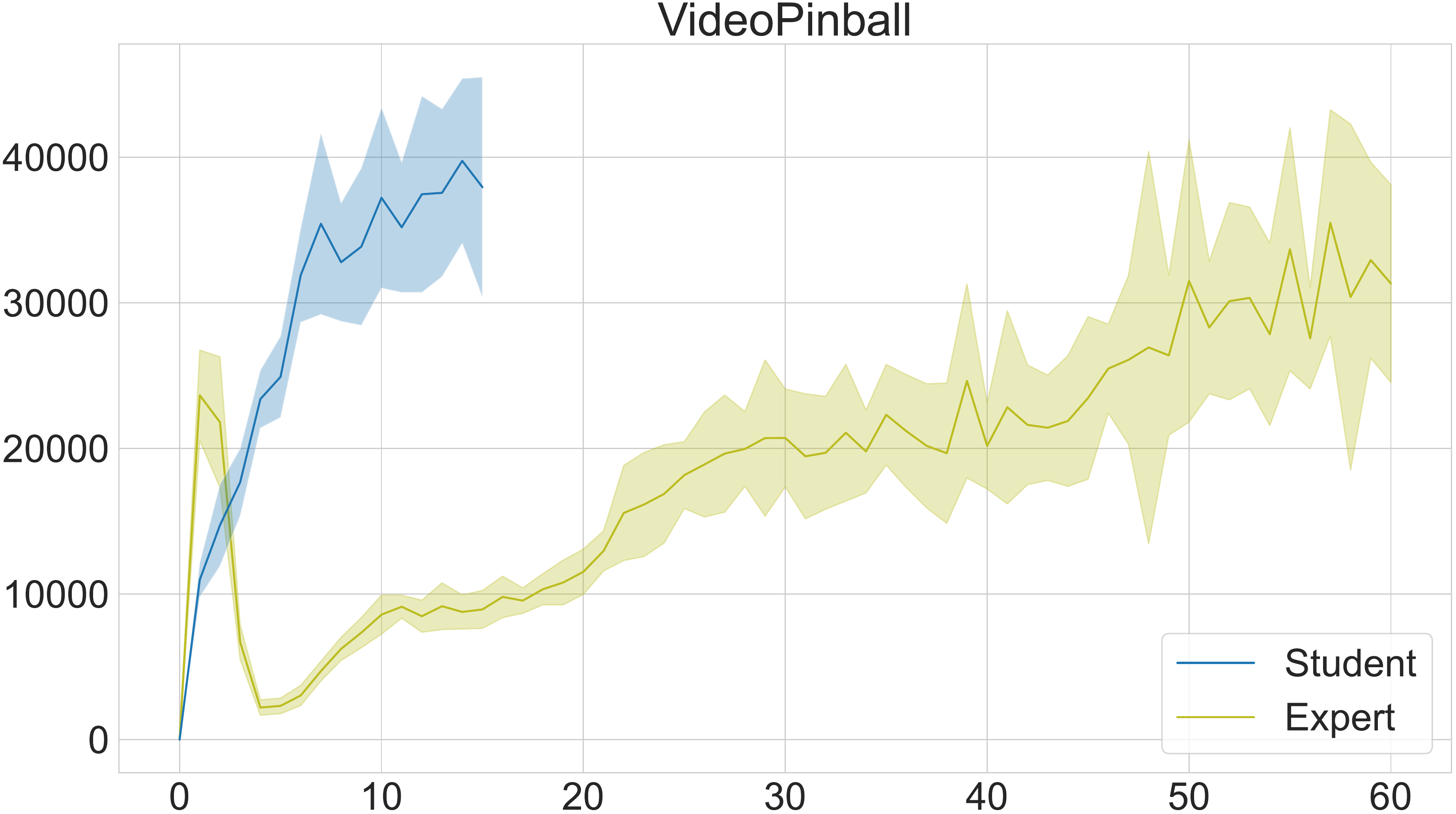}
    \caption{Training curves for the experts and the student on visual control tasks.}
\end{figure}

\clearpage
\section{Saliency maps}
\label{app:saliencies}

\begin{figure}[H]
    \centering
    \includegraphics[width=0.32\textwidth]{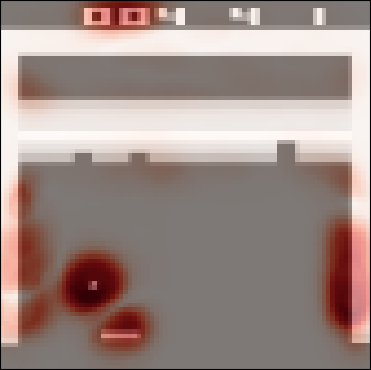}
    \includegraphics[width=0.32\textwidth]{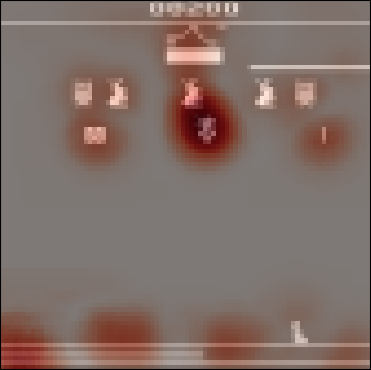}
    \includegraphics[width=0.32\textwidth]{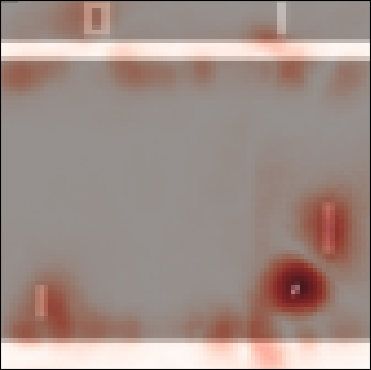} \\
    \vspace{1em}
    \includegraphics[width=0.32\textwidth]{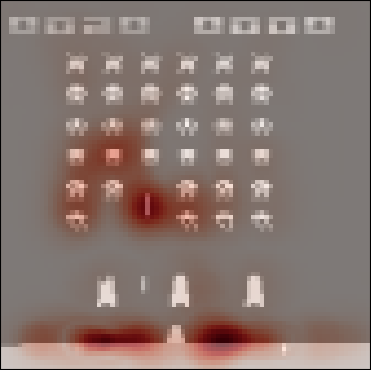}
    \includegraphics[width=0.32\textwidth]{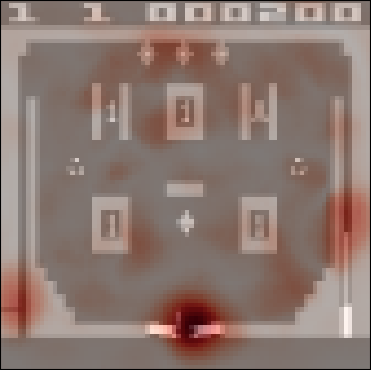} \\
    \vspace{1em}
    \includegraphics[width=0.32\textwidth]{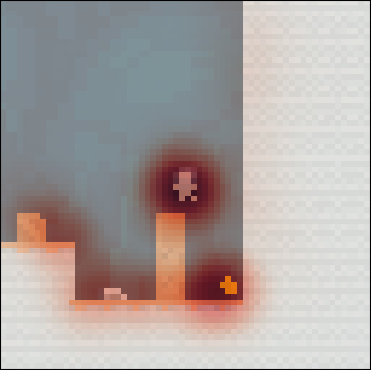}
    \includegraphics[width=0.32\textwidth]{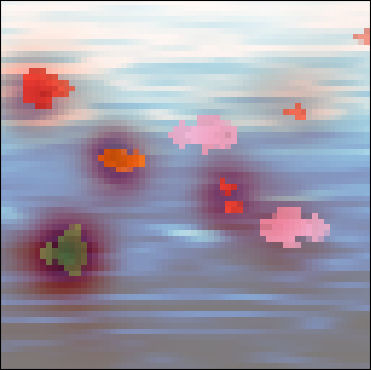}
    \includegraphics[width=0.32\textwidth]{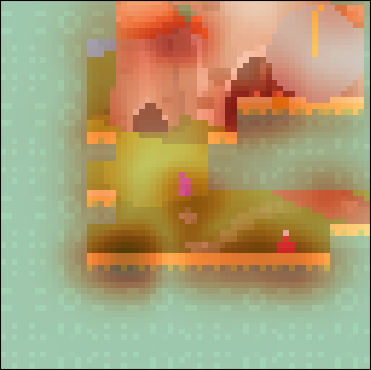}
    \caption{Saliency maps of expert networks superimposed on states sampled from each visual task.}
\end{figure}

\newpage

\section{Separability of state embeddings on visual tasks}
\label{app:sep_visual}

\begin{figure}[h]
    \vspace{8em}
    \centering
    \includegraphics[width=0.49\textwidth]{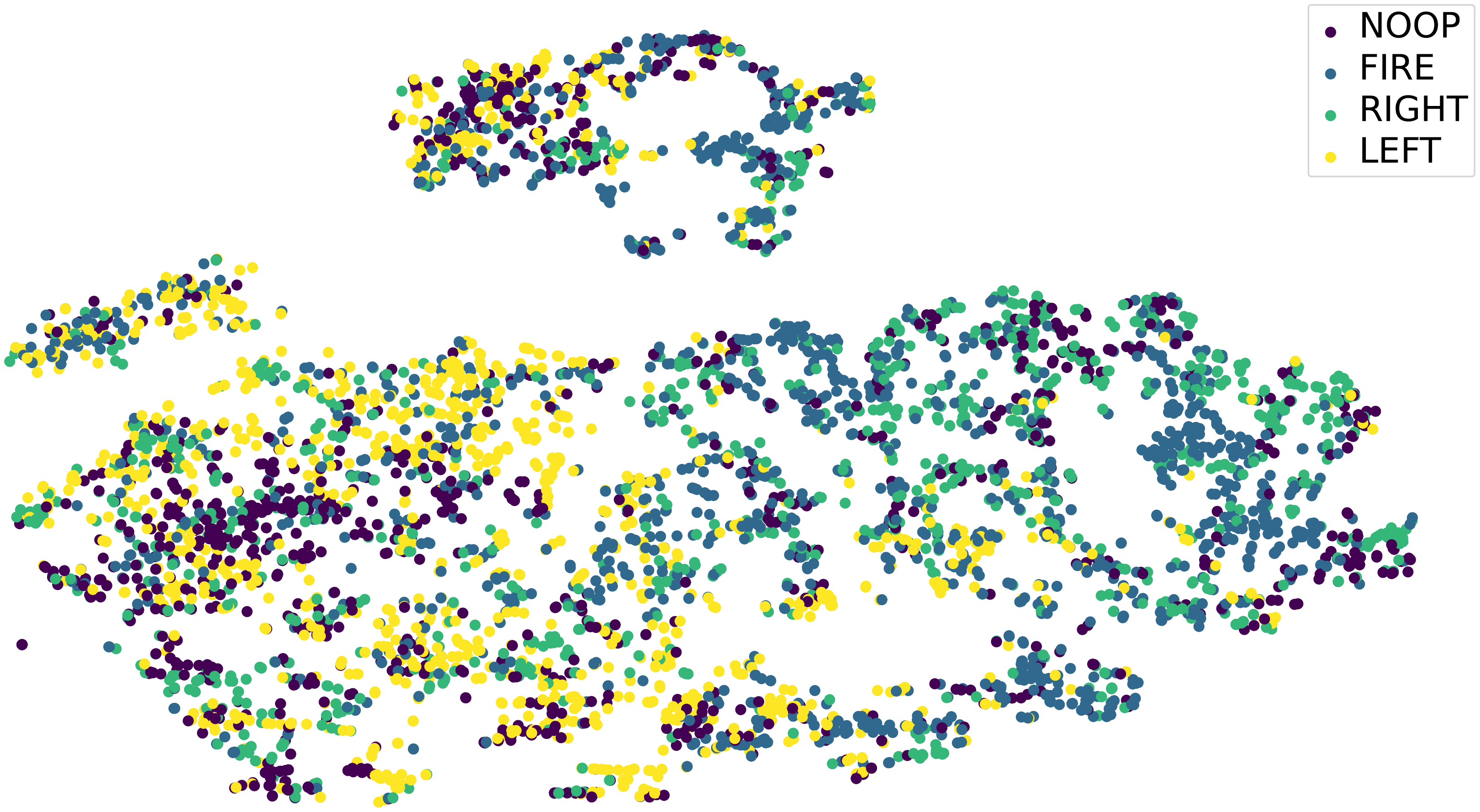}
    \includegraphics[width=0.49\textwidth]{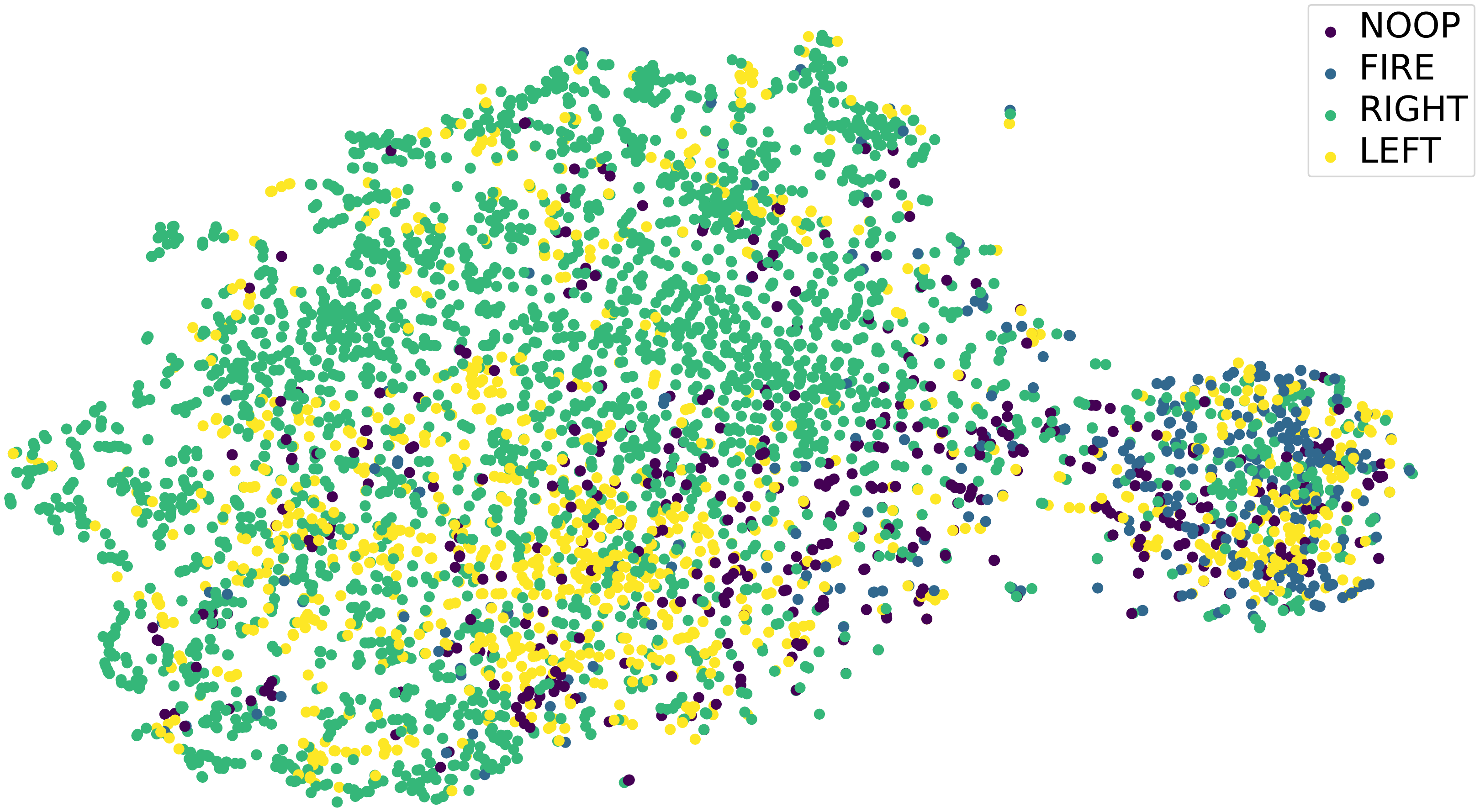}
    \vspace{1em}
    \includegraphics[width=0.49\textwidth]{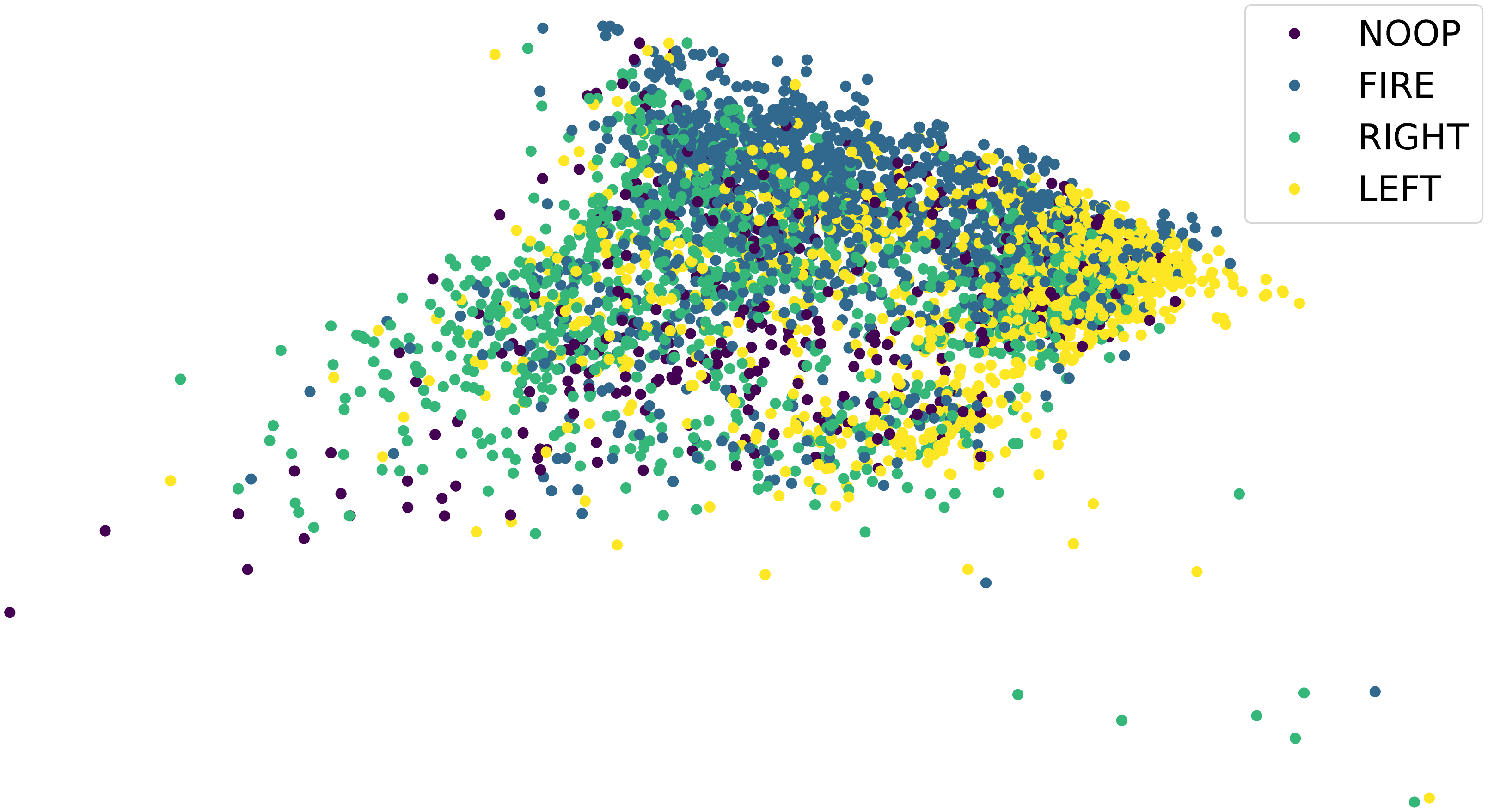}
    \includegraphics[width=0.49\textwidth]{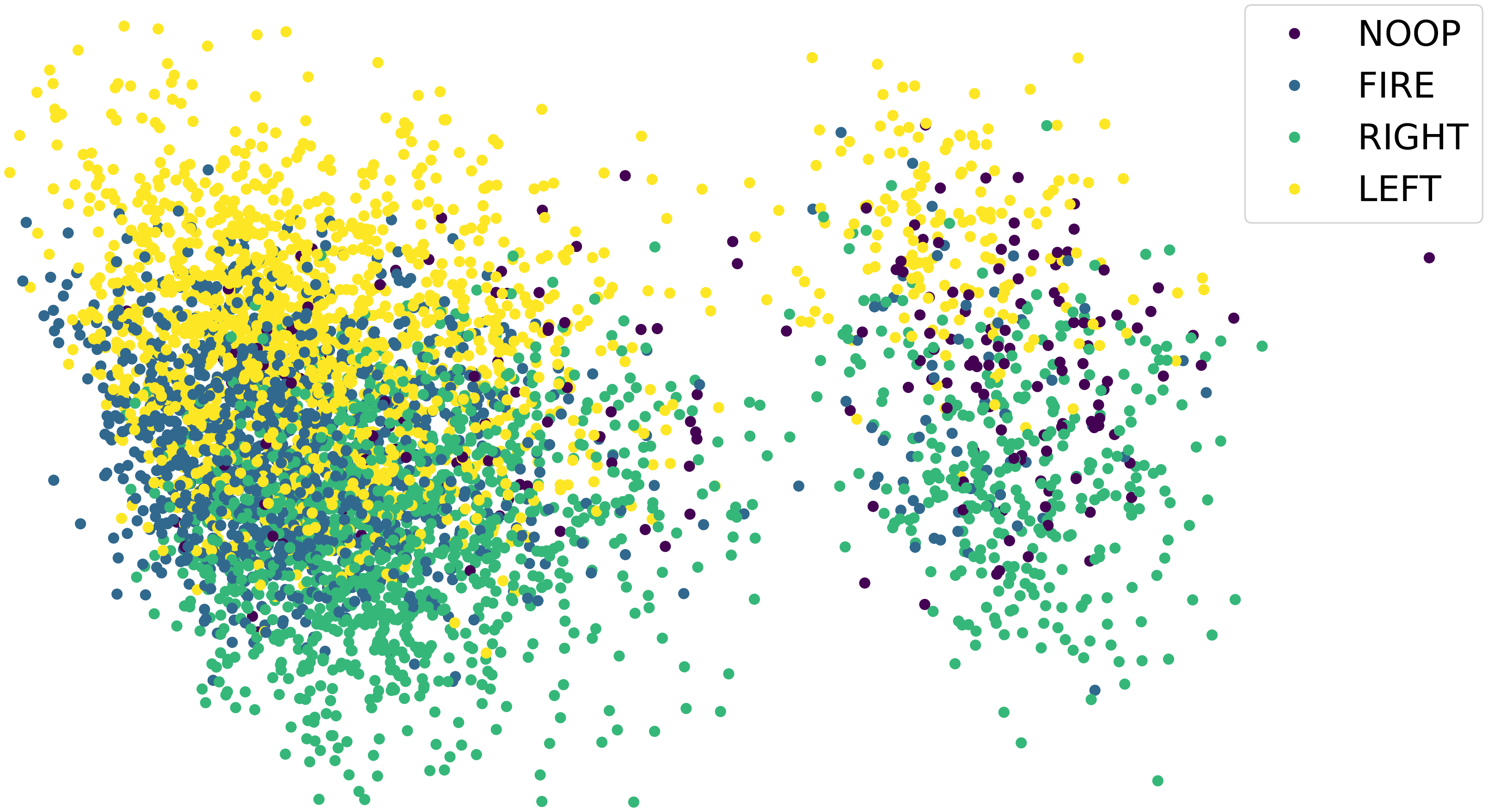}
    \caption{t-SNE (top) and PLS (bottom) plots of the state embedding for an expert (left) and the student (right) on Breakout.}
    \label{fig:tsne_breakout}
\end{figure}

\begin{figure}
    \centering
    \includegraphics[width=0.49\textwidth]{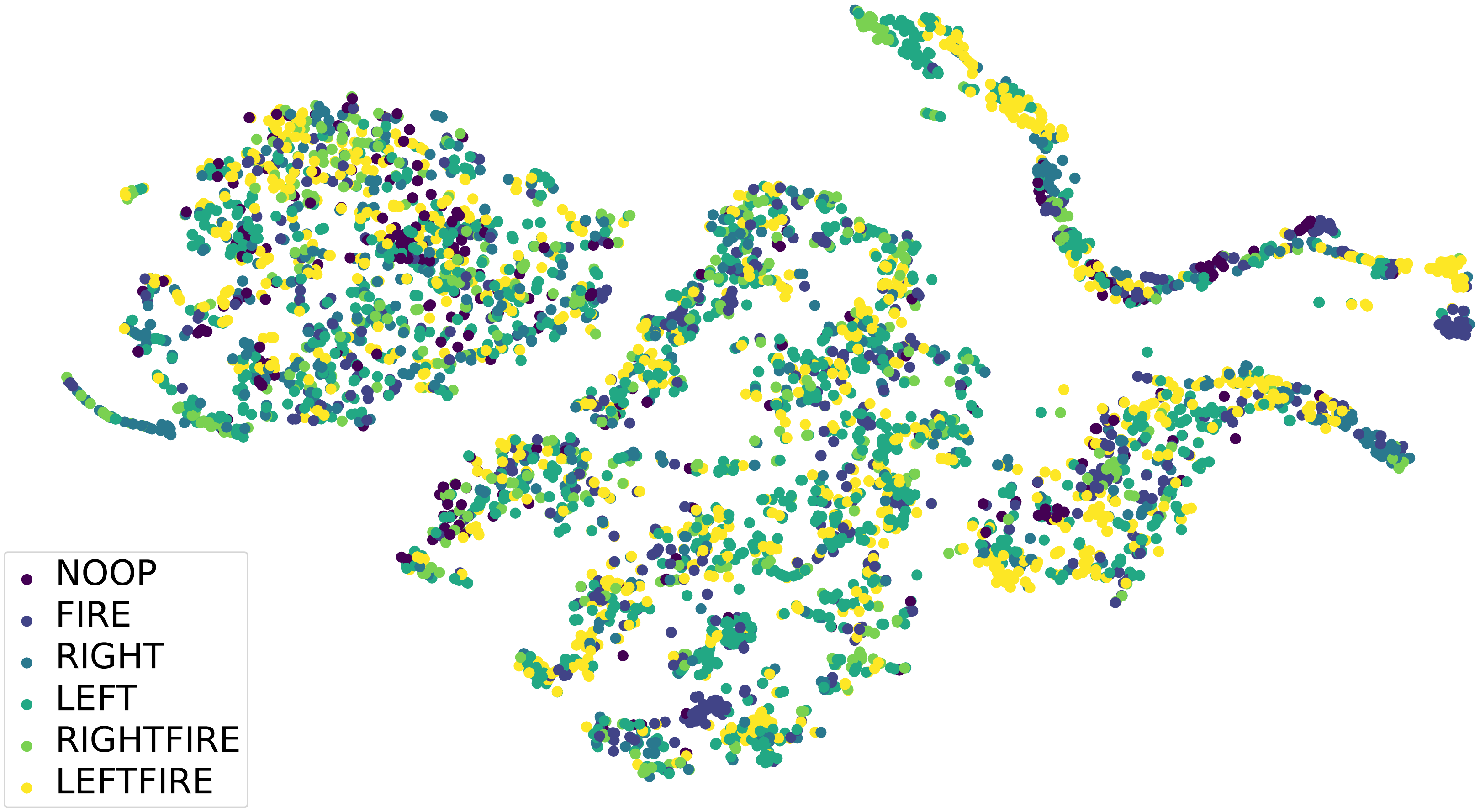}
    \includegraphics[width=0.49\textwidth]{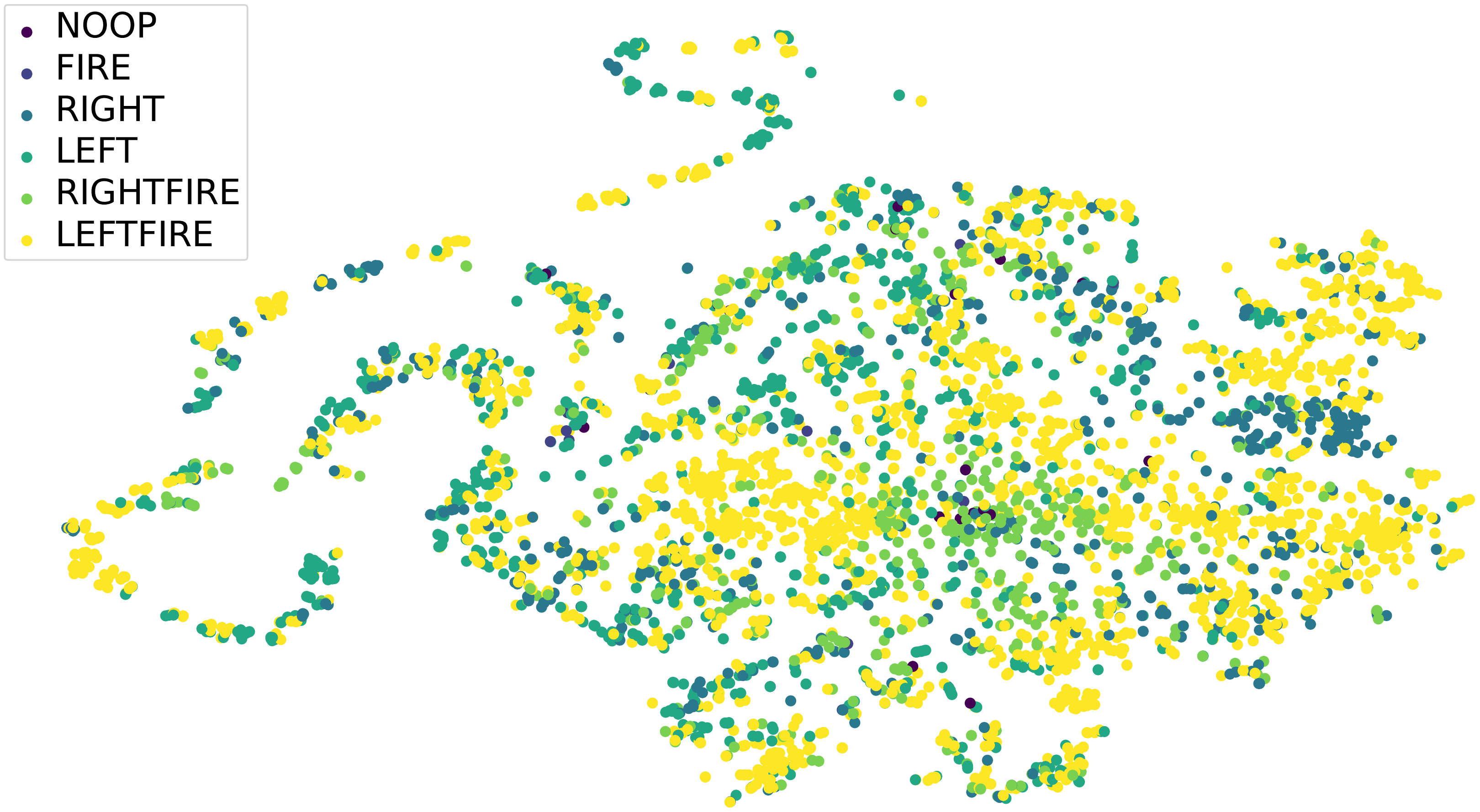}
    \vspace{1em}
    \includegraphics[width=0.49\textwidth]{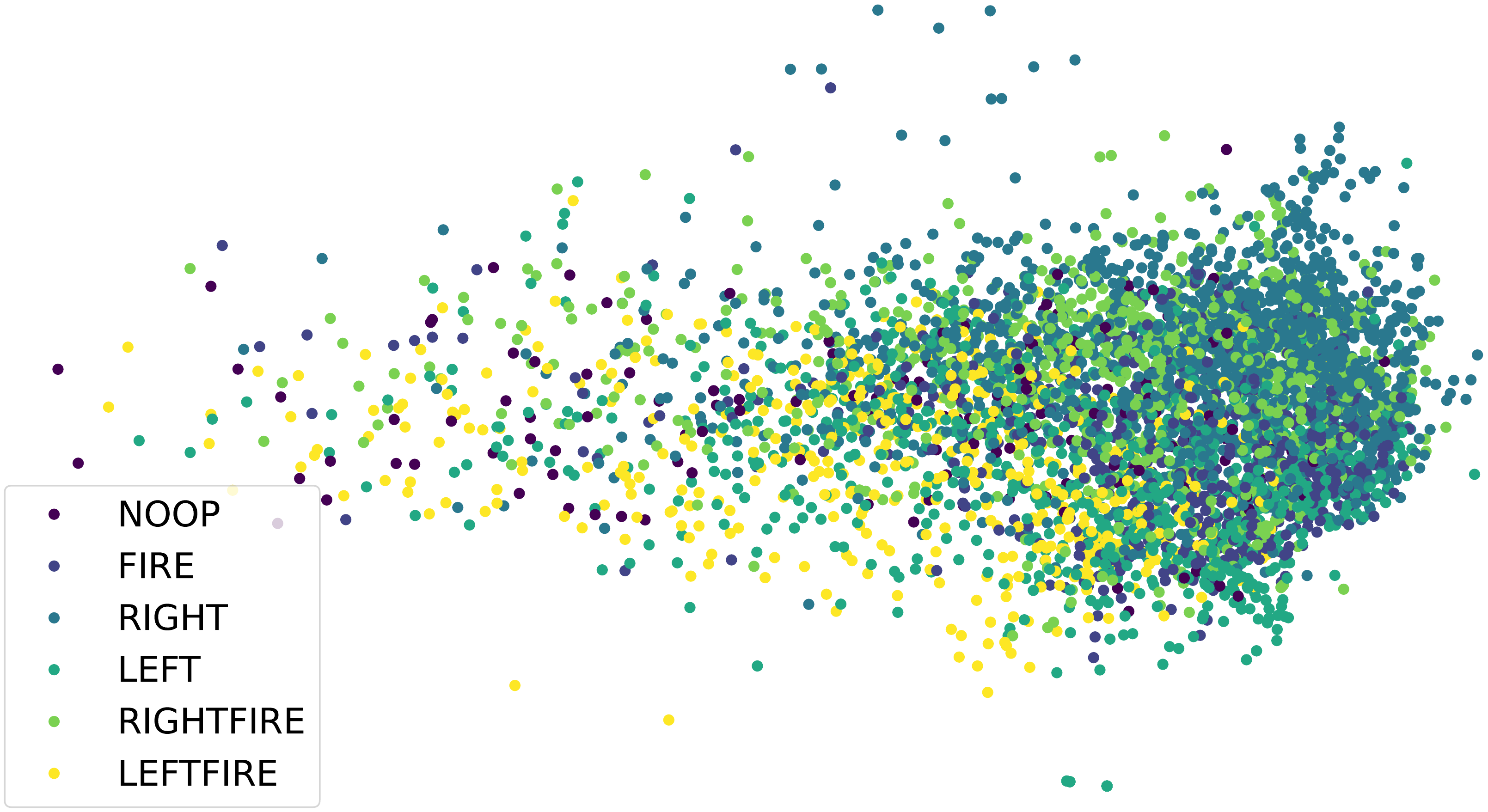}
    \includegraphics[width=0.49\textwidth]{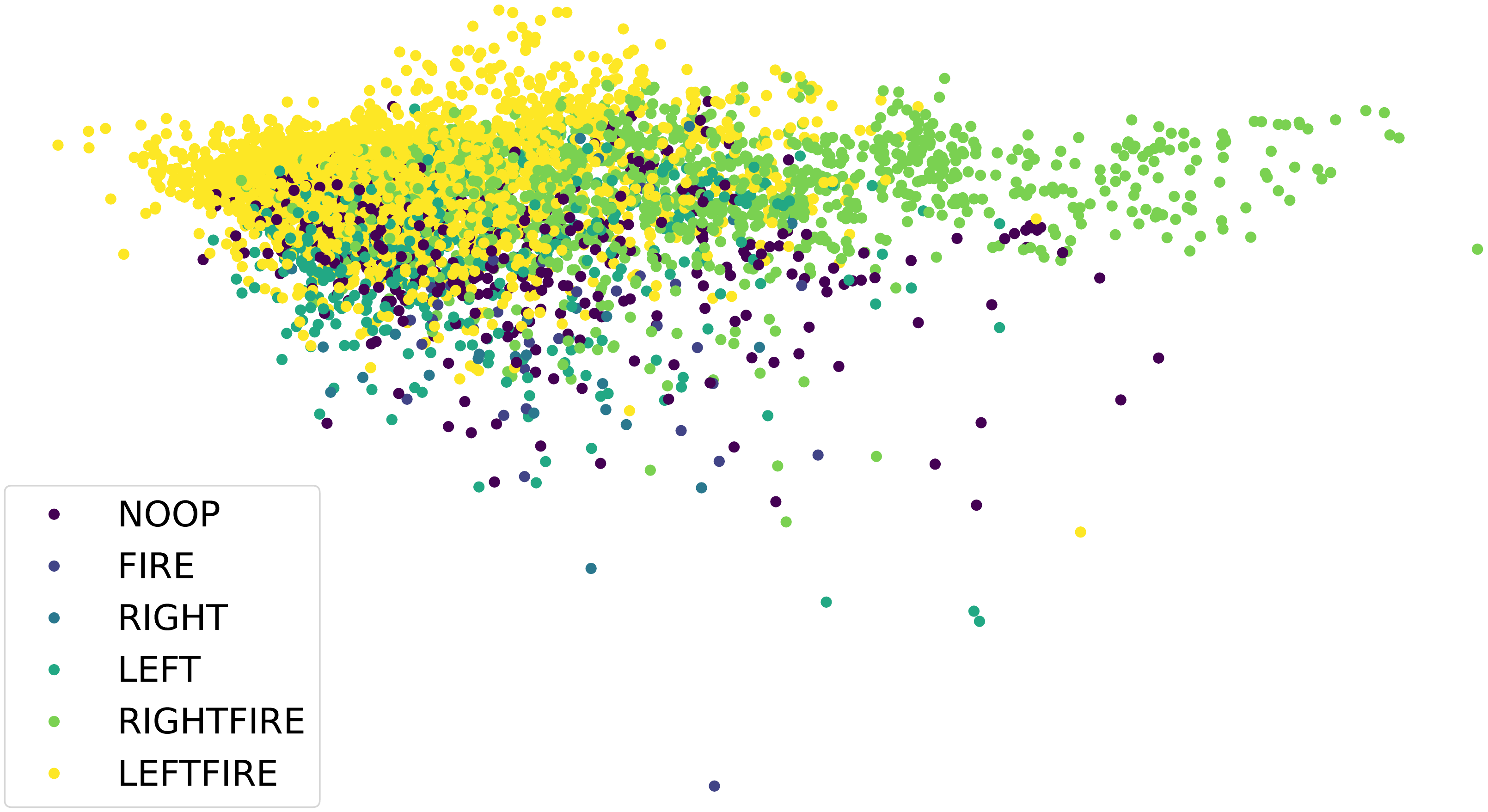}
    \caption{t-SNE (top) and PLS (bottom) plots of the state embedding for an expert (left) and the student (right) on Carnival.}
    \label{fig:tsne_carnival}
\end{figure}

\begin{figure}
    \centering
    \includegraphics[width=0.49\textwidth]{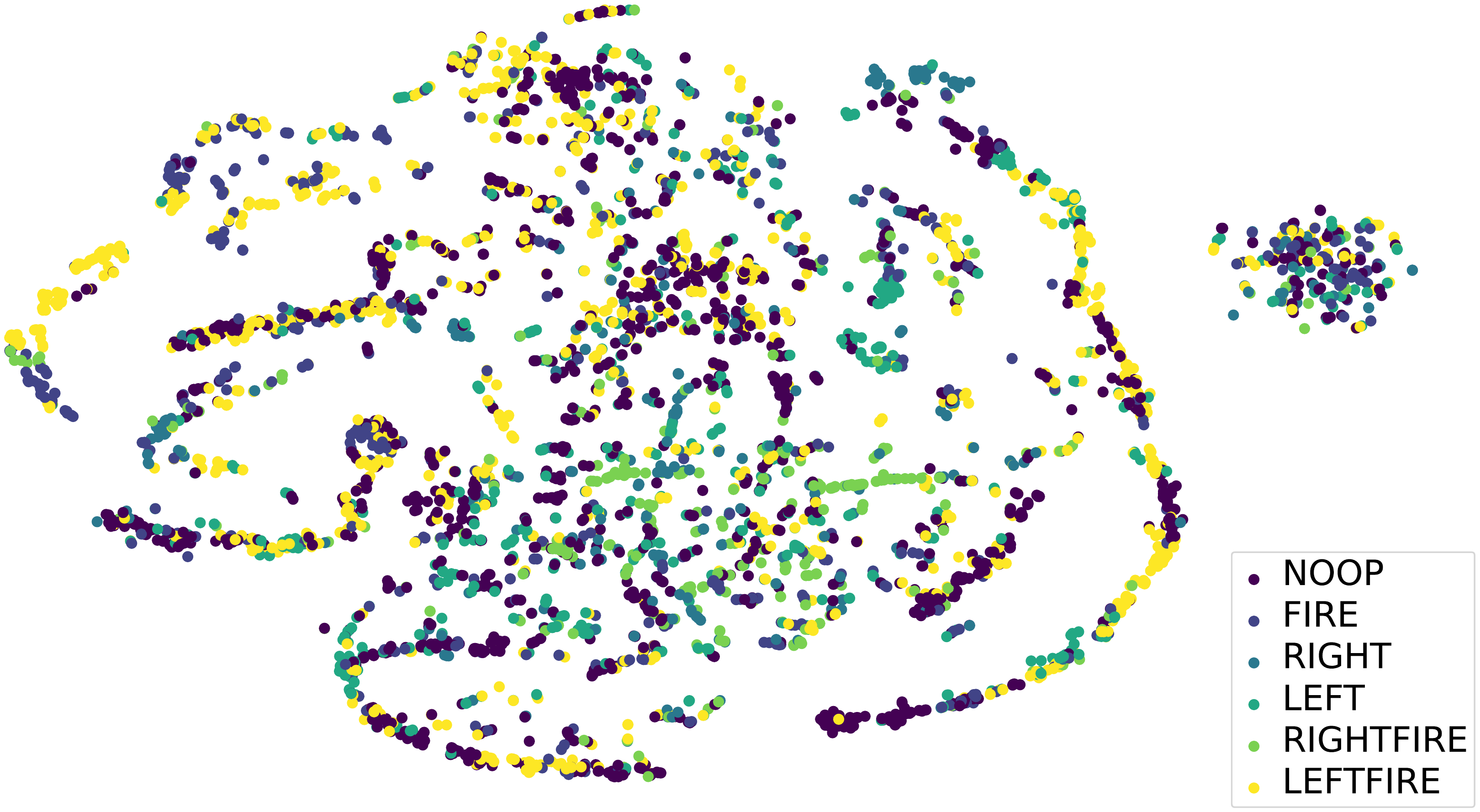}
    \includegraphics[width=0.49\textwidth]{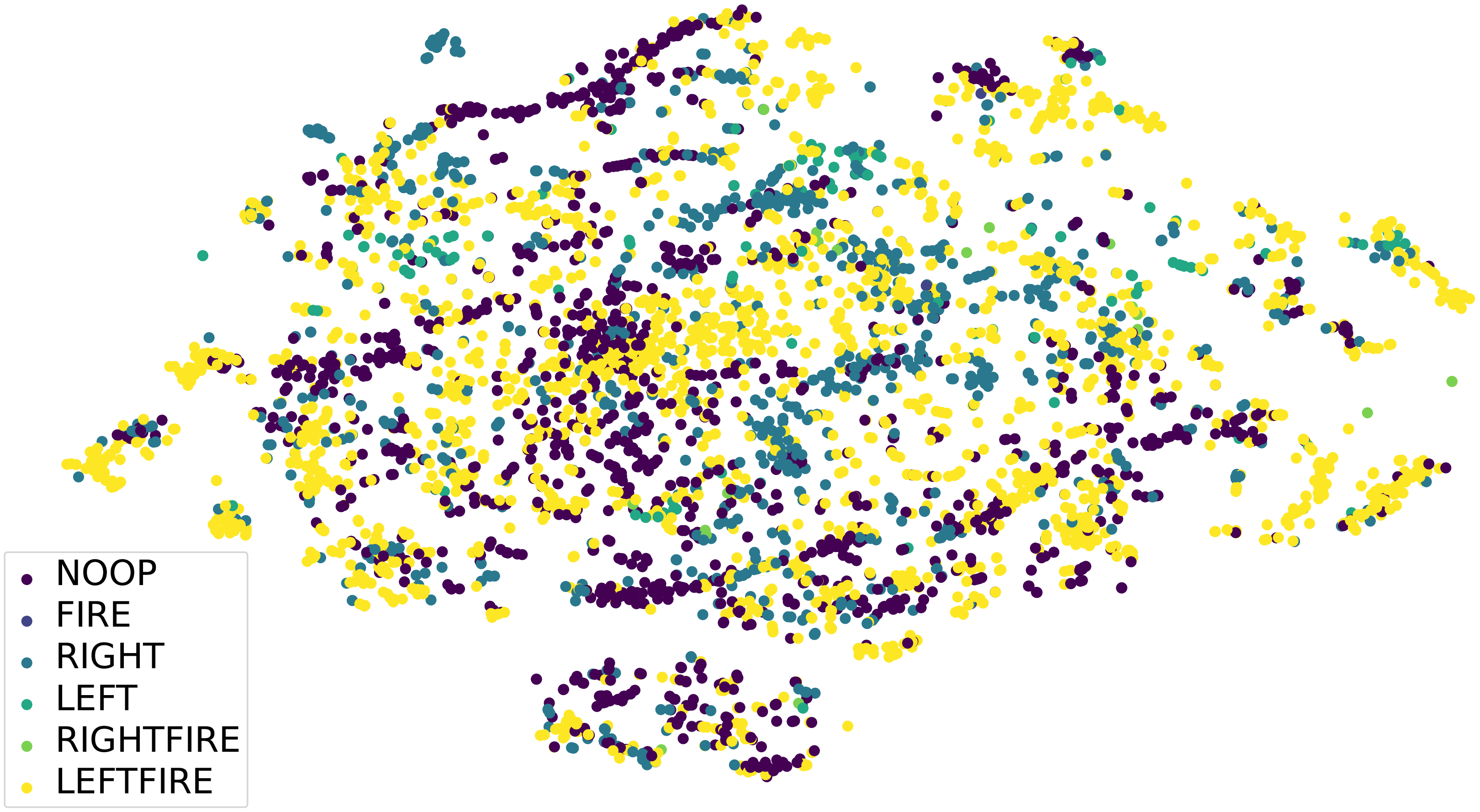}
    \vspace{1em}
    \includegraphics[width=0.49\textwidth]{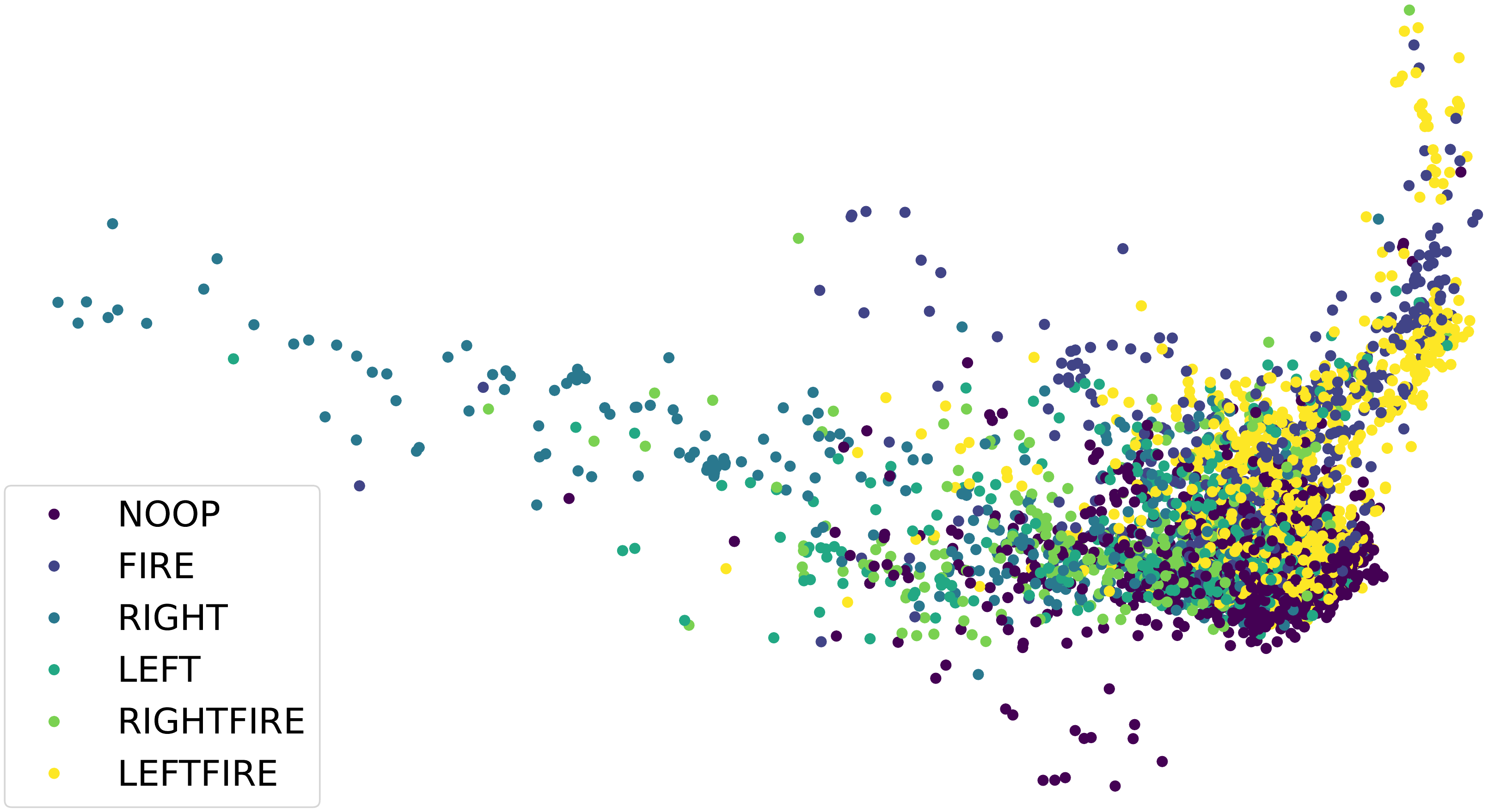}
    \includegraphics[width=0.49\textwidth]{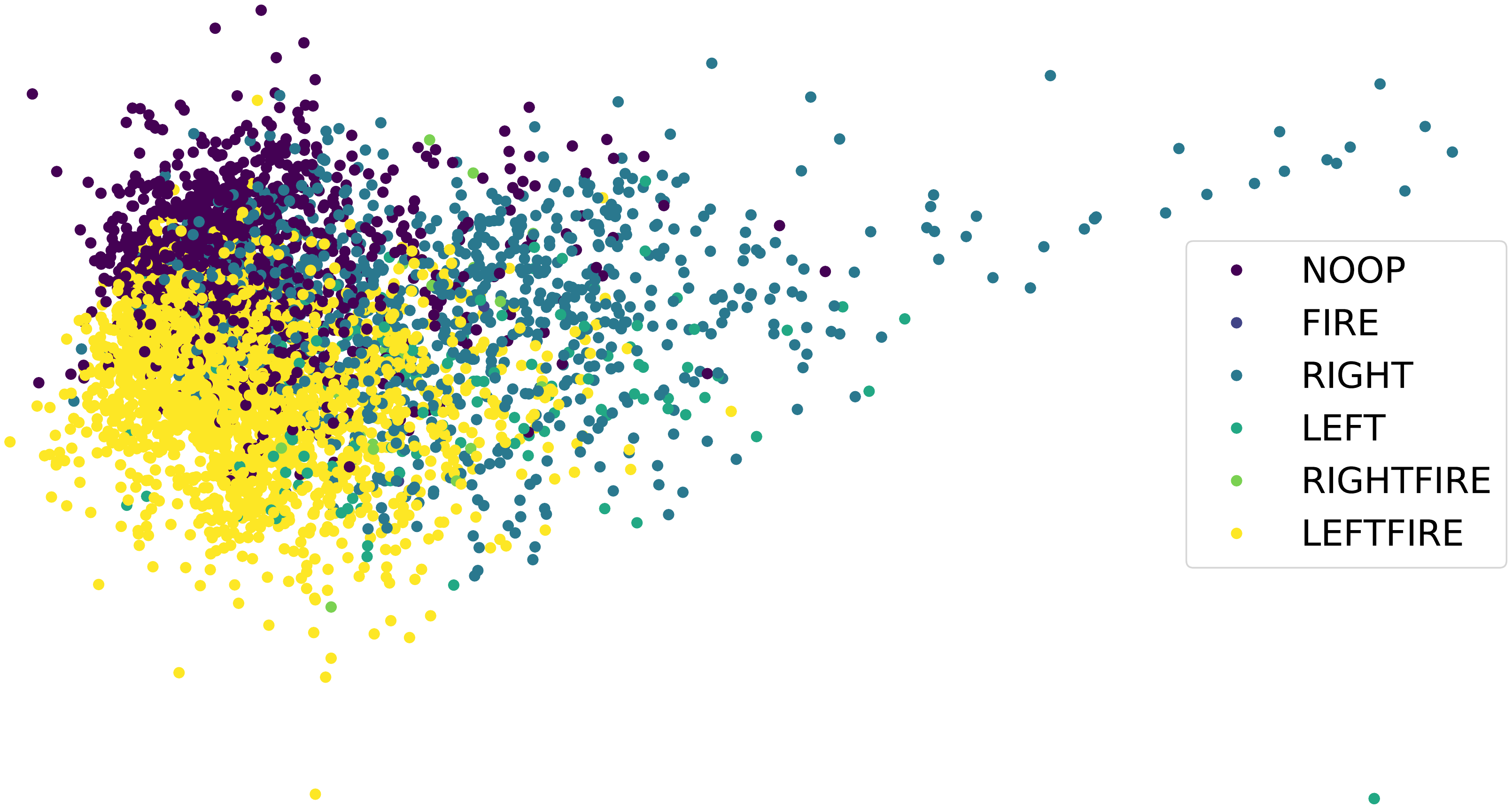}
    \caption{t-SNE (top) and PLS (bottom) plots of the state embedding for an expert (left) and the student (right) on Pong.}
    \label{fig:tsne_pong}
\end{figure}

\begin{figure}
    \centering
    \includegraphics[width=0.49\textwidth]{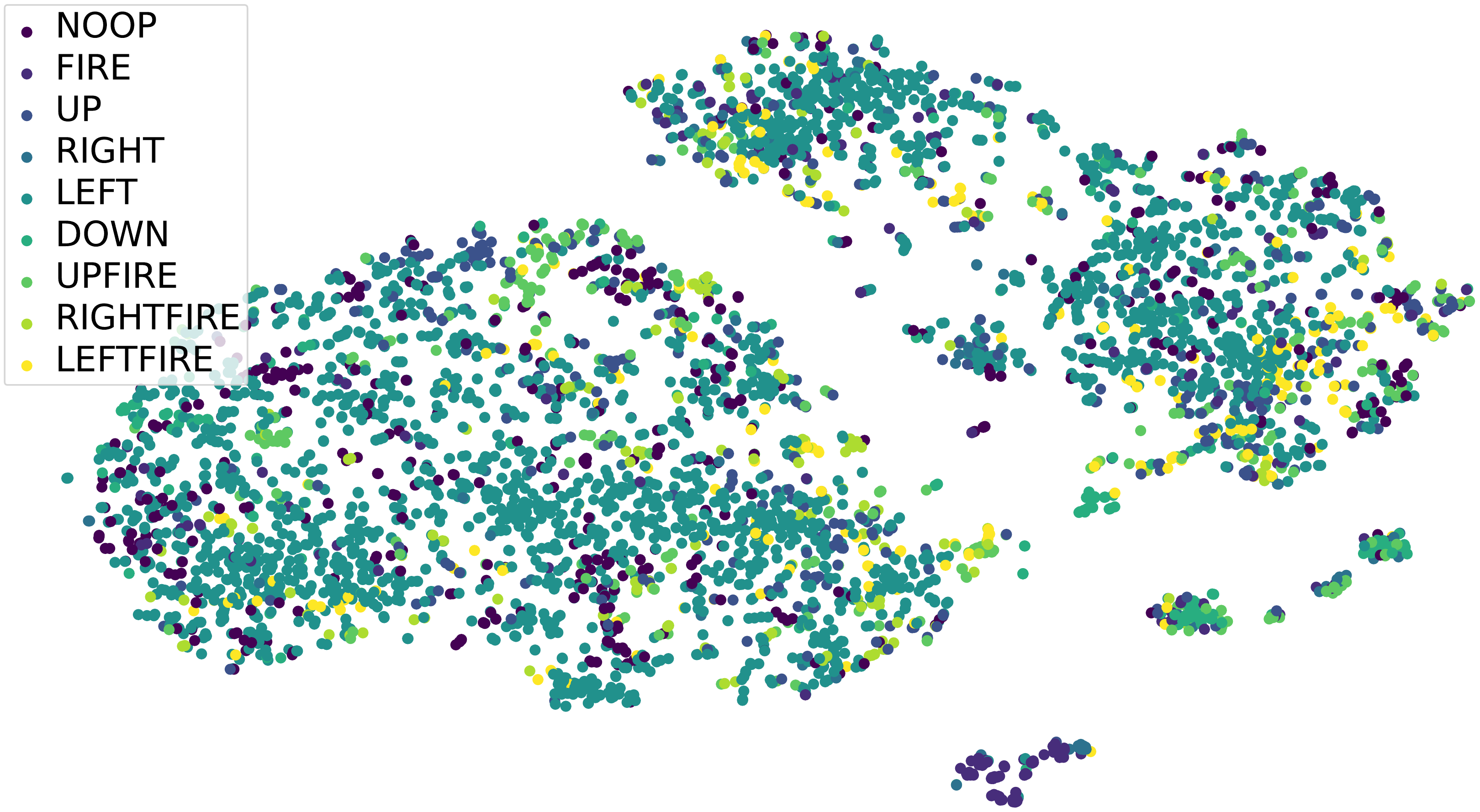}
    \includegraphics[width=0.49\textwidth]{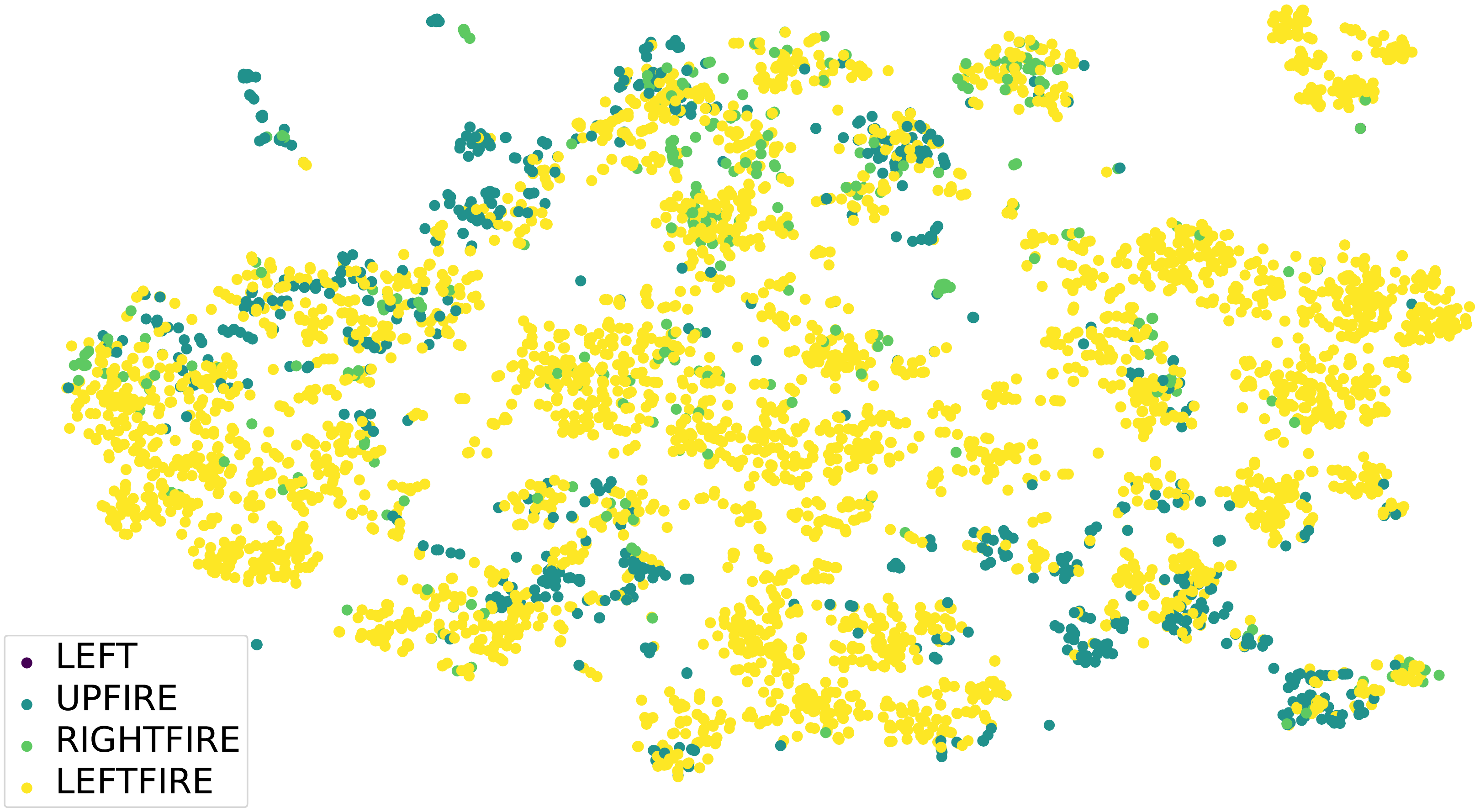}
    \vspace{1em}
    \includegraphics[width=0.49\textwidth]{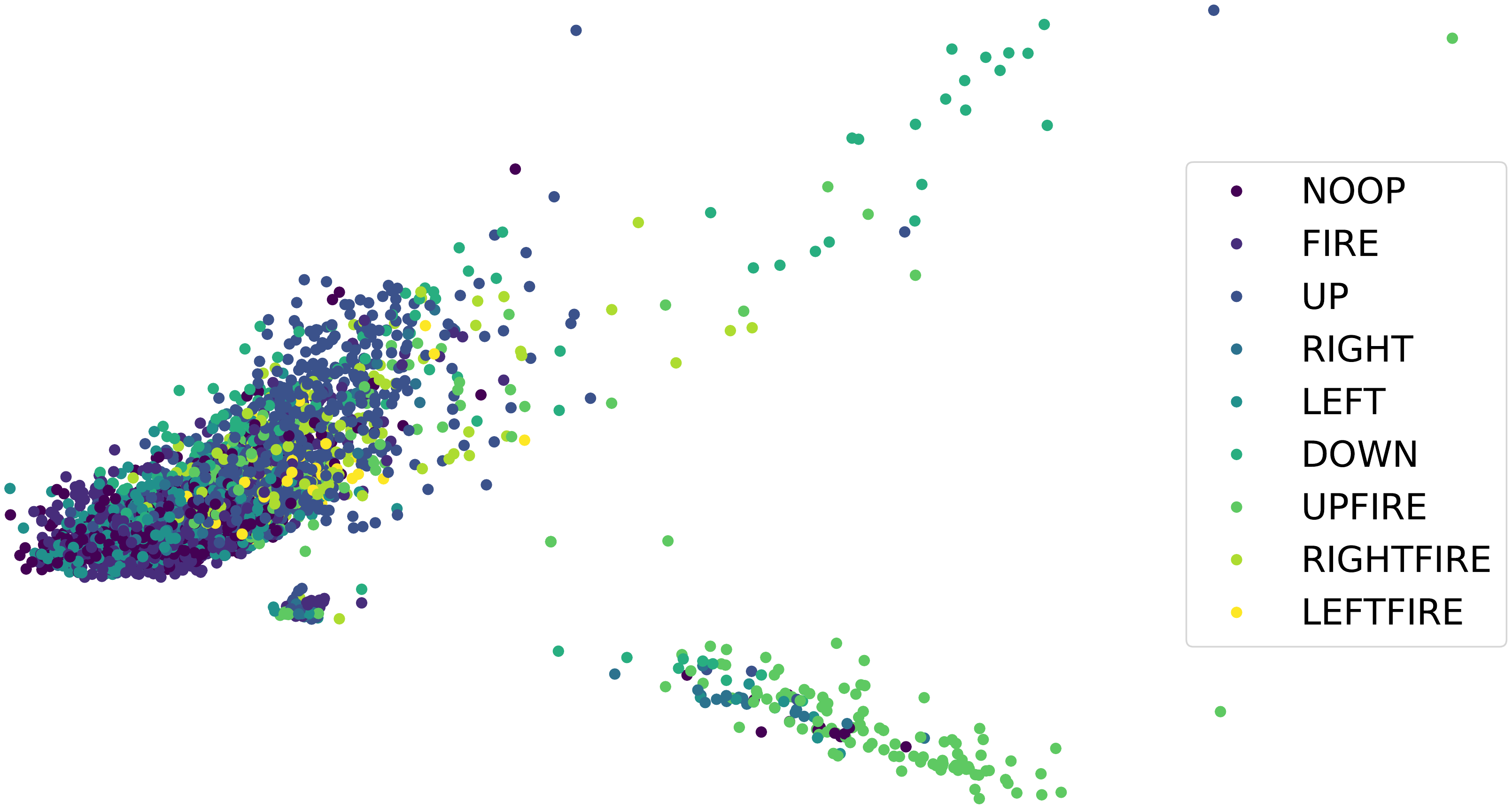}
    \includegraphics[width=0.49\textwidth]{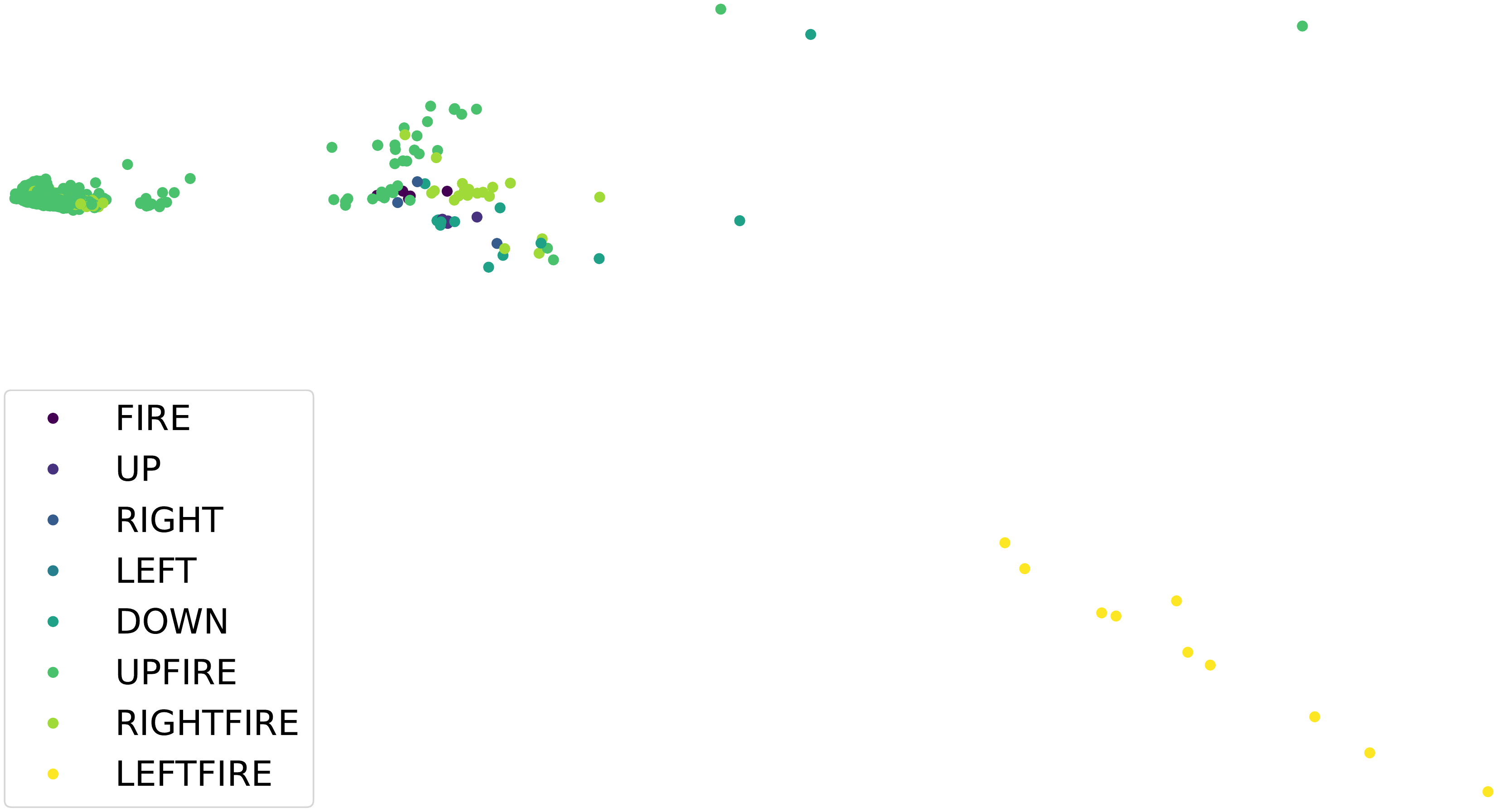}
    \caption{t-SNE (top) and PLS (bottom) plots of the state embedding for an expert (left) and the student (right) on VideoPinball.}
    \label{fig:tsne_videopinball}
\end{figure}

\begin{figure}
    \centering
    \includegraphics[width=0.49\textwidth]{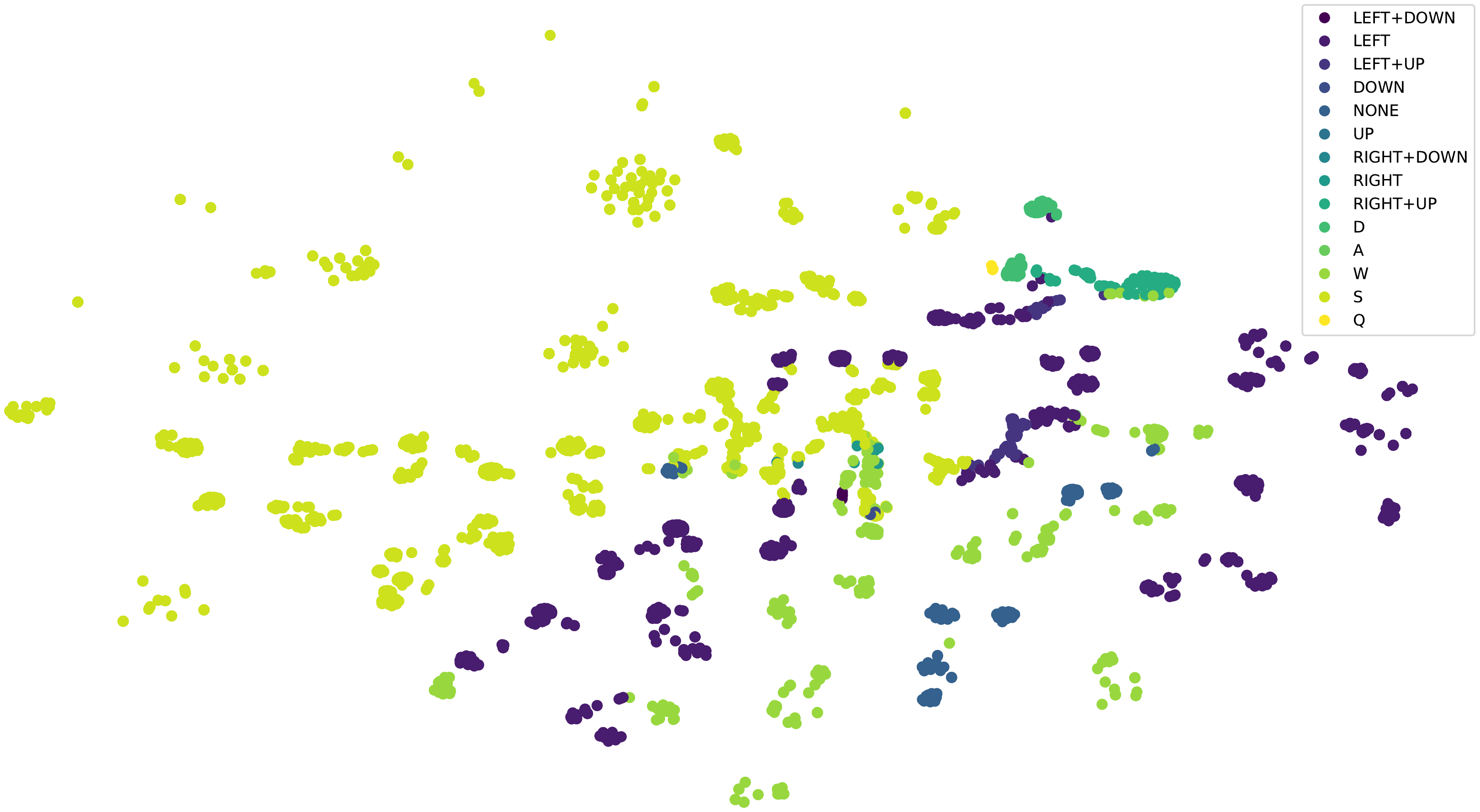}
    \includegraphics[width=0.49\textwidth]{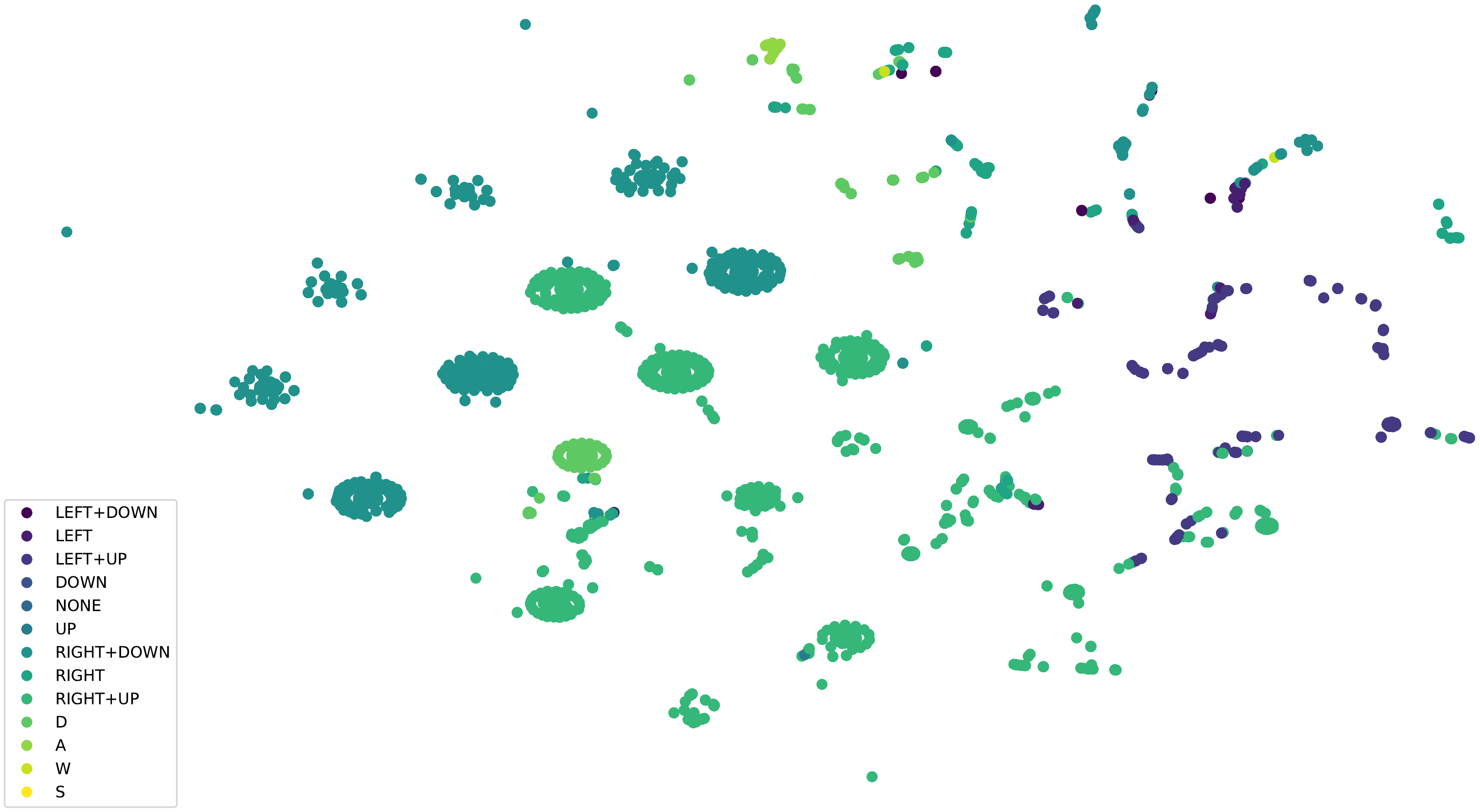}
    \vspace{1em}
    \includegraphics[width=0.49\textwidth]{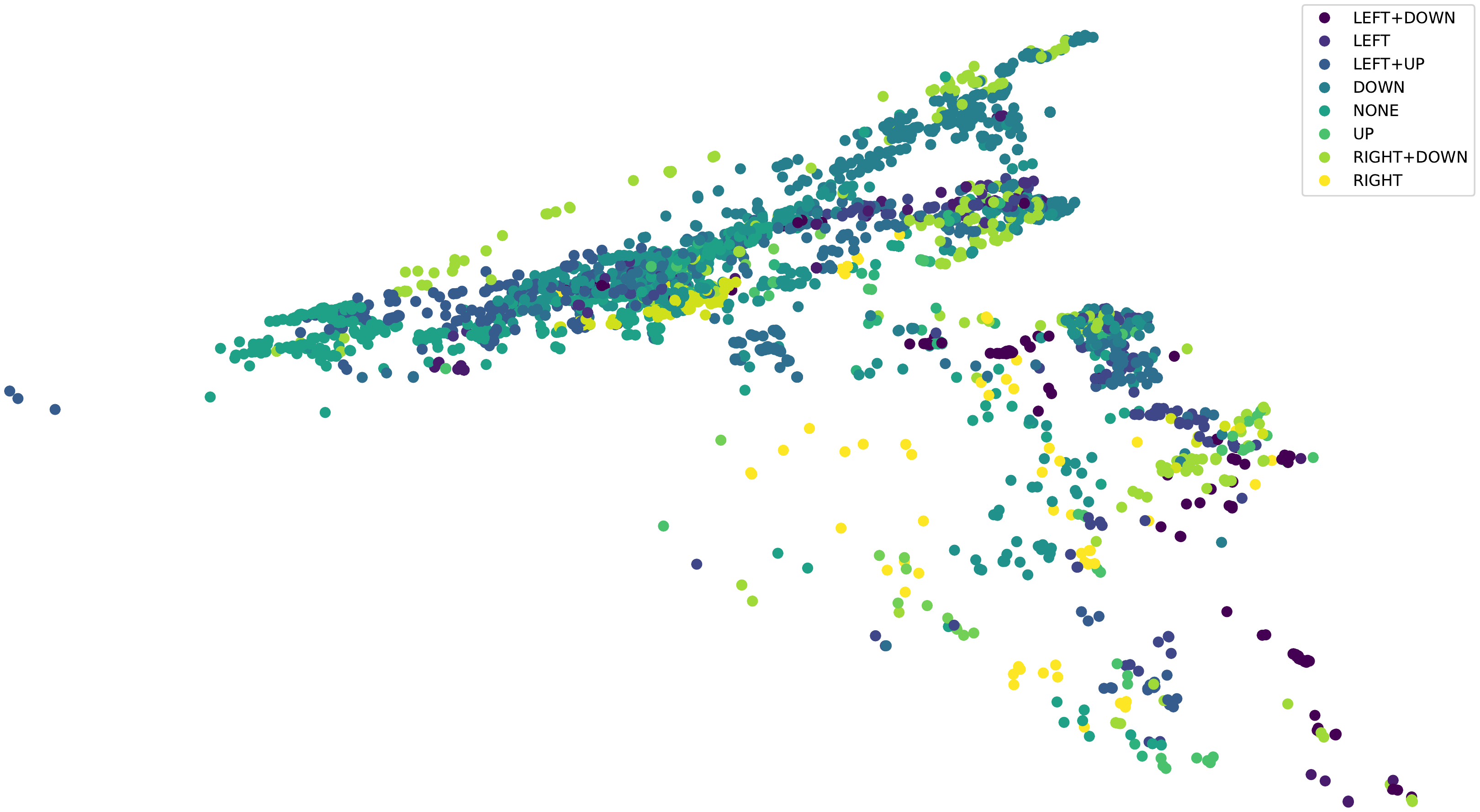}
    \includegraphics[width=0.49\textwidth]{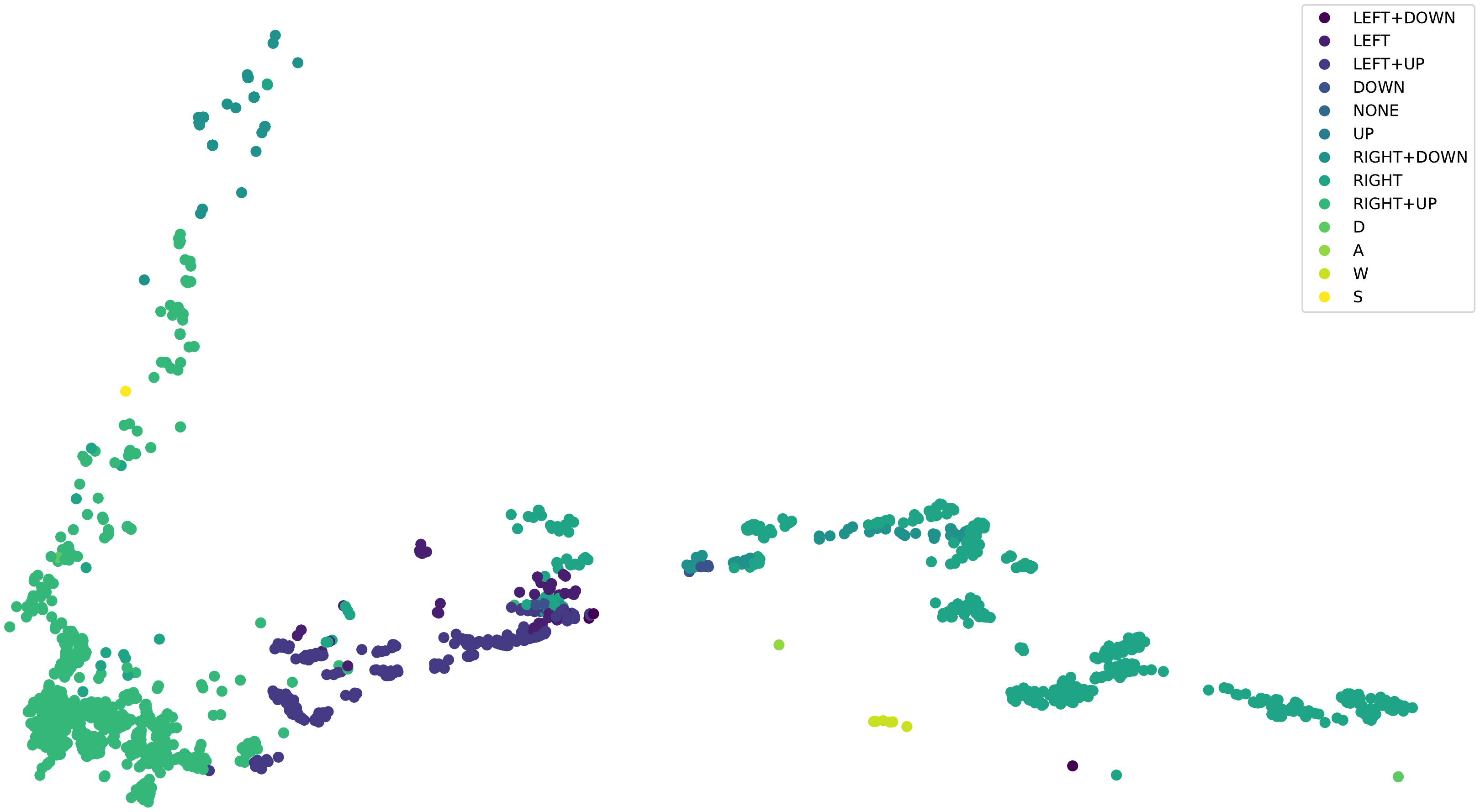}
    \caption{t-SNE (top) and PLS (bottom) plots of the state embedding for an expert (left) and the student (right) on CoinRun.}
    \label{fig:tsne_coinrun}
\end{figure}

\begin{figure}
    \centering
    \includegraphics[width=0.49\textwidth]{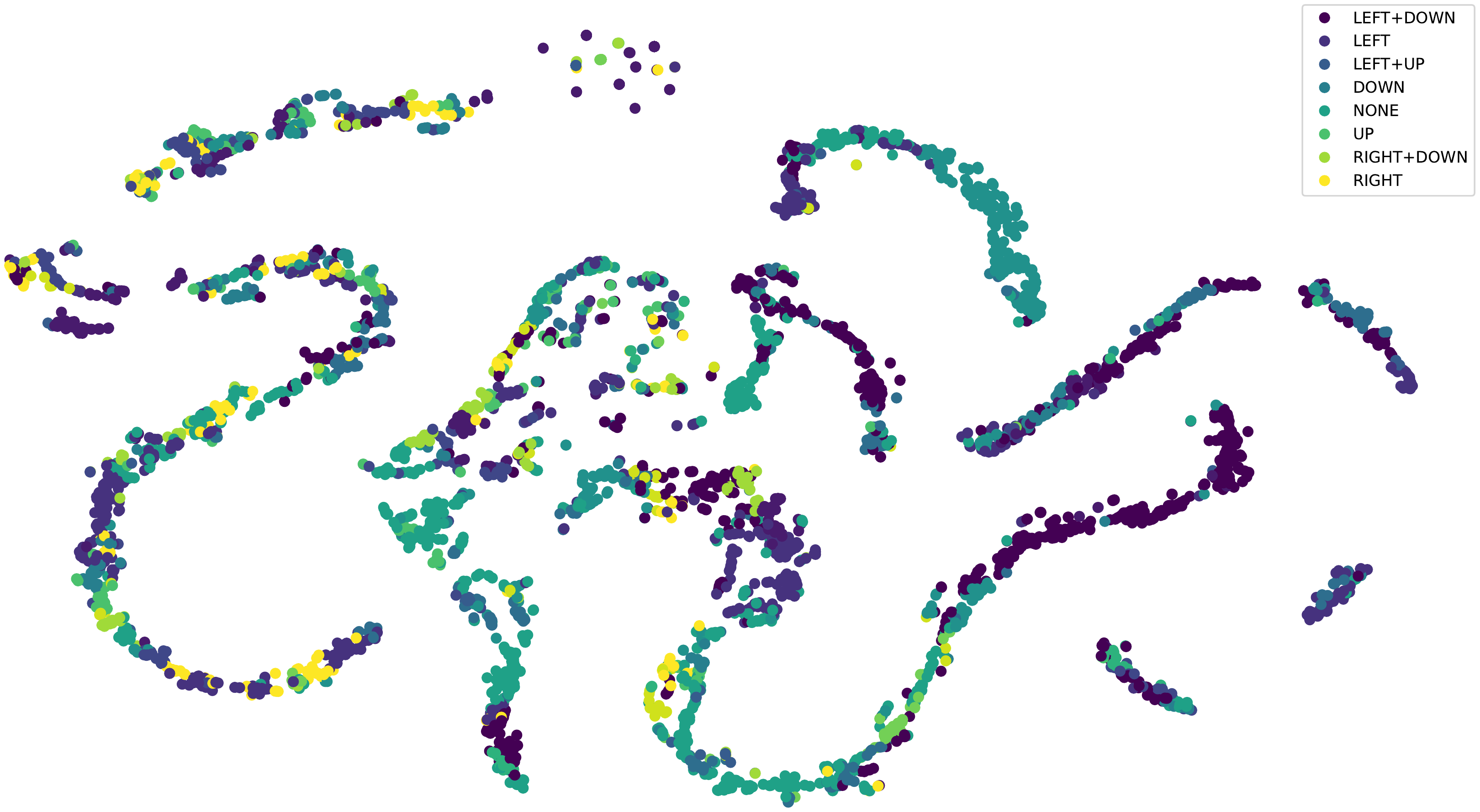}
    \includegraphics[width=0.49\textwidth]{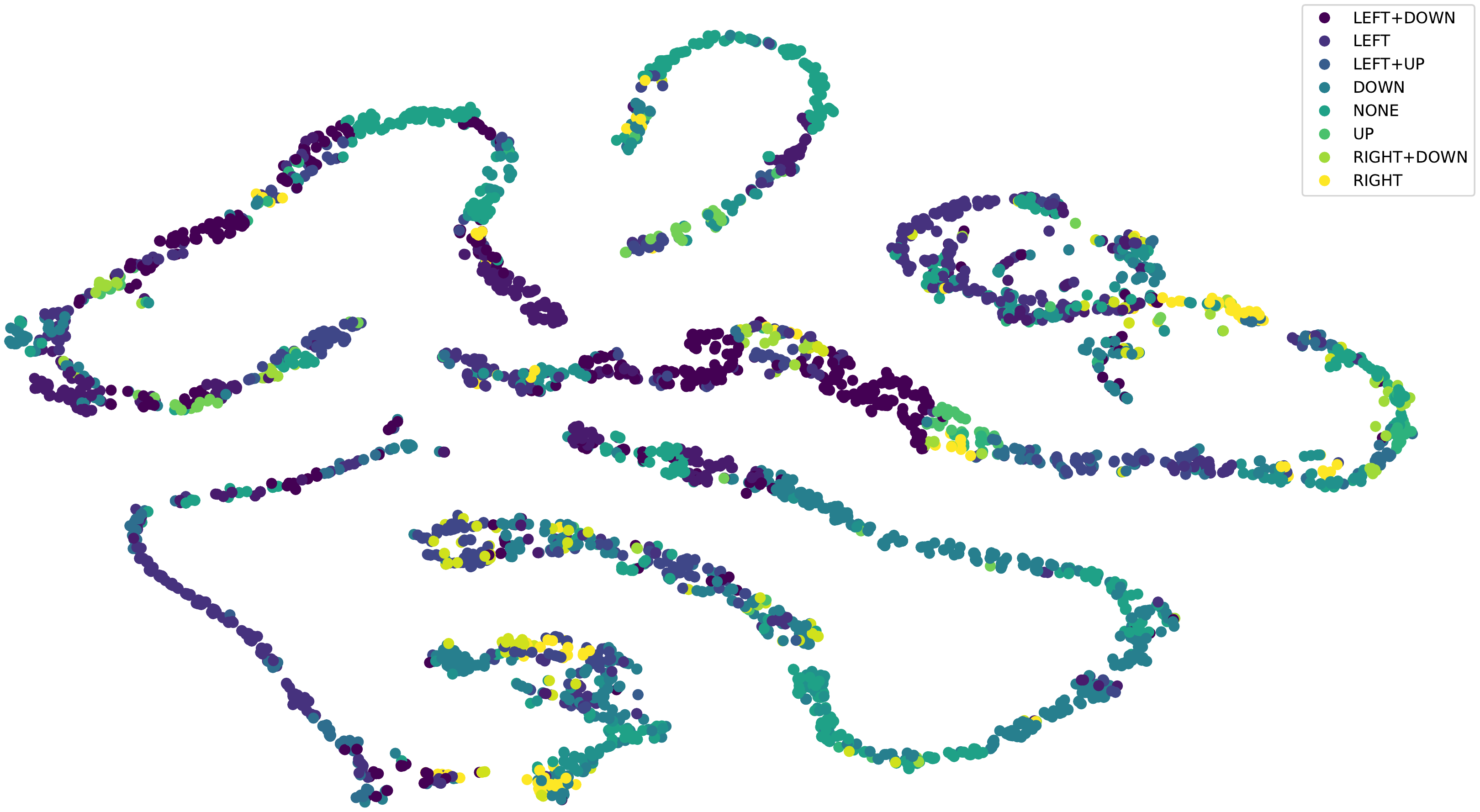}
    \vspace{1em}
    \includegraphics[width=0.49\textwidth]{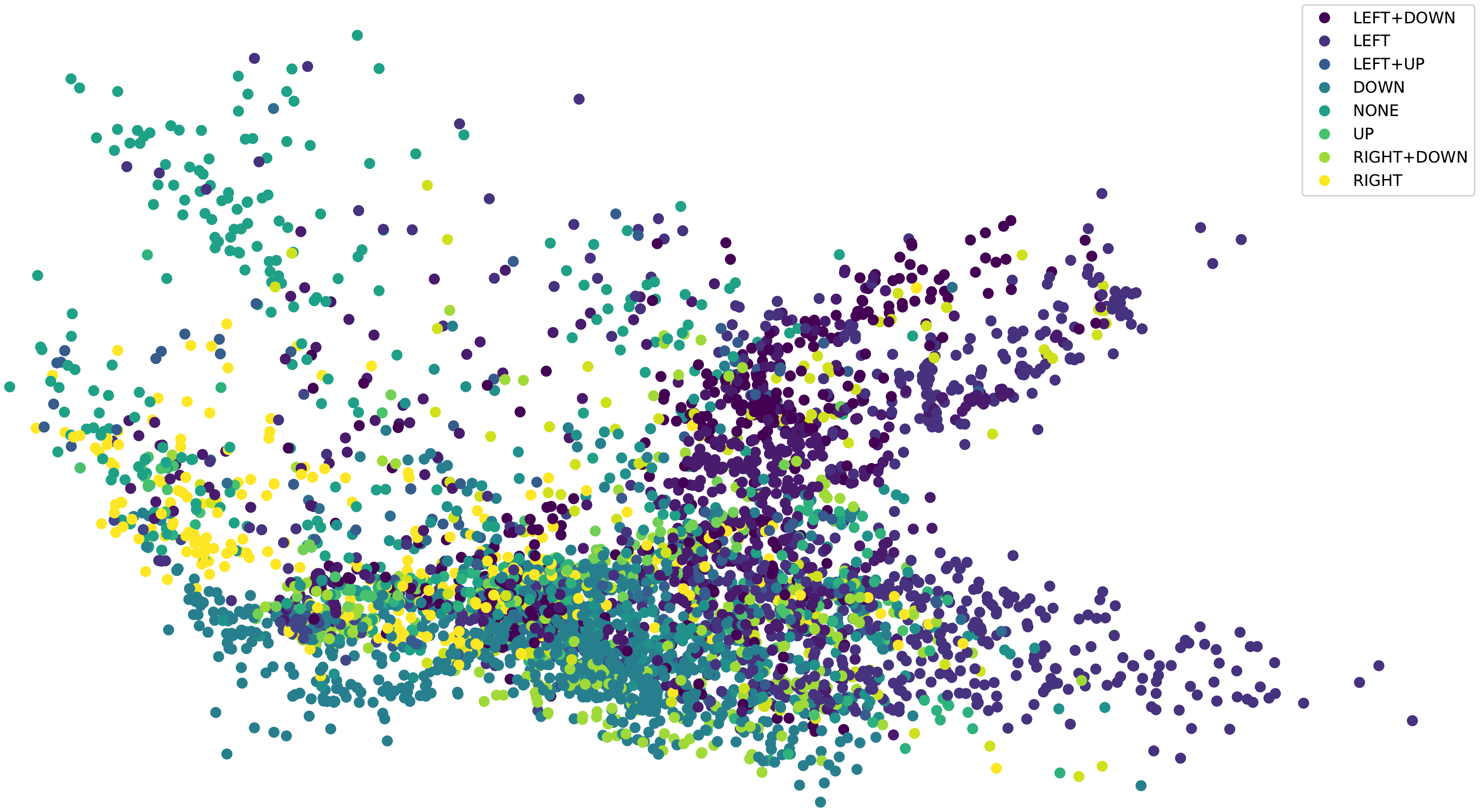}
    \includegraphics[width=0.49\textwidth]{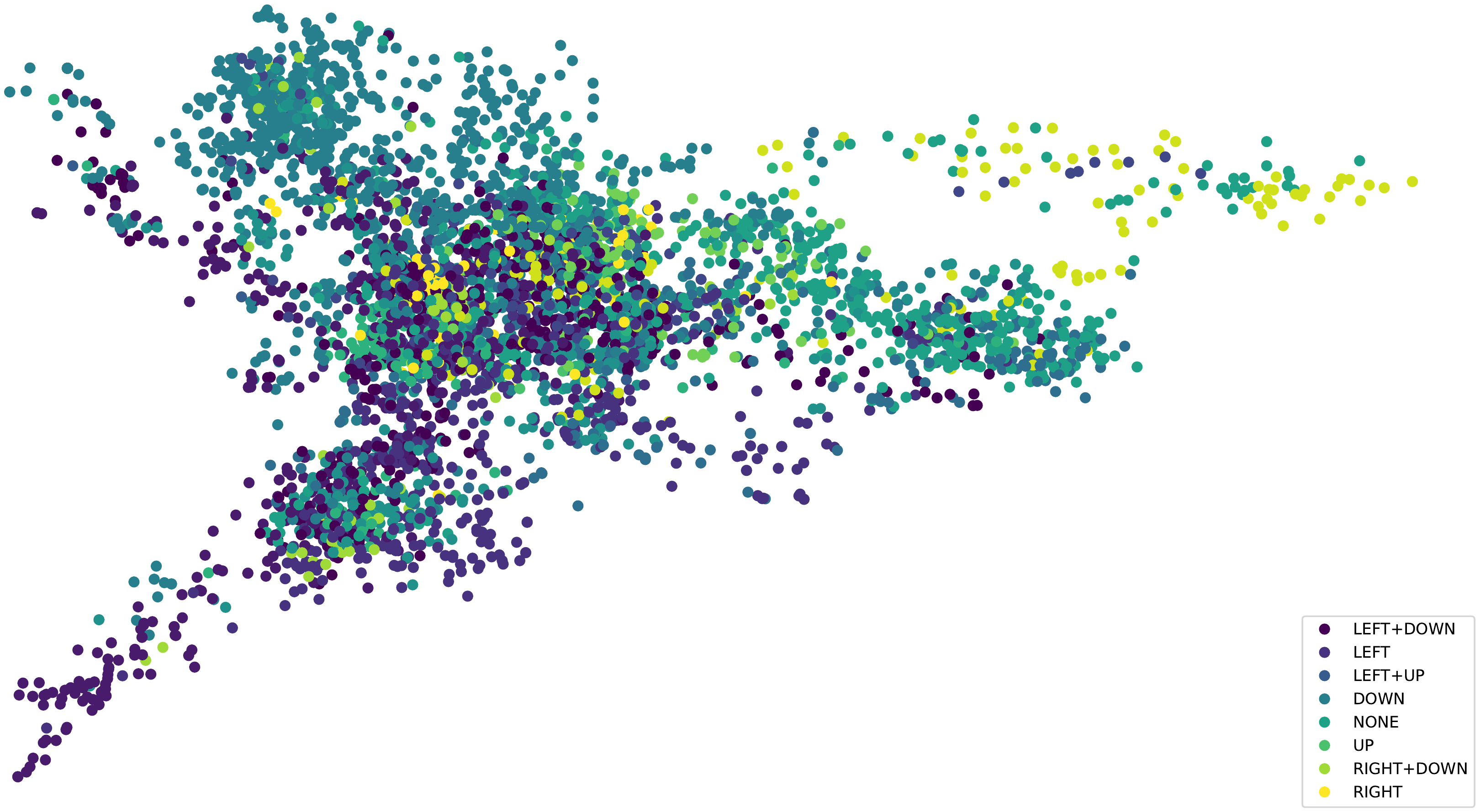}
    \caption{t-SNE (top) and PLS (bottom) plots of the state embedding for an expert (left) and the student (right) on BigFish.}
    \label{fig:tsne_bigfish}
\end{figure}

\begin{figure}
    \centering
    \includegraphics[width=0.49\textwidth]{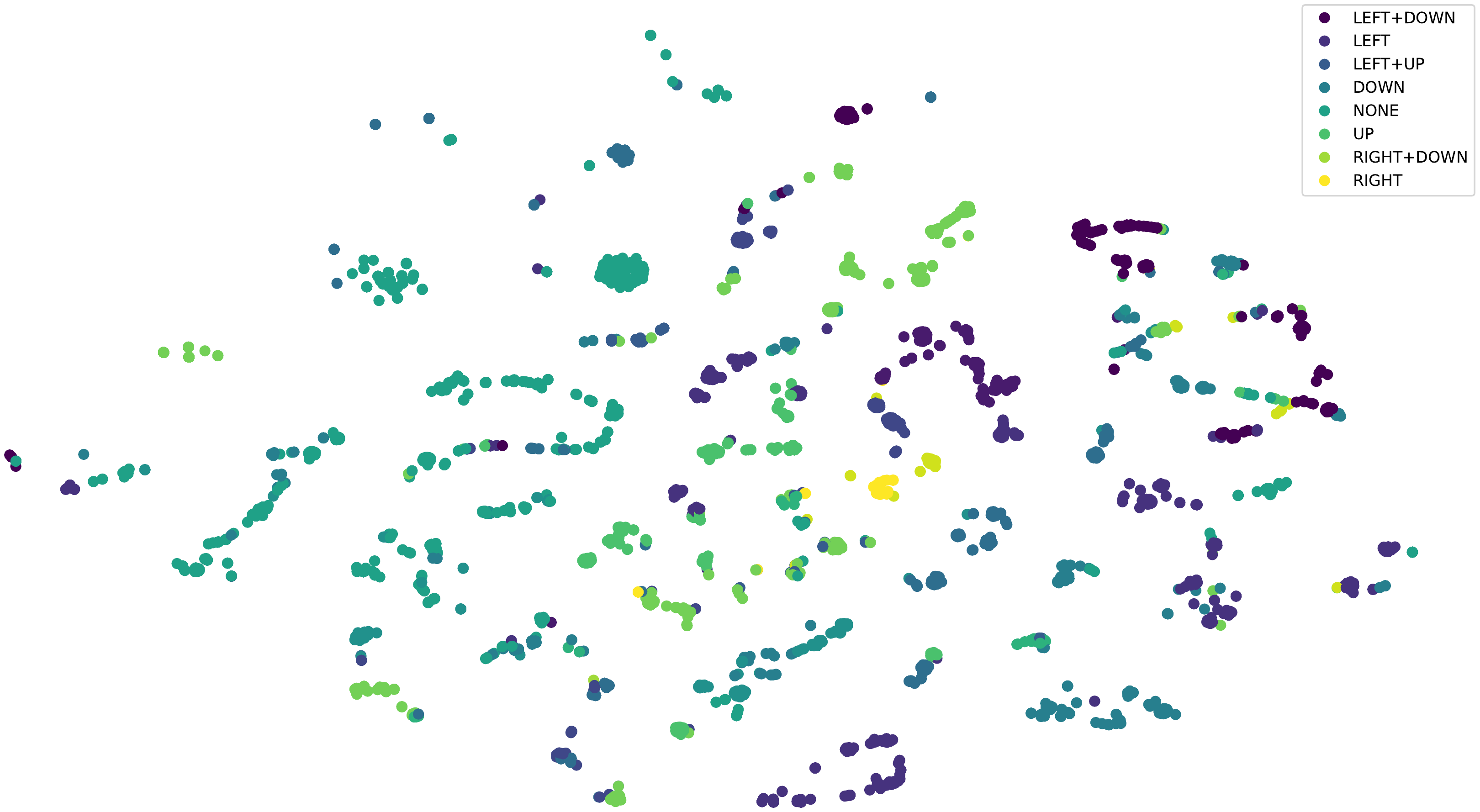}
    \includegraphics[width=0.49\textwidth]{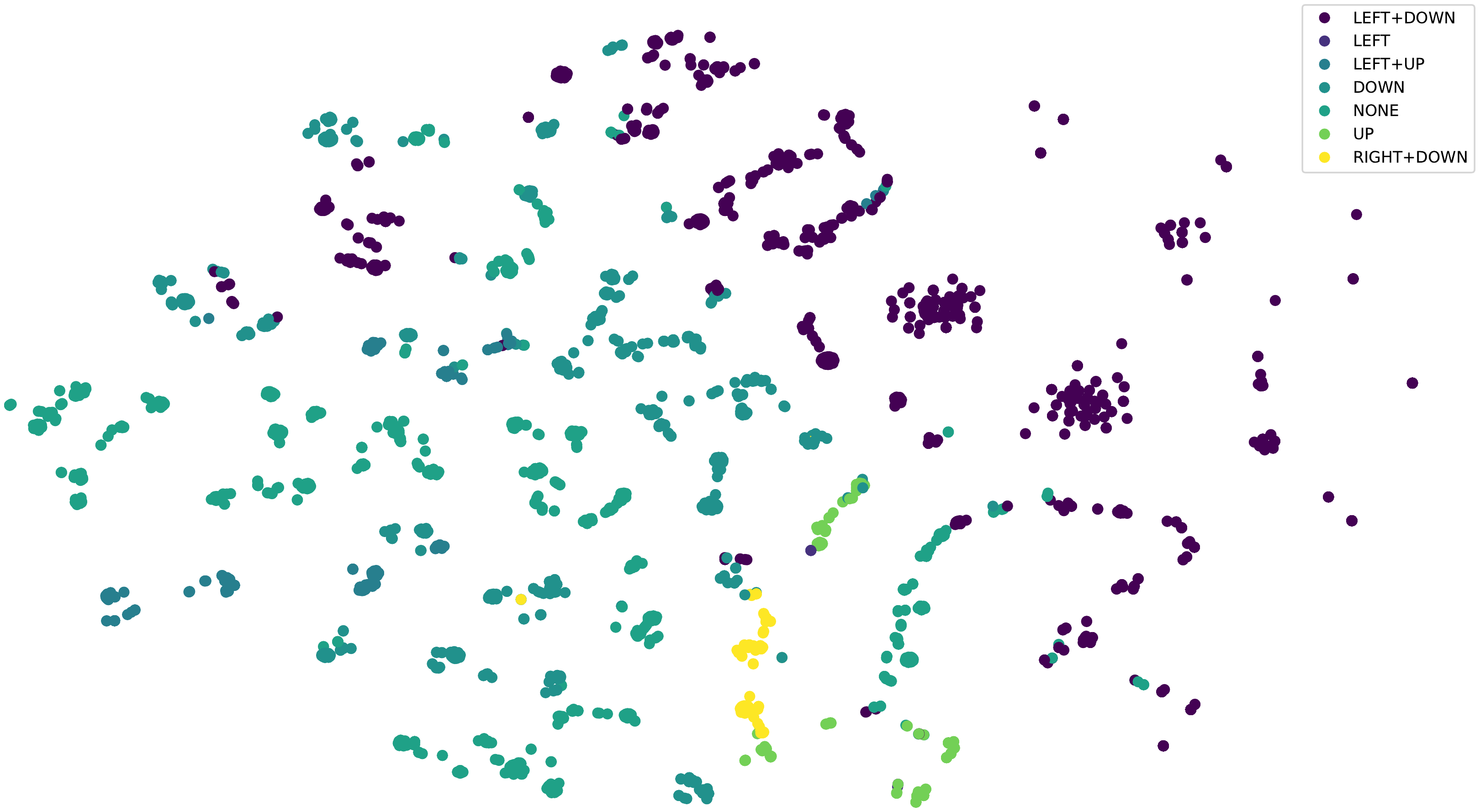}
    \vspace{1em}
    \includegraphics[width=0.49\textwidth]{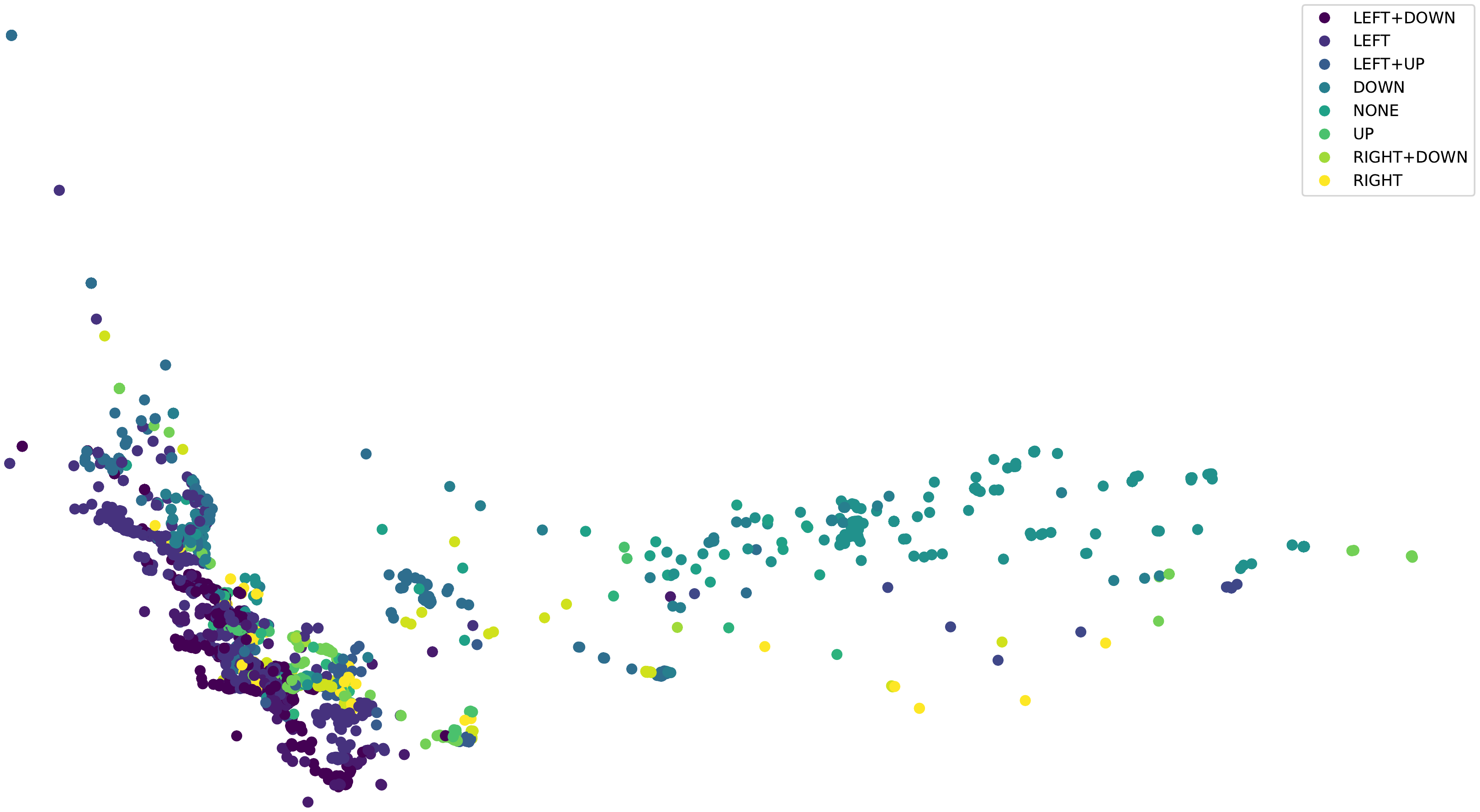}
    \includegraphics[width=0.49\textwidth]{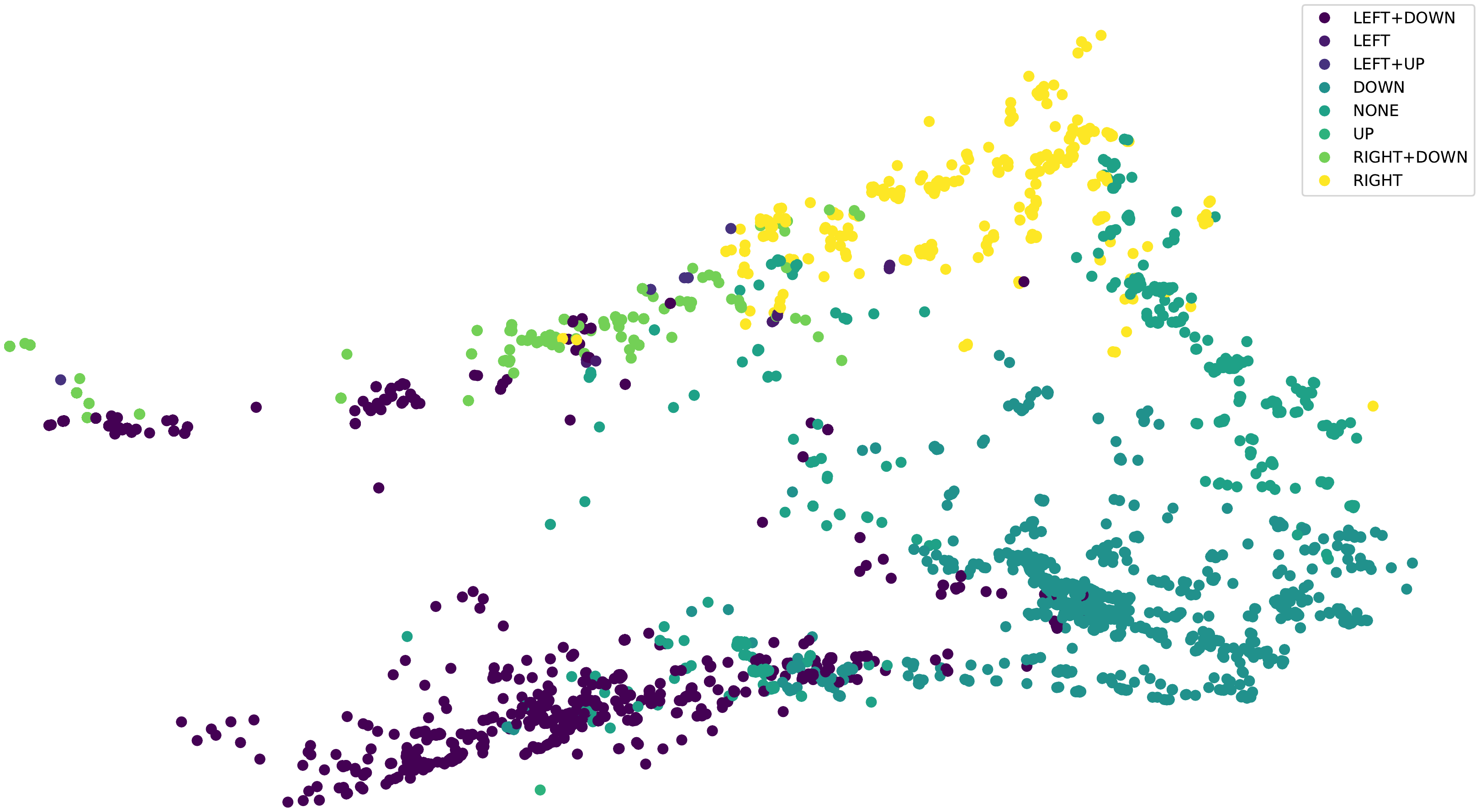}
    \caption{t-SNE (top) and PLS (bottom) plots of the state embedding for an expert (left) and the student (right) on Jumper.}
    \label{fig:tsne_jumper}
\end{figure}

Figures \ref{fig:tsne_breakout}, \ref{fig:tsne_pong}, \ref{fig:tsne_carnival}, \ref{fig:tsne_videopinball}, \ref{fig:tsne_coinrun}, \ref{fig:tsne_bigfish} and \ref{fig:tsne_jumper} report the t-SNE and PLS representations of the state embeddings for the experts and the students.

\end{document}